\documentclass[10pt,twocolumn,letterpaper]{article}
\usepackage{bbding}
\usepackage[pagenumbers]{cvpr}      
\usepackage{tabularx}
\usepackage{url}
\usepackage{graphicx}
\makeatletter
\def\hlinew#1{%
  \noalign{\ifnum0=`}\fi\hrule \@height #1 \futurelet
   \reserved@a\@xhline}
\makeatother
\usepackage{multirow}
\usepackage{algorithm}
\usepackage{float}
\usepackage{algorithmic}
\usepackage{amsmath,epstopdf,amssymb,color,url,multirow,verbatim,threeparttable,diagbox,makecell,booktabs}
\definecolor{light-gray}{gray}{0.82}
\definecolor{aliceblue}{rgb}{0.94,0.97,1.0}
\usepackage[table]{xcolor}
\usepackage{}
\usepackage{colortbl}
\usepackage{tcolorbox}
\definecolor{cvprblue}{rgb}{0.21,0.49,0.74}
\usepackage[pagebackref,breaklinks,colorlinks,allcolors=cvprblue]{hyperref}
\def\paperID{***} 
\def\confName{CVPR}
\def\confYear{2026}

\title{Your One-Stop Solution for AI-Generated Video Detection}

\author{
\textbf{Long Ma}\textsuperscript{1}\quad
\textbf{Zihao Xue}\textsuperscript{2}\quad
\textbf{Yan Wang}\textsuperscript{3}\quad
\textbf{Zhiyuan Yan}\textsuperscript{4}\quad \\
\textbf{Jin Xu}\textsuperscript{5}\quad
\textbf{Xiaorui Jiang}\textsuperscript{1}\quad
\textbf{Haiyang Yu}\textsuperscript{1,6}\quad
\textbf{Yong Liao}\textsuperscript{1,$\dag$}\quad
\textbf{Zhen Bi}\textsuperscript{2,$\dag$}\quad \\ [0.6em]
\textsuperscript{1}School of Cyber Science and Technology, University of Science and Technology of China\\
\textsuperscript{2}Huzhou University \quad
\textsuperscript{3}Alibaba Group\\
\textsuperscript{4}School of Electronic and Computer Engineering, Peking University\\
\textsuperscript{5}School of Information Science and Technology, University of Science and Technology of China\\
\textsuperscript{6}Institute of Dataspace, Hefei Comprehensive National Science Center\\[0.6em]
{\tt longm@mail.ustc.edu.cn, yliao@ustc.edu.cn, bizhen\_zju@zju.edu.cn}\\
{\small $^{\dag}$Corresponding authors.}}

\begin{document}
\let\oldtwocolumn\twocolumn 
\renewcommand\twocolumn[1][]{%
    \oldtwocolumn[{#1}{ 
    \begin{center}
    \vspace{-0.5cm}
        \includegraphics[width=0.9\textwidth]{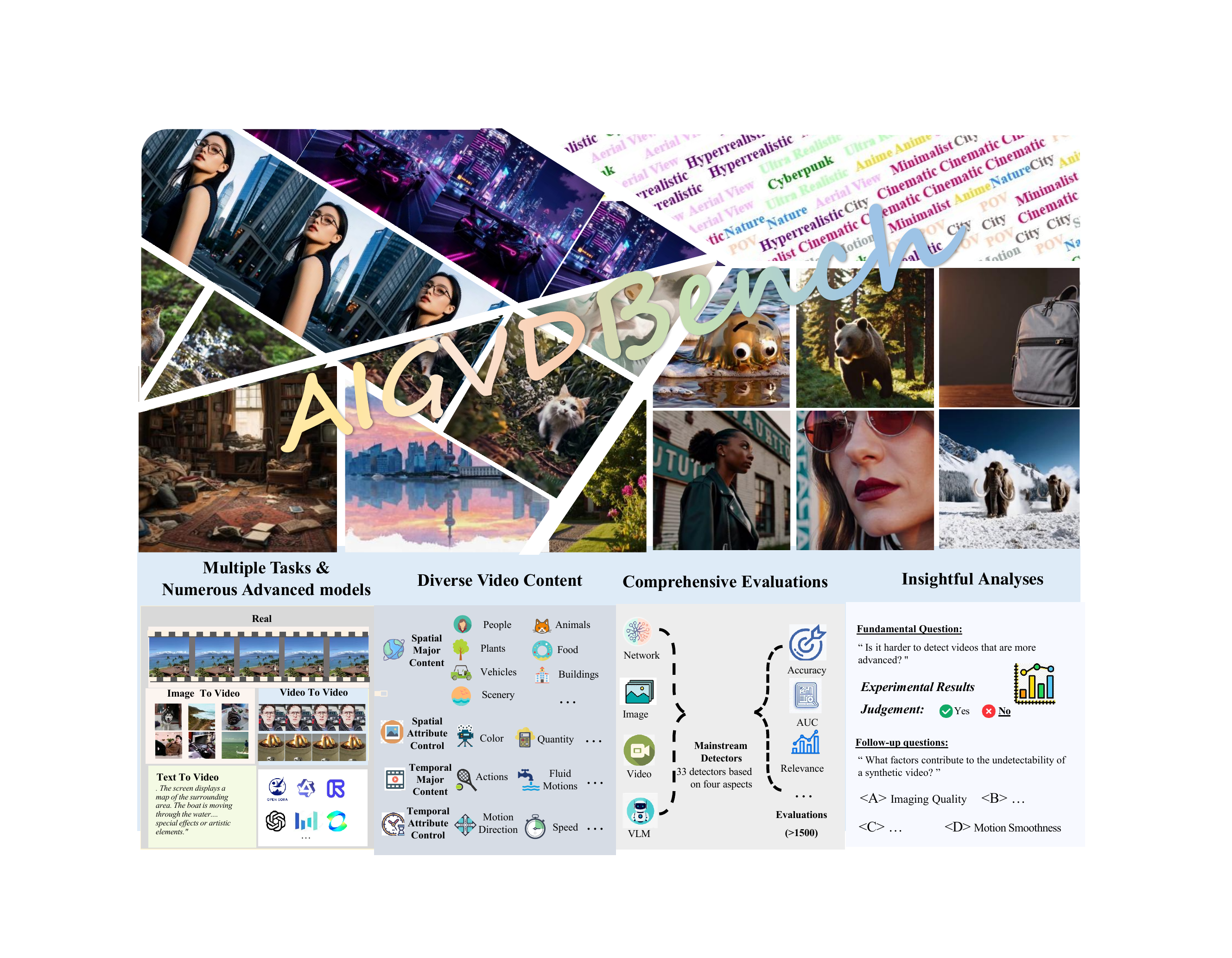} 
        \label{fig:fig1}
    \end{center}
    }] %
} 
\maketitle
\begin{abstract}
Recent advances in generative modeling can create remarkably realistic synthetic videos, making it increasingly difficult for humans to distinguish them from real ones and necessitating reliable detection methods. 
However, two key limitations hinder the development of this field.
\textbf{From the dataset perspective}, existing datasets are often limited in scale and constructed using outdated or narrowly scoped generative models, making it difficult to capture the diversity and rapid evolution of modern generative techniques. Moreover, the dataset construction process frequently prioritizes quantity over quality, neglecting essential aspects such as semantic diversity, scenario coverage, and technological representativeness. 
\textbf{From the benchmark perspective}, current benchmarks largely remain at the stage of dataset creation, leaving many fundamental issues and in-depth analysis yet to be systematically explored.

Addressing this gap, we propose AIGVDBench, a benchmark designed to be comprehensive and representative, covering \textbf{31} state-of-the-art generation models and over \textbf{440,000} videos. By executing more than \textbf{1,500} evaluations on \textbf{33} existing detectors belonging to four distinct categories. This work presents \textbf{8 in-depth analyses} from multiple perspectives and identifies \textbf{4 novel findings} that offer valuable insights for future research. We hope this work provides a solid foundation for advancing the field of AI-generated video detection.

Our benchmark is open-sourced at \url{https://github.com/LongMa-2025/AIGVDBench}.
\end{abstract}
 
\section{Introduction}
\label{sec:intro}
Between 2024 and 2025, the rise of AI-generated videos (\eg, Sora~\cite{Sora}, Veo~\cite{Veo}) has ignited widespread creative enthusiasm. However, their rapidly improving realism and accessibility pose formidable challenges to social trust, leading to a pervasive “synthetic skepticism” where viewers increasingly attribute suspicious videos to AI generated.

This trust crisis originates from the ongoing evolution of synthetic technology: it has advanced from deepfake face swapping~\cite{rossler2019faceforensics++,li2020celeb} to realistic image generation, and now to coherent video synthesis from text or images. Accordingly, detection efforts have expanded from initial deepfake detection~\cite{yan2023ucf,zhu2023gendet,zhao2021multi} to AI-generated image identification~\cite{effort,CNN-generated,Universal}. Yet, as synthesis has shifted entirely to video, detection research has failed to keep pace. Despite sporadic explorations~\cite{decof,DeMamba}, AI-generated video detection (AIGVD) remains significantly underdeveloped in both scale and depth compared to earlier detection tasks.
\begin{table*}[t]
\caption{Comparison with existing AI-generated video detection datasets. AIGVDBench significantly surpasses all others in scale, content diversity, coverage of generation tasks, and variety of video generation models. It comprises videos generated by \textbf{31} distinct models (\textbf{20} open-source and \textbf{11} closed-source), spanning a wide range of tasks including \textbf{23} text-to-video (T2V), \textbf{6} image-to-video (I2V), and \textbf{2} video-to-video (V2V) models,  with the closed-source models being primarily but not exclusively focused on T2V generation. 
}
\label{tab:cmp_other_datasets}
\footnotesize
\resizebox{\textwidth}{!}{
\begin{tabular}{l|ccccccccccc}
\hlinew{1.1pt}
Dataset&Publication&Latest Models&Methods&Open Source &  \multicolumn{1}{c}{Closed Source} & \multicolumn{1}{c}{T2V} & \multicolumn{1}{c}{I2V} & \multicolumn{1}{c}{V2V}&\multirow{1}{*}{Real world}&\multirow{1}{*}{Generated Videos}&\multirow{1}{*}{Content diversity}\\
\hline\hline
GVD~\cite{gvd} & PRCV'24 & Sora~\cite{Sora} (2024.2) & 11 & 3 & 8 & 8 &3 & - & \Checkmark & 11.6k&- \\
GVF~\cite{decof} & ICME'25 & kling~\cite{Kling} (2024.6) & 9 & 4 & 5 & 9 &- & - & \Checkmark & 4.2k&\Checkmark\kern-1.2ex\raisebox{1ex}{\rotatebox[origin=c]{125}{\textbf{--}}} \\
GenVideo~\cite{DeMamba} & ArXiv’24 & OpenSora~\cite{opensora} (2024.3) & 20 & 14 & 6 & 16 &4 & - & -& 100k&- \\
GenVidBench~\cite{genvidbench} & ArXiv’25 & Mora~\cite{Mora} (2024.3) & 8 & 7 & 1 & 6 &2 & -& -& 109.2k&\Checkmark\kern-1.2ex\raisebox{1ex}{\rotatebox[origin=c]{125}{\textbf{--}}}  \\
GenBuster-200K~\cite{busterx} & ArXiv’25 &  EasyAnimate~\cite{easyanimate} (2025.1) & 12 & 4 & 8 & 8 &- & - & -& 101.1k&\Checkmark\kern-1.2ex\raisebox{1ex}{\rotatebox[origin=c]{125}{\textbf{--}}}  \\

 GenWorld~\cite{Genworld} & ArXiv’25 &  Cosmos~\cite{cosmos} (2025.1) & 10 & 10 & - & 7 &2 & 1 & -& 89.4k&- \\

\midrule
\textbf{Ours} & - &Open-sora~\cite{opensorav2}(2025.3) & \textbf{31} & \textbf{20} & \textbf{11} & \textbf{23} & \textbf{6} & \textbf{2} & \Checkmark&\textbf{422k} &\Checkmark \\ 
\hlinew{1.1pt}
\end{tabular}
}
\vspace{0.1cm}
\begin{itemize}
    \item  \textbf{Latest Models}: The year when the \textit{latest}  video generation models was added to the dataset.
    \item \textbf{Methods}: The \textit{number} of different video generation models used to generate videos in the dataset.
    \item  \textbf{Model Type}: The methods are  classified into: open source and closed source.
    \item  \textbf{Generation tasks}: Based on  generation task types, the models are categorized into text-to-video (T2V), image-to-video (I2V), and video-to-video (V2V).
    \item \textbf{Generated Videos}: The number of \textit{generated} videos involved in the dataset.
    \item  \textbf{Real-world:} Whether the dataset includes videos that real-user scenarios, as opposed to being created from a predefined set of prompts.
    \item \textbf{Content Diversity:} Evaluates the balance and variety of the prompts used for video generation. The prompt sourcing strategy is labeled as non-automated (\Checkmark\kern-1.2ex\raisebox{1ex}{\rotatebox[origin=c]{125}{\textbf{--}}}) or automated (\Checkmark).
\end{itemize}
 
\end{table*}

\begin{table*}[]
\caption{Involved video generation methods of the AIGVDBench. We use the following 31 methods to create fake videos for evaluation.}
\label{tab:detail}
 \belowrulesep=0pt \aboverulesep=0pt
\footnotesize
\centering
\resizebox{0.95\textwidth}{!}{
\begin{tabular}{l|c|ccccccc} 
\hlinew{1.1pt}
Type & Sub-Types & ID-Number & Method & Venue & Time & Version & Data Used & Data Scale (Videos) \\ 
\hline 
\multirow{20}{*}{Open Source} 
& \multirow{12}{*}{Text to Video (T2V)} & 1 & Open-Sora~\cite{opensorav2}
  & ArXiv 2025 & 2025.3 & Open-Sora v2.0 & Train \& Test \dag & 20000 \\ 
 & & 2 & RepVideo~\cite{RepVideo}
   & ArXiv 2025 & 2025.1 & RepVideo v0.1 & Train \& Test & 20000 \\ 
 & & 3 & AccVideo~\cite{accvideo}
   & ArXiv 2025 & 2025.3 & AccVideo (Hunyuan) & Train \& Test & 20000 \\ 
& & 4 & CogVideoX~\cite{cogvideox}
   & ICLR 2025 & 2024.8 & CogVideoX-2B & Train \& Test & 20000 \\ 
& & 5 & EasyAnimate~\cite{easyanimate}
   & ArXiv 2024 & 2025.1 & EasyAnimate-V5.1-12B & Train \& Test & 20000 \\ 
& & 6 & Wan2.1~\cite{wan2025}
   & ArXiv 2025 & 2025.2 & Wan2.1 -T2V-1.3B & Train \& Test & 20000 \\ 
& & 7 & VideoCrafter~\cite{VideoCrafter2}
   & CVPR 2024 & 2024.1 & VideoCrafter2 & Train \& Test & 20000 \\ 
 & & 8 & Pyramid-Flow~\cite{pyramidal}
   & ICLR 2025 & 2024.10 & - & Train \& Test & 20000 \\ 
 & & 9 & IPOC~\cite{ipoc}
   & ArXiv 2025 & 2025.2 & IPOC-2B-v1.0 & Train \& Test & 20000 \\ 
 &  & 10 & Hunyuan~\cite{hunyuanvideo}
   & ArXiv 2024 & 2024.12 & - & Train \& Test & 20000 \\ 
 &   & 11 & LTX~\cite{LTXVideo}
   & ArXiv 2024 & 2024.12 & LTX v0.9.1 & Train \& Test & 20000 \\ 
 &    & 12 & AnimateDiff-L~\cite{animatediff}
   & ArXiv 2024 & 2024.3 & - & Train \& Test & 20000 \\ 
\cmidrule{2-9}
 & \multirow{6}{*}{Image to Video (I2V)} & 13 & Pyramid-Flow~\cite{pyramidal}
 & ICLR 2025 & 2024.10 & - & Train \& Test & 20000 \\ 
& & 14 & LTX~\cite{LTXVideo} & ArXiv 2025 & 2024.12 & LTX v0.9.1 & Train \& Test & 20000 \\
 & & 15 & EasyAnimate~\cite{easyanimate}
 & ArXiv 2024 & 2025.1 & EasyAnimate-V5.1-12B & Train \& Test & 20000 \\ 
 & & 16 & VideoCrafter~\cite{VideoCrafter2}
   & ArXiv 2023 & 2023.10 & VideoCrafter1 & Train \& Test & 20000 \\ 
 &  & 17 & SVD~\cite{SVD}
   & ArXiv 2023 & 2023.11 & SVD & Train \& Test & 20000 \\ 
 & & 18 & SEINE~\cite{seine}
& ICLR 2024 & 2023.11 & - & Train \& Test & 20000 \\ 
 \cmidrule{2-9}
  &\multirow{2}{*}{Video to Video (V2V)} & 19 & LTX~\cite{LTXVideo} & ArXiv 2025 & 2025.3 & LTX v0.9.5 & Train \& Test & 20000 \\ 
& & 20 & CogVideoX~\cite{cogvideox}
   & ICLR 2025 & 2024.8 & CogVideoX-2B & Train \& Test & 20000 \\ 
\midrule
\multirow{11}{*}{Closed Source}  
& \multirow{11}{*}{Real World} & 21 & Sora~\cite{Sora}
  & - & 2024.2 & - & Test & 2000 \\ 
&  & 22 & kling~\cite{Kling}
  & - & 2024.6 & Multi & Test & 2000 \\ 
  &  & 23 & Gen2~\cite{Gen-2}
  & - & 2023.2 & - & Test & 2000 \\ 
    &  & 24 & Gen3~\cite{Gen-3}
  & - & 2023.6 & Multi & Test & 2000 \\ 
      &  & 25 & Causvid*~\cite{yin2025causvid}
  & CVPR 2025 & 2024.12 & - & Test & 2000 \\
        &  & 26 & Luma~\cite{Luma}
  & - & 2024.6 & Multi & Test & 2000 \\ 
          &  & 27& Pika~\cite{Pika}
  & - & 2023.11 & Multi & Test & 2000 \\ 
          &  & 28& Open-Sora~\cite{opensora}
  & ArXiv 2025 & 2024.3 & Multi & Test & 2000 \\ 
            &  & 29& Wan~\cite{Wan}
  & - & 2025.2 & Multi & Test & 2000 \\ 
            &  & 30& Jimeng~\cite{Jimeng}
  & - & 2024.5 & Multi & Test & 2000 \\ 
              &  & 31& Vidu~\cite{Vidu}
  & - & 2024.5 & Multi & Test & 2000 \\ 
\hlinew{1.1pt}
\end{tabular}
}
\vspace{0.1cm}
\begin{itemize}
    \item  \textbf{Train \& Test}: For each open-source model, the dataset contains 20,000 generated videos, partitioned into training, validation, and test sets with a split of 14,000, 3,000, and 3,000 videos, respectively. Additionally, 20,000 real videos, each paired with a generated video based on the same prompt, are included and partitioned identically.  \dag Open-Sora is selected as the default training set.
    \item \textbf{Test}: To evaluate closed-source models, each is provided with a test set comprising 2,000 generated videos. A matched real-video test set is formed for each by selecting the first 2,000 videos from the complete real-video test set.
    \item  \textbf{Causvid*}:  During the dataset creation period, Causvid was still closed-source and was therefore classified as such.
\end{itemize}
\end{table*}

In the field of AI-generated video detection, researchers face two major limitations that impede research progress: \\
\textbf{From the dataset perspective}: Existing collections often rely on outdated generative models with limited variety~\cite{Genworld,genvidbench} and quantity~\cite{gvd,DeMamba,busterx}, further compounded by insufficient video sample sizes~\cite{decof}. Moreover, during dataset construction, the focus tends to be narrowly placed on expanding the number of videos, while overlooking critical aspects such as semantic diversity, comprehensive scene coverage, and balanced representation across different generation techniques. These shortcomings fundamentally undermine both the practical utility and generalizability of evaluation results.\\
\textbf{From the benchmark perspective}: Most current efforts remain at the preliminary stage of dataset creation, leaving a range of fundamental questions and in-depth analyses largely unaddressed. Several key issues have yet to be systematically investigated: How can AI-generated videos be effectively detected? Is it more effective to analyze the video sequence as a whole or to perform frame-by-frame recognition? Why are videos produced by certain models more easily detectable? Can training detectors with higher-quality generated samples improve their robustness and generalization? Conversely, does ongoing advancement in generative technology inevitably render existing detection methods obsolete? 

Given these limitations, there is a pressing need for a comprehensive benchmark to propel the field forward. In response, we propose AIGVDBench, a novel benchmark for AI-generated video detection designed to offer a more solid and instructive foundation for research. Our work makes the following key contributions:

(1) To address the challenges in building benchmarks for AI-generated video detection, we propose a standardized pipeline that ensures representativeness through attribute-balancing and comprehensive model selection, while guaranteeing replicability via alignment with public benchmarks and strict quality control.

(2) We construct a high-quality benchmark  comprising over 440,000 videos (more than 14 million images) from 20 open-source and 11 closed-source generative models, with 20,000 videos per open-source model and 2,000 per closed-source model simulated under real-world conditions.

(3) Leveraging this AIGVDBench, we systematically investigate four promising research directions and provide instructive insights through extensive experiments to inspire future work in generated video detection. 

We hope that this study will offer researchers a solid foundation and facilitate the crucial first step in advancing the field of AI-generated video detection.

\section{AIGVDBench}
In this work, we concentrate on detecting AI-generated videos, leaving the detection of deepfake faces outside the scope of our discussion. Our dataset encompasses a total of 31 generation methods, comprising 20 open-source and 11 closed-source approaches. These methods are further classified according to their generation paradigms, such as text-to-video, image-to-video, video-to-video, and real-world scenarios (where task types are unconstrained). Thanks to its substantial scale and methodological diversity, our dataset exhibits notable advantages over existing alternatives, as demonstrated in ~\cref{tab:cmp_other_datasets} and ~\cref{tab:detail}.
\begin{figure}[htbp]
    \centering

    \begin{subfigure}[t]{0.45\linewidth}
        \centering
        \includegraphics[width=\linewidth]{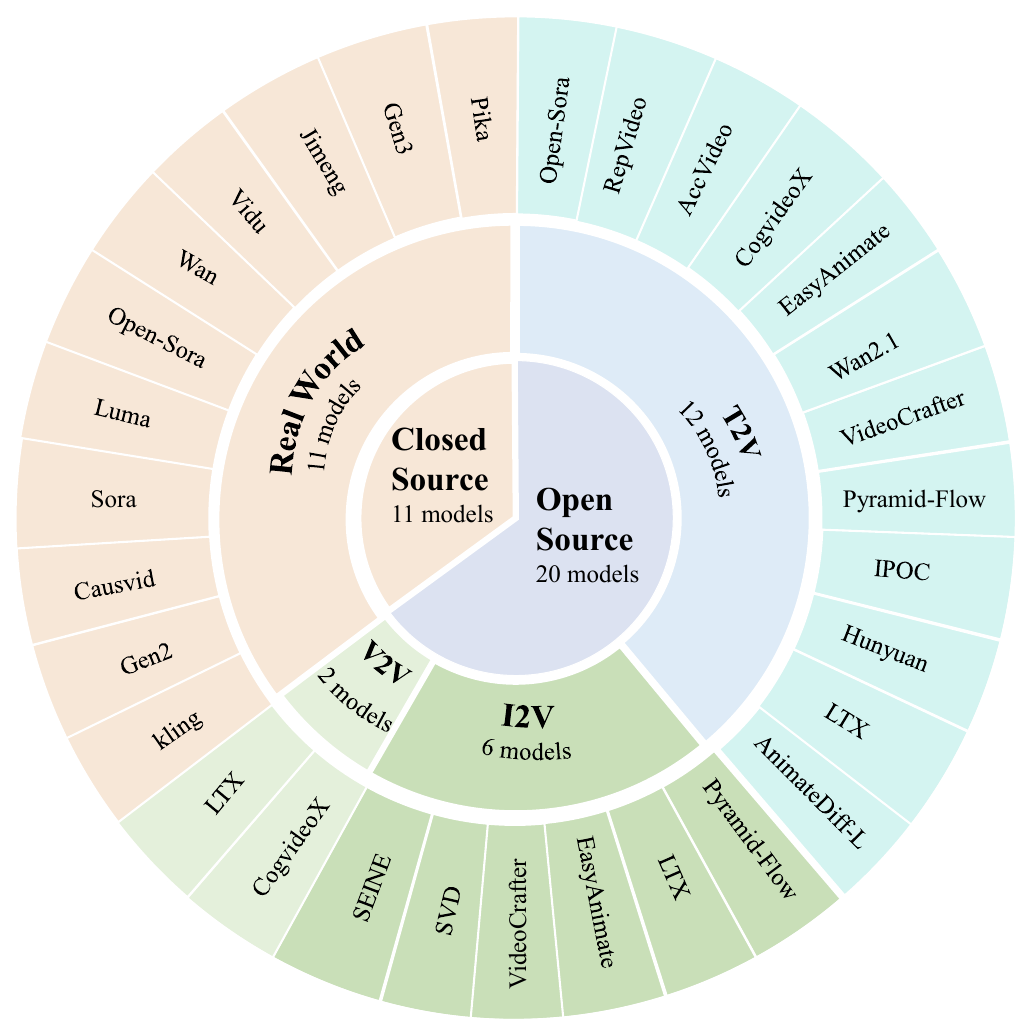}
        \caption{Video generation models}
        \label{fig:subfig_a}
    \end{subfigure}
    \hfill
    \begin{subfigure}[t]{0.45\linewidth}
        \centering
        \includegraphics[width=\linewidth]{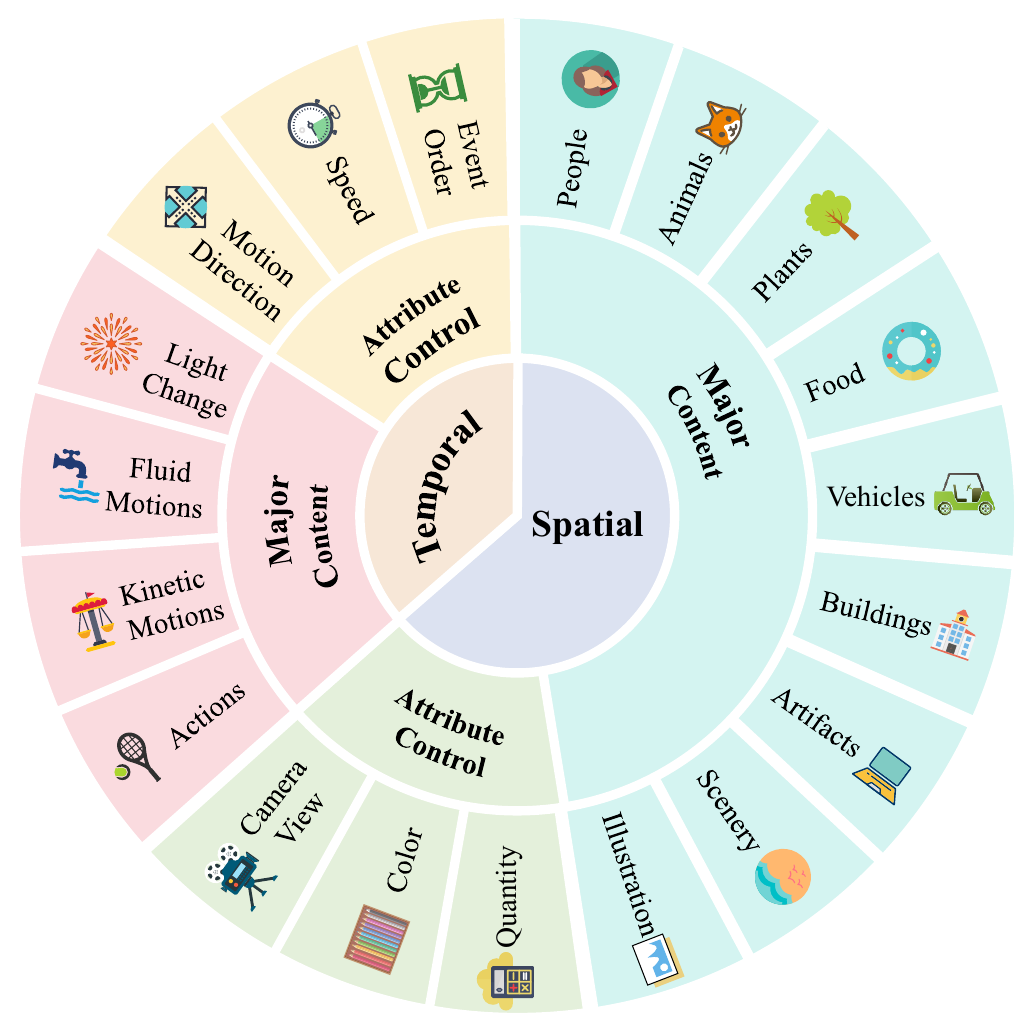}
        \caption{Categorization of prompts}
        \label{fig:main_subfig_b}
    \end{subfigure}

    \vspace{0.2cm}

    \begin{subfigure}[t]{\linewidth}
        \centering
        \includegraphics[width=\linewidth]{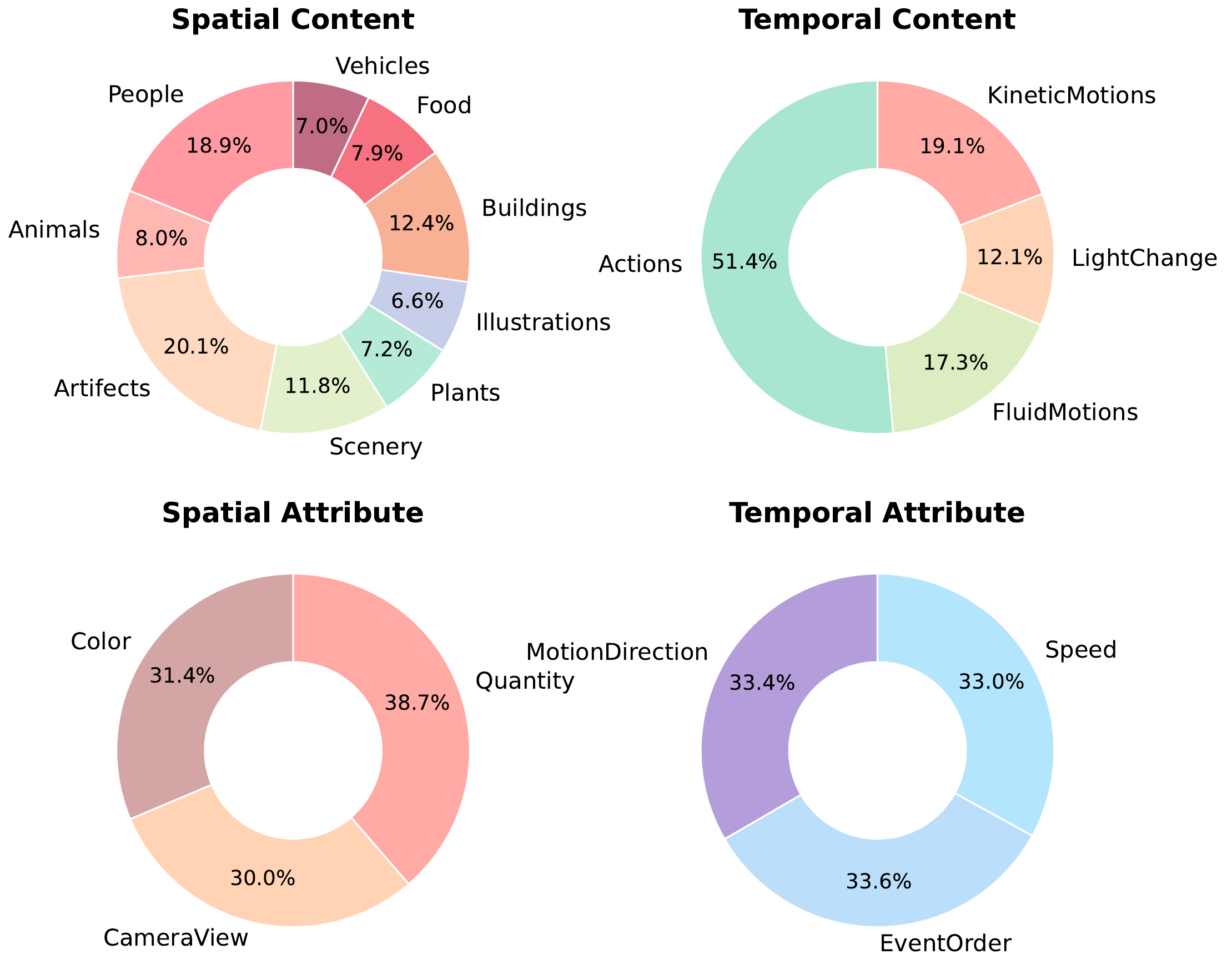}
        \caption{Balance in content distribution}
        \label{fig:subfig_c_single}
    \end{subfigure}

    \caption{Overview of AIGVDBench. Our benchmark demonstrates advantages in scale, content diversity, coverage of generation tasks, diversity of video generation models, and video realism.}
    \label{fig:overview_aigvdbench}
\end{figure}

\begin{figure}[htbp]
    \centering
    \includegraphics[width=\linewidth]{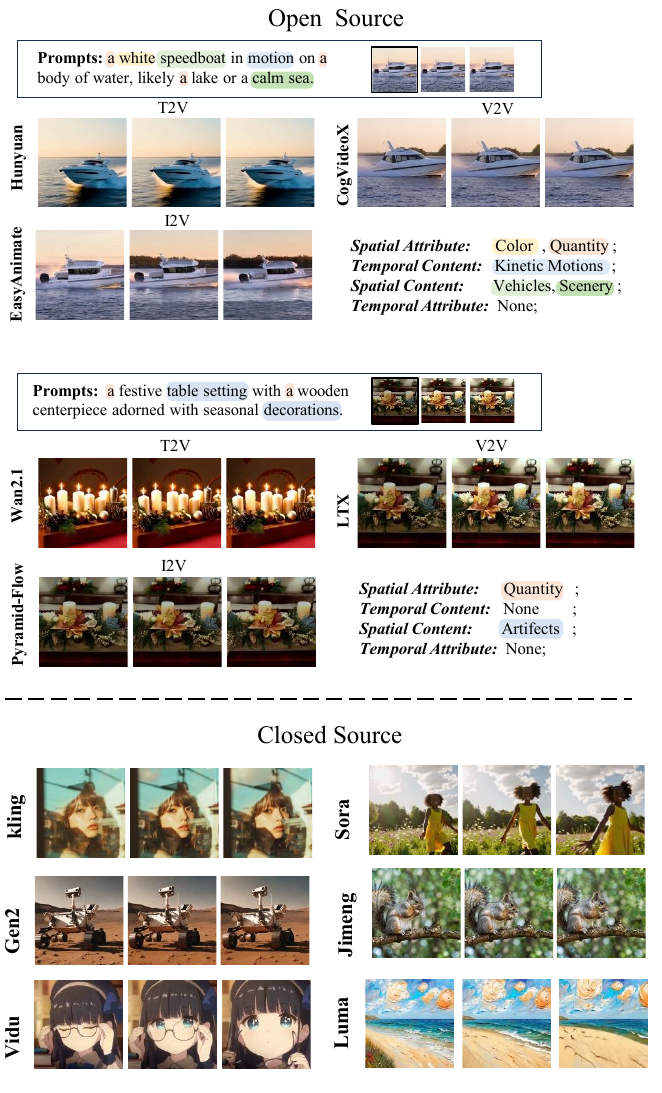}
    \caption{Visual examples of the AIGVDBench.} 
    \label{fig:subfig_d_single}
\end{figure}
\subsection{Selecting Prompts for Diversity.} 
A common practice for constructing datasets to detect AI-generated videos is to leverage real videos and their associated prompts from existing video datasets~\cite{nanopenvid,yang2024vript,wang2023videofactory,kay2017kinetics,xu2016msr} as source material. However, such datasets are typically large-scale and exhibit inherent distribution imbalances. Direct random sampling from them would therefore risk reinforcing these biases. Although prior studies have recognized this issue, a systematic solution remains lacking. For example, GVF~\cite{decof} manually curated prompts, while GenVidBench~\cite{genvidbench} and GenBuster~\cite{busterx} opted for random sampling from several relatively comprehensive datasets~\cite{yang2024vript,nanopenvid,wang2023videofactory}. None of these approaches, however, has fundamentally addressed the sampling bias problem.
\vspace{0.2cm}

\noindent\textbf{The Source of Real Videos:} To achieve a balanced selection of prompts, we first select OpenVid-HD~\cite{nanopenvid} as the source of real videos. This text-to-video dataset is relatively well-balanced in content distribution, providing a solid foundation for extracting attribute-balanced prompts. Additionally, the consistent resolution of its videos helps prevent  unintended bias in detector evaluation. 

\noindent\textbf{Prompts Categorization:} We then systematically categorized 400,000 prompts from this dataset using a structured attribute taxonomy. As shown in~\cref{fig:main_subfig_b}, prompts are organized along the two primary dimensions of "Main Content" and "Attribute Control", with a further subdivision based on their temporal and spatial characteristics, which results in 9 spatial content, 3 spatial attribute, 4 temporal content, and 3 temporal attribute categories. The detailed classification methodology is provided in the ~\cref{7.4} of Appendix.

\noindent\textbf{Attribute Balancing Selection Algorithm:} 
Given a pre-classified set of prompts $P$, our objective is to select a subset $P_B$ that exhibits a balanced distribution across all considered attributes. A key challenge in this task arises from the multi-label nature of the prompts—each prompt can be associated with multiple attributes simultaneously (\eg, containing both "People" and "Animals"). Such multi-label assignments introduce complex correlations between candidate prompts and the attribute space,  which complicate the selection process. To tackle this issue, we propose an Attribute Balancing Algorithm designed to systematically account for these dependencies while ensuring distributional balance in the resulting subset.

The algorithm proceeds in four main stages. First, based on the complexity of attribute combinations, the original set $P$ is divided into four mutually exclusive subsets: $P_{1}$ contains only spatial and temporal content; $P_{2}$ includes spatial content, temporal content, and temporal attributes; $P_{3}$ includes spatial content, temporal content, and spatial attributes; and $P_{4}$ contains spatial content, temporal content, temporal attributes, and spatial attributes. Next, each subset  $P_i$ , it is further partitioned into  $N_i$  categories based on the attribute categories it contains, where $N_i = \prod_{j=1}^k \text{Num}(\text{CLS}(j)) $,  $k$  denotes the number of attributes in    $P_i$ , and  $cls(j)$ denotes the set of categories under the  $j$-th attribute.

We then assign each prompt $p$ is assigned to one or more categories according to its attribute combination: if a prompt belongs to multiple categories, it is labeled as \textit{"Multi"}; if it belongs to only a single category, it is labeled as \textit{"Single"}. Before the selection process begins, the number of prompts in the category with the fewest prompts $m$ is used as the initial selection base.

An iterative selection process is then carried out as follows. Categories are sorted in ascending order based on their current prompt count, and selection begins with the category containing the fewest prompts.  Priority is given to prompts labeled \textit{"Single"};  if these are insufficient, \textit{"Multi"} prompts are supplemented; If the total number of prompts in a category is still inadequate, all remaining prompts in the category are selected. After completing the selection for a category, the chosen prompts are added to the total subset $P_B$. If a \textit{"Multi"} prompt in a given category $N_i$ has already been selected, it is removed from the current category, and the required number of prompts $S_{N_i}$ to be selected is reduced accordingly. This process repeats until all categories have been processed, ultimately resulting in a balanced subset $P_B$ in terms of attributes.  Through a combination of algorithmic filtering and manual screening, we arrived at a final collection of 20,000 real videos. The algorithm is described in ~\cref{alg:attribute_balanced}, beyond the qualitative validation in \cref{fig:subfig_c_single}, we further provide a quantitative appraisal in the ~\cref{7.5} of Appendix, contrasting the balance quality with prior datasets and thoroughly assessing the algorithm's efficacy.

\begin{algorithm}[t]
\caption{Attribute Balanced Selection}
\label{alg:attribute_balanced}
\begin{algorithmic}[1]
\REQUIRE Prompt set $P$ with attribute annotations
\ENSURE Balanced subset $P_B$ 

\STATE Partition $P$ into $\{P_1, P_2, P_3, P_4\}$ by attribute combinations
\STATE $P_B \gets \emptyset$
\FOR{each subset $P_i$}
    \STATE $\mathcal{C} \gets$ mapping where each category $c$ has:
    \STATE \quad $\bullet$ $\text{sing}$: single-attribute prompts
    \STATE \quad $\bullet$ $\text{multi}$: multi-attribute prompts
    
    \STATE $m \gets \min\limits_{c} |\mathcal{C}[c]|$ \COMMENT{Min category size}
    \STATE $\mathcal{L} \gets$ categories sorted by size ascending
    
    \WHILE{$\mathcal{L} \neq \emptyset$}
        \STATE $c \gets$ first category in $\mathcal{L}$
        \STATE Select $\min(m, |\mathcal{C}[c].\text{sing}|)$ from sing
        \STATE Select remaining from multi if needed
        \STATE Add to $P_B$
        
        \FOR{each $c_j \in \mathcal{L}$}
            \STATE Remove selected from $c_j$.multi
        \ENDFOR
        
        \STATE Remove $c$ from $\mathcal{L}$
        \STATE Re-sort $\mathcal{L}$ by updated size
    \ENDWHILE
\ENDFOR
\RETURN $P_B$
\end{algorithmic}
\end{algorithm}

\subsection{Selection of Generation Models.}
Benefiting from the model leaderboard provided by VBench~\cite{vbench}, we construct our dataset by selecting 20 open-source and 11 closed-source video generation models. The task categories, release dates, version details, and official links of these models are summarized in ~\cref{fig:subfig_a} and ~\cref{tab:detail}.
For open-source models, we cover three primary generation tasks: text-to-video, image-to-video, and video-to-video. Instead of relying solely on ranking, we also take into account factors such as model representativeness, GPU memory consumption, and inference speed under limited resources. We give priority to models that support multiple task types (\eg, LTX~\cite{LTXVideo} and EasyAnimate~\cite{easyanimate}) to facilitate a deeper exploration of the research questions outlined in Appendix.

As for closed-source models, we include 11 widely recognized and frequently used systems. Due to budget constraints, we collected 2,000 generated videos per model category. The video sources encompass the VBench evaluation set~\cite{vbench}, official demonstrations, community-shared examples, and self-generated samples. During collection, we imposed no restrictions on video content, resolution, or task type, aiming to create a more challenging and realistic test set. It is worth noting that the Veo3~\cite{veo3} and Sora2~\cite{Sora2} models were excluded from this study due to both regional access restrictions and the substantial workload involved in dataset construction and evaluation.

\subsection{Compression Format Unification.}
Numerous studies highlight that deepfake detection datasets can be biased by irrelevant low-level cues, particularly the systematic difference in compression formats between real (often PNG) and synthetic (often JPEG) content~\cite{Fakeorjpeg,wang2023dire,zhu2023genimage}. In response to concerns that such format discrepancies may introduce spurious correlations and compromise model fairness~\cite{seeing}, we preprocess all videos by standardizing them to the H.264 codec. This step ensures a consistent compression landscape, thereby mitigating a known source of bias. The implications of compression formats are further discussed in the ~\cref{7.6} of Appendix.

\begin{table*}[htbp]

  \caption{AUC Performance Comparison of Video Classification, Generated Image Detection, and  Generated Video Detection Models across Open-Source Generation Models.}
  \resizebox{\textwidth}{!}{
    \begin{tabular}{>{\centering}m{2.5cm}|*{20}{c|}c}
    \toprule
    \multirow{3}{*}{\textbf{Method}} & \multicolumn{6}{c|}{\textbf{I2V}} & \multicolumn{12}{c|}{\textbf{T2V}} & \multicolumn{2}{c|}{\textbf{V2V}} & \multirow{3}{*}{\textbf{AVG}} \\
    \cmidrule(lr){2-7} \cmidrule(lr){8-19} \cmidrule(lr){20-21}
    & \multirow{2}{*}{\makecell[c]{\scriptsize Easy\\\scriptsize Animate}} & \multirow{2}{*}{\makecell[c]{\scriptsize LTX}} & \multirow{2}{*}{\makecell[c]{\scriptsize Pyramid\\\scriptsize Flow}} & \multirow{2}{*}{\makecell[c]{\scriptsize SEINE}} & \multirow{2}{*}{\makecell[c]{\scriptsize SVD}} & \multirow{2}{*}{\makecell[c]{\scriptsize Video\\\scriptsize Crafter}} & \multirow{2}{*}{\makecell[c]{\scriptsize Acc\\\scriptsize Video}} & \multirow{2}{*}{\makecell[c]{\scriptsize Animate\\\scriptsize Diff}} & \multirow{2}{*}{\makecell[c]{\scriptsize Cogvideo\\\scriptsize x1.5}} & \multirow{2}{*}{\makecell[c]{\scriptsize Easy\\\scriptsize Animate}} & \multirow{2}{*}{\makecell[c]{\scriptsize Hunyuan}} & \multirow{2}{*}{\makecell[c]{\scriptsize IPOC}} & \multirow{2}{*}{\makecell[c]{\scriptsize LTX}} & \multirow{2}{*}{\makecell[c]{\scriptsize Open\\\scriptsize Sora}} & \multirow{2}{*}{\makecell[c]{\scriptsize Pyramid\\\scriptsize Flow}} & \multirow{2}{*}{\makecell[c]{\scriptsize Rep\\\scriptsize Video}} & \multirow{2}{*}{\makecell[c]{\scriptsize Video\\\scriptsize Crafter}} & \multirow{2}{*}{\makecell[c]{\scriptsize Wan\\\scriptsize 2.1}} & \multirow{2}{*}{\makecell[c]{\scriptsize Cogvideo\\\scriptsize x1.5}} & \multirow{2}{*}{\makecell[c]{\scriptsize LTX}} & \\
    & & & & & & & & & & & & & & & & & & & & & \\
    \midrule
    
    \multicolumn{22}{c}{\textbf{Video Classification Models}} \\
    \midrule
    
    MViTv2~\cite{MViTv2} & 52.26 & 67.74 & 65.93 & 72.64 & 64.57 & 70.45 & 64.85 & 72.82 & 63.61 & 60.65 & 69.13 & 59.28 & 69.52 & 100.00 & 70.02 & 56.33 & 68.61 & 48.72 & 62.21 & 64.93 & 66.21 \\
    UniFormer~\cite{uniformer}& 48.46 & 60.05 & 66.36 & 44.41 & 32.03 & 68.78 & 85.00 & 80.53 & 67.57 & 65.95 & 84.76 & 72.75 & 52.92 & 89.49 & 84.92 & 66.43 & 83.73 & 55.38 & 75.06 & 69.35 & 67.70 \\
    VideoSwin~\cite{videoswin} & 60.43 & 70.88 & 65.63 & 90.68 & 85.10 & 77.10 & 62.89 & 73.71 & 65.64 & 62.99 & 60.80 & 63.69 & 70.08 & 100.00 & 63.88 & 60.15 & 76.33 & 54.95 & 57.49 & 66.52 & 69.45 \\
     VideoMAE~\cite{videomae}& 57.50 & 69.09 & 77.04 & 69.39 & 66.95 & 65.40 & 78.65 & 66.42 & 75.33 & 66.88 & 80.99 & 83.60 & 75.01 & 98.85 & 92.66 & 74.40 & 86.68 & 62.66 & 71.25 & 71.54 & 74.52 \\
      TSM~\cite{lin2019tsm} & 79.41 & 70.38 & 85.88 & 89.65 & 93.82 & 84.25 & 60.07 & 58.39 & 68.53 & 77.64 & 70.59 & 72.68 & 80.06 & 100.00 & 87.01 & 68.86 & 94.45 & 55.97 & 68.89 & 66.67 & 76.66 \\
    SlowFast~\cite{slowfast} & 63.92 & 71.62 & 83.83 & 74.02 & 72.29 & 68.36 & 82.71 & 71.03 & 80.23 & 74.91 & 83.52 & 89.42 & 76.22 & 99.78 & 92.75 & 79.87 & 91.33 & 65.91 & 78.86 & 77.50 & 78.90 \\
    
     TimeSformer~\cite{timesformer} & 66.33 & 70.38 & 80.32 & 82.08 & 88.27 & 65.38 & 83.86 & 76.31 & 84.55 & 84.34 & 84.81 & 91.30 & 85.51 & 100.00 & 96.17 & 85.13 & 96.42 & 73.66 & 69.63 & 65.55 & 81.50 \\
     I3D~\cite{i3d} & 81.00 & 81.68 & 94.01 & 96.67 & 96.13 & 84.79 & 66.62 & 46.75 & 83.49 & 78.40 & 75.47 & 90.21 & 94.62 & 100.00 & 96.75 & 80.28 & 91.49 & 65.30 & 79.74 & 76.40 & 82.99 \\

    UniFormerV2~\cite{uniformerv2} & 73.94 & 77.59 & 77.11 & 82.38 & 94.29 & 82.28 & 82.48 & 86.58 & 82.95 & 84.35 & 81.54 & 84.47 & 82.60 & 100.00 & 92.95 & 79.86 & 97.14 & 76.38 & 73.94 & 72.65 & 83.27 \\

    X3D~\cite{x3d} & 90.98 & 68.43 & 92.77 & 96.12 & 97.53 & 96.56 & 79.47 & 67.78 & 76.92 & 93.81 & 80.44 & 84.32 & 84.17 & 100.00 & 97.06 & 78.81 & 97.48 & 63.56 & 73.52 & 61.67 & 84.07 \\
    \midrule
    
    \multicolumn{22}{c}{\textbf{AI-Generated Image Detection Models}} \\
    \midrule
        NPR~\cite{NPR} & 69.58 & 67.60 & 77.80 & 83.69 & 92.90 & 59.54 & 53.33 & 24.19 & 75.61 & 65.39 & 60.23 & 70.01 & 80.67 & 100.00 & 85.35 & 68.48 & 70.85 & 48.08 & 64.72 & 58.58 & 68.83 \\
            FreDect~\cite{freqdet} & 59.70 & 61.23 & 69.19 & 71.29 & 96.10 & 77.62 & 66.12 & 49.33 & 71.51 & 67.01 & 69.02 & 68.49 & 70.06 & 100.00 & 73.32 & 67.17 & 81.54 & 54.75 & 65.00 & 52.62 & 69.55 \\
        Fusing~\cite{fusing} & 81.69 & 73.76 & 93.07 & 93.94 & 97.43 & 77.91 & 62.75 & 47.86 & 81.90 & 76.79 & 70.94 & 80.41 & 90.64 & 100.00 & 95.98 & 75.84 & 87.98 & 57.19 & 80.28 & 64.38 & 79.54 \\
            Gram-Net~\cite{Gram-Net} & 77.22 & 71.57 & 88.61 & 87.68 & 98.74 & 77.04 & 71.52 & 46.04 & 84.00 & 81.26 & 81.49 & 85.47 & 92.95 & 99.99 & 95.48 & 77.63 & 93.49 & 62.30 & 79.61 & 59.62 & 80.59 \\
            D3~\cite{d3} & 74.69 & 77.69 & 82.45 & 81.89 & 92.65 & 42.52 & 79.62 & 64.73 & 90.09 & 84.82 & 83.23 & 95.64 & 83.82 & 100.00 & 98.37 & 88.23 & 95.37 & 83.46 & 72.69 & 77.86 & 82.49 \\
    CNNSpot~\cite{CNN-generated} & 88.54 & 81.41 & 97.33 & 97.19 & 99.78 & 73.17 & 73.96 & 49.72 & 89.74 & 88.20 & 78.61 & 92.16 & 96.82 & 100.00 & 99.21 & 85.80 & 91.97 & 71.46 & 85.39 & 71.81 & 85.61 \\
        UnivFD~\cite{Universal} & 87.39 & 89.91 & 89.57 & 96.42 & 97.40 & 81.12 & 60.13 & 84.01 & 84.69 & 88.73 & 69.45 & 88.52 & 94.70 & 99.99 & 97.39 & 81.16 & 97.55 & 61.36 & 75.02 & 89.38 & 85.69 \\
    ForgeLens3~\cite{ForgeLens} & 95.79 & 89.67 & 97.78 & 99.92 & 99.96 & 98.76 & 81.06 & 72.92 & 84.34 & 96.17 & 80.51 & 91.22 & 93.76 & 100.00 & 99.59 & 87.98 & 99.50 & 66.25 & 69.98 & 86.79 & 89.60 \\

    Effort~\cite{effort} & 88.74 & 79.15 & 94.99 & 97.15 & 99.64 & 58.25 & 82.79 & 88.99 & 89.31 & 97.01 & 85.38 & 93.40 & 83.13 & 100.00 & 99.87 & 84.29 & 99.61 & 78.63 & 71.05 & 78.42 & 87.49 \\
    ForgeLens1~\cite{ForgeLens} & 98.21 & 92.89 & 98.68 & 99.94 & 99.95 & 98.86 & 79.18 & 95.84 & 87.03 & 98.68 & 75.03 & 88.81 & 95.63 & 100.00 & 99.71 & 88.06 & 99.81 & 74.87 & 73.85 & 91.41 & 91.82 \\

    \midrule
    
    \multicolumn{22}{c}{\textbf{AI-Generated Video Detection Models}} \\
    \midrule
    DeMamba~\cite{DeMamba} & 76.37 & 77.59 & 92.84 & 94.76 & 98.94 & 52.02 & 76.82 & 64.82 & 82.17 & 76.59 & 73.80 & 90.93 & 85.28 & 100.00 & 98.90 & 78.70 & 88.01 & 66.40 & 67.48 & 77.46 & 80.99 \\
    DeCoF~\cite{decof} & 70.64 & 81.53 & 83.74 & 85.67 & 93.12 & 45.21 & 79.52 & 68.11 & 90.71 & 78.03 & 79.82 & 94.64 & 87.74 & 99.90 & 96.57 & 89.49 & 90.77 & 72.90 & 81.84 & 81.34 & 82.56 \\
    
    \bottomrule
    \end{tabular}%
  }
  \label{tab:video_detection_benchmark}%
\end{table*}%
\begin{table*}[htbp]
  \centering
  \caption{ACC Performance Comparison of Vision-Language Models on Open-Source Generative Models. This comparison uses ACC because VLMs produce binary decisions, which preclude the calculation of AUC.}
  
  \resizebox{\textwidth}{!}{
    \footnotesize
    \begin{tabular}{>{\centering\arraybackslash}m{2.8cm}|*{20}{c|}c}
    \toprule
    \multirow{3}{*}{\textbf{Method}} & \multicolumn{6}{c|}{\textbf{I2V}} & \multicolumn{12}{c|}{\textbf{T2V}} & \multicolumn{2}{c|}{\textbf{V2V}} & \multirow{3}{*}{\textbf{AVG}} \\
    \cmidrule(lr){2-7} \cmidrule(lr){8-19} \cmidrule(lr){20-21}
    & \makecell[c]{\scriptsize Easy\\\scriptsize Animate} & \makecell[c]{\scriptsize LTX} & \makecell[c]{\scriptsize Pyramid\\\scriptsize Flow} & \makecell[c]{\scriptsize SEINE} & \makecell[c]{\scriptsize SVD} & \makecell[c]{\scriptsize Video\\\scriptsize Crafter} & \makecell[c]{\scriptsize Acc\\\scriptsize Video} & \makecell[c]{\scriptsize Animate\\\scriptsize Diff} & \makecell[c]{\scriptsize Cogvideo\\\scriptsize x1.5} & \makecell[c]{\scriptsize Easy\\\scriptsize Animate} & \makecell[c]{\scriptsize Hunyuan} & \makecell[c]{\scriptsize IPOC} & \makecell[c]{\scriptsize LTX} & \makecell[c]{\scriptsize Open\\\scriptsize Sora} & \makecell[c]{\scriptsize Pyramid\\\scriptsize Flow} & \makecell[c]{\scriptsize Rep\\\scriptsize Video} & \makecell[c]{\scriptsize Video\\\scriptsize Crafter} & \makecell[c]{\scriptsize Wan\\\scriptsize 2.1} & \makecell[c]{\scriptsize Cogvideo\\\scriptsize x1.5} & \makecell[c]{\scriptsize LTX} & \\
    \midrule
    Emu3-Stage1~\cite{wang2024emu3} & 33.22  & 33.57  & 33.03  & 32.27  & 28.70  & 32.82  & 31.03  & 40.60  & 31.82  & 32.32  & 30.07  & 31.28  & 33.67  & 27.65  & 30.65  & 31.67  & 42.87  & 31.95  & 34.65  & 35.32  & 32.96 \\
    FastVLM-7B~\cite{fastvlm} & 46.47  & 47.68  & 47.37  & 47.60  & 50.55  & 49.50  & 47.57  & 48.97  & 48.03  & 46.65  & 47.45  & 48.43  & 48.68  & 48.62  & 46.80  & 48.60  & 50.92  & 46.47  & 48.12  & 47.52  & 48.10  \\
   DeepseekVL2-S~\cite{DeepSeek-VL2}& 49.43  & 49.45  & 49.45  & 49.93  & 49.48  & 49.43  & 49.42  & 49.42  & 49.42  & 49.42  & 49.40  & 49.43  & 49.48  & 49.40  & 49.62  & 49.45  & 49.42  & 49.40  & 49.45  & 49.45  & 49.47 \\
   LLaVA-v1.5-7B~\cite{liu2023improvedllava} & 50.00 & 50.00 & 50.00 & 50.00 & 50.00 & 50.00 & 50.00 & 50.00 & 50.00 & 50.00 & 50.00 & 50.00 & 50.00 & 50.00 & 50.00 & 50.00 & 50.00 & 50.00 & 50.00 & 50.00 & 50.00 \\
   Kimi-VL-A3B~\cite{Kimi-vl} & 49.97  & 50.05  & 50.08  & 50.08  & 50.85  & 50.63  & 50.22  & 50.62  & 50.22  & 50.08  & 50.12  & 50.12  & 50.27  & 49.93  & 50.52  & 50.30  & 50.38  & 49.88  & 50.12  & 50.07  & 50.23 \\
    Deepseek-VL-7B~\cite{lu2024deepseekvl} & 50.02 & 50.13 & 50.10 & 50.12 & 51.20 & 50.57 & 50.23 & 50.02 & 50.27 & 50.12 & 50.10 & 50.05 & 50.80 & 50.00 & 50.15 & 50.25 & 50.05 & 50.00 & 50.35 & 50.15 & 50.23 \\
    Qwen2.5-VL-3B~\cite{Qwen2.5-VL} & 49.30  & 50.13  & 49.95  & 49.60  & 51.37  & 53.22  & 49.87  & 53.77  & 50.17  & 48.93  & 49.85  & 50.02  & 50.48  & 48.63  & 49.03  & 50.32  & 52.23  & 49.28  & 50.68  & 50.25  & 50.35 \\
    Qwen2.5-VL-32B~\cite{Qwen2.5-VL} & 49.45  & 50.03  & 49.80  & 49.43  & 50.78  & 53.32  & 51.07  & 55.58  & 51.87  & 49.88  & 50.17  & 51.58  & 50.13  & 49.82  & 50.63  & 52.18  & 55.42  & 49.75  & 51.20  & 50.32  & 51.12   \\
    DeepseekVL2~\cite{DeepSeek-VL2} & 50.12  & 51.38  & 51.10  & 52.05  & 56.37  & 55.25  & 50.28  & 51.27  & 51.40  & 49.32  & 50.08  & 50.87  & 53.48  & 49.87  & 52.45  & 52.50  & 51.10  & 49.62  & 50.95  & 51.07  & 51.53  \\
    Qwen2.5-VL-7B~\cite{Qwen2.5-VL} & 50.20  & 50.98  & 50.93  & 50.97  & 53.05  & 52.95  & 51.95  & 54.62  & 51.70  & 50.50  & 51.73  & 52.05  & 51.35  & 50.73  & 51.28  & 53.22  & 55.30  & 51.10  & 51.13  & 50.98  & 51.84  \\

    InternVL-8B~\cite{zhu2025internvl3} & 48.88  & 53.22  & 51.45  & 53.27  & 60.95  & 62.62  & 55.03  & 62.62  & 55.93  & 48.05  & 54.02  & 56.28  & 56.53  & 53.47  & 54.22  & 57.50  & 67.12  & 52.53  & 53.57  & 52.30  & 55.48  \\

    \bottomrule
    \end{tabular}%
  }
  \label{tab:vision_language_video_detection}%
\end{table*}%

\begin{figure*}[tbp]
  \begin{minipage}{0.68\textwidth}
   \centering
    \captionof{table}{\centering AUC Performance Comparison of Video Classification, Generated Image Detection, and Generated Video Detection Models across Closed-Source Generation Models.}
    \resizebox{\textwidth}{!}{
      \begin{tabular}{>{\centering\arraybackslash}m{3cm}|*{11}{c|}c}
      \toprule
      \multirow{3}{*}{\textbf{Method}} & \multicolumn{11}{c|}{\textbf{Closed-Source Approaches}} & \multirow{3}{*}{\textbf{AVG}} \\
      \cmidrule(lr){2-12}
      & \makecell[c]{\scriptsize Luma} & \makecell[c]{\scriptsize Open\\\scriptsize Sora} & \makecell[c]{\scriptsize Sora} & \makecell[c]{\scriptsize Causvid\\\scriptsize 24fps} & \makecell[c]{\scriptsize Gen2} & \makecell[c]{\scriptsize Gen3} & \makecell[c]{\scriptsize Jimeng} & \makecell[c]{\scriptsize Kling} & \makecell[c]{\scriptsize Pika} & \makecell[c]{\scriptsize Vidu} & \makecell[c]{\scriptsize Wan} &  \\
      \midrule
      
      \multicolumn{13}{c}{\textbf{Video Classification Models}} \\
      \midrule
        VideoSwin~\cite{videoswin} & 44.62 & 55.43 & 48.17 & 32.96 & 59.02 & 67.27 & 50.80 & 45.77 & 62.81 & 48.15 & 62.13 & 52.47 \\
      TSM~\cite{lin2019tsm} & 48.69 & 68.99 & 47.67 & 28.96 & 70.39 & 62.77 & 39.27 & 52.14 & 71.51 & 62.18 & 64.25 & 56.08 \\
        MViTv2~\cite{MViTv2} & 53.76 & 67.64 & 43.85 & 35.40 & 68.10 & 62.27 & 60.79 & 53.21 & 69.00 & 50.66 & 61.64 & 56.94 \\
      VideoMAE~\cite{videomae} & 52.78 & 70.62 & 42.85 & 56.13 & 61.45 & 63.46 & 31.91 & 51.20 & 80.69 & 64.23 & 58.95 & 57.66 \\
      Uniformer~\cite{uniformer} & 51.11 & 65.09 & 42.76 & 24.31 & 79.04 & 57.23 & 51.77 & 51.26 & 83.74 & 61.19 & 59.70 & 57.02 \\
      I3D~\cite{i3d} & 56.72 & 75.98 & 51.04 & 73.30 & 56.82 & 65.53 & 27.48 & 57.29 & 70.14 & 73.05 & 65.62 & 61.18 \\
    
      SlowFast~\cite{slowfast} & 53.28 & 75.41 & 48.21 & 56.74 & 67.17 & 69.31 & 44.47 & 54.30 & 86.28 & 66.57 & 68.12 & 62.71 \\
    X3D~\cite{x3d} & 54.97 & 79.31 & 52.42 & 75.64 & 79.14 & 64.73 & 52.40 & 55.60 & 77.40 & 66.94 & 67.93 & 66.04 \\

      TimeSformer~\cite{timesformer} & 84.15 & 94.92 & 85.25 & 90.12 & 86.92 & 86.93 & 84.19 & 80.34 & 92.90 & 84.07 & 81.73 & 86.50 \\

      UniformerV2~\cite{uniformerv2} & 82.61 & 90.81 & 81.45 & 88.81 & 94.67 & 85.53 & 97.56 & 85.68 & 86.46 & 77.96 & 86.20 & 87.07 \\

      \midrule
      
      \multicolumn{13}{c}{\textbf{AI-Generated Image Detection Models}} \\
      \midrule
        NPR~\cite{NPR} & 52.90 & 56.69 & 57.64 & 54.71 & 41.10 & 58.69 & 17.97 & 48.23 & 66.97 & 67.49 & 53.54 & 52.36 \\
      FreDect~\cite{freqdet} & 59.85 & 56.16 & 53.75 & 39.12 & 48.33 & 54.53 & 52.70 & 55.86 & 78.15 & 59.28 & 54.76 & 55.68 \\
      UnivFD~\cite{Universal} & 66.38 & 87.45 & 65.96 & 94.25 & 42.05 & 54.58 & 29.35 & 72.18 & 64.65 & 82.31 & 58.70 & 65.26 \\
      Fusing~\cite{fusing} & 66.98 & 72.11 & 66.15 & 63.22 & 48.92 & 67.33 & 45.66 & 65.60 & 82.26 & 79.65 & 62.23 & 65.46 \\
      Gram-Net~\cite{Gram-Net} & 70.33 & 75.63 & 62.84 & 75.17 & 44.20 & 67.32 & 39.47 & 66.60 & 86.00 & 74.12 & 60.84 & 65.68 \\
      CNNSpot~\cite{CNN-generated} & 69.49 & 84.83 & 74.15 & 88.26 & 42.41 & 73.42 & 41.48 & 65.53 & 82.93 & 86.27 & 67.79 & 70.60 \\
      D3~\cite{d3} & 75.22 & 96.05 & 75.92 & 94.62 & 50.30 & 77.22 & 49.97 & 78.00 & 81.98 & 84.92 & 78.13 & 76.58 \\
      ForgeLens3~\cite{ForgeLens} & 62.81 & 78.84 & 69.04 & 98.11 & 75.80 & 79.62 & 89.33 & 61.67 & 84.47 & 73.83 & 72.07 & 76.87 \\
      ForgeLens1~\cite{ForgeLens} & 74.18 & 85.54 & 86.57 & 99.58 & 86.56 & 88.58 & 97.81 & 74.55 & 85.92 & 76.84 & 79.23 & 85.03 \\
      Effort~\cite{effort} & 92.14 & 99.45 & 92.82 & 99.54 & 95.82 & 91.81 & 97.00 & 94.77 & 89.18 & 93.26 & 88.72 & 94.05 \\

      \midrule
      
      \multicolumn{13}{c}{\textbf{AI-Generated Video Detection Models}} \\
      \midrule
      DeMamba~\cite{DeMamba} & 61.82 & 85.27 & 61.76 & 81.75 & 64.42 & 71.75 & 53.05 & 67.34 & 71.44 & 76.27 & 68.88 & 69.43 \\
      DeCoF~\cite{decof} & 78.76 & 91.84 & 69.58 & 95.91 & 51.89 & 66.71 & 32.10 & 77.26 & 80.29 & 84.59 & 73.00 & 72.90 \\

      \bottomrule
      \end{tabular}%
    }
    \label{tab:closed_source_auc_benchmark}%
    \vspace{0.5cm}
    \hfill
    \centering
    \captionof{table}{ACC Performance Comparison of Vision-Language Models on Closed-Source Generative Models. This comparison uses ACC because VLMs produce binary decisions, which preclude the calculation of AUC.}
     \vspace{0.2cm}
    \resizebox{\textwidth}{!}{
      \begin{tabular}{>{\centering\arraybackslash}m{3.2cm}|*{11}{c|}c}
      \toprule
      \multirow{3}{*}{\textbf{Method}} & \multicolumn{11}{c|}{\textbf{Closed-Source Approaches}} & \multirow{3}{*}{\textbf{AVG}} \\
      \cmidrule(lr){2-12}
      & \makecell[c]{\scriptsize Luma} & \makecell[c]{\scriptsize Open\\\scriptsize Sora} & \makecell[c]{\scriptsize Sora} & \makecell[c]{\scriptsize Causvid\\\scriptsize 24fps} & \makecell[c]{\scriptsize Gen2} & \makecell[c]{\scriptsize Gen3} & \makecell[c]{\scriptsize Jimeng} & \makecell[c]{\scriptsize Kling} & \makecell[c]{\scriptsize Pika} & \makecell[c]{\scriptsize Vidu} & \makecell[c]{\scriptsize Wan} &  \\
      \midrule
      Emu3-Stage1~\cite{wang2024emu3} & 36.83 & 38.95 & 37.35 & 41.88 & 40.20 & 34.92 & 42.80 & 39.73 & 37.33 & 37.80 & 36.23 & 38.55 \\
      DeepseekVL2-S~\cite{DeepSeek-VL2} & 49.43 & 49.30 & 49.30 & 49.30 & 49.30 & 49.30 & 49.30 & 49.30 & 49.35 & 49.33 & 49.30 & 49.32 \\  
      LLaVA-v1.5-7B~\cite{liu2023improvedllava} & 50.00 & 50.00 & 50.00 & 50.00 & 50.00 & 50.00 & 50.00 & 50.00 & 50.00 & 50.00 & 50.00 & 50.00 \\   
      Deepseek-VL-7B~\cite{lu2024deepseekvl} & 50.18 & 49.98 & 50.15 & 50.05 & 49.93 & 50.05 & 49.98 & 50.10 & 50.30 & 50.10 & 50.03 & 50.08 \\
      Kimi-VL-A3B~\cite{Kimi-vl} & 51.28 & 50.60 & 51.70 & 52.18 & 49.68 & 52.75 & 49.88 & 54.15 & 52.38 & 51.92 & 51.95 & 51.68 \\
      FastVLM-7B~\cite{fastvlm} & 51.85 & 52.85 & 50.88 & 53.77 & 48.60 & 52.58 & 51.50 & 59.88 & 57.05 & 52.80 & 53.57 & 53.21 \\
      Qwen2.5-VL-7B~\cite{Qwen2.5-VL} & 54.00 & 53.35 & 51.85 & 54.25 & 49.62 & 53.10 & 51.55 & 62.90 & 58.13 & 55.38 & 54.60 & 54.43 \\
      Qwen2.5-VL-32B~\cite{Qwen2.5-VL} & 55.62 & 56.97 & 52.35 & 54.47 & 50.10 & 53.85 & 57.30 & 62.60 & 61.98 & 56.20 & 57.05 & 56.23 \\
      InternVL3-8B~\cite{zhu2025internvl3} & 58.45 & 56.75 & 51.78 & 56.85 & 47.62 & 55.77 & 54.27 & 62.98 & 67.83 & 52.42 & 55.25 & 56.36 \\
      Qwen2.5-VL-3B~\cite{Qwen2.5-VL} & 54.75 & 54.60 & 55.45 & 54.60 & 53.45 & 59.33 & 54.07 & 66.40 & 64.40 & 57.67 & 58.05 & 57.52 \\
    DeepseekVL2~\cite{DeepSeek-VL2} & 76.22 & 77.68 & 71.05 & 72.67 & 70.45 & 75.05 & 71.43 & 79.15 & 88.52 & 72.55 & 74.08 & 75.35 \\

      \bottomrule
      \end{tabular}%
    }
    \label{tab:closed_source_acc_benchmark}%
  \end{minipage}
  \hfill
  \begin{minipage}{0.3\textwidth}
    \centering
    \includegraphics[width=\textwidth]{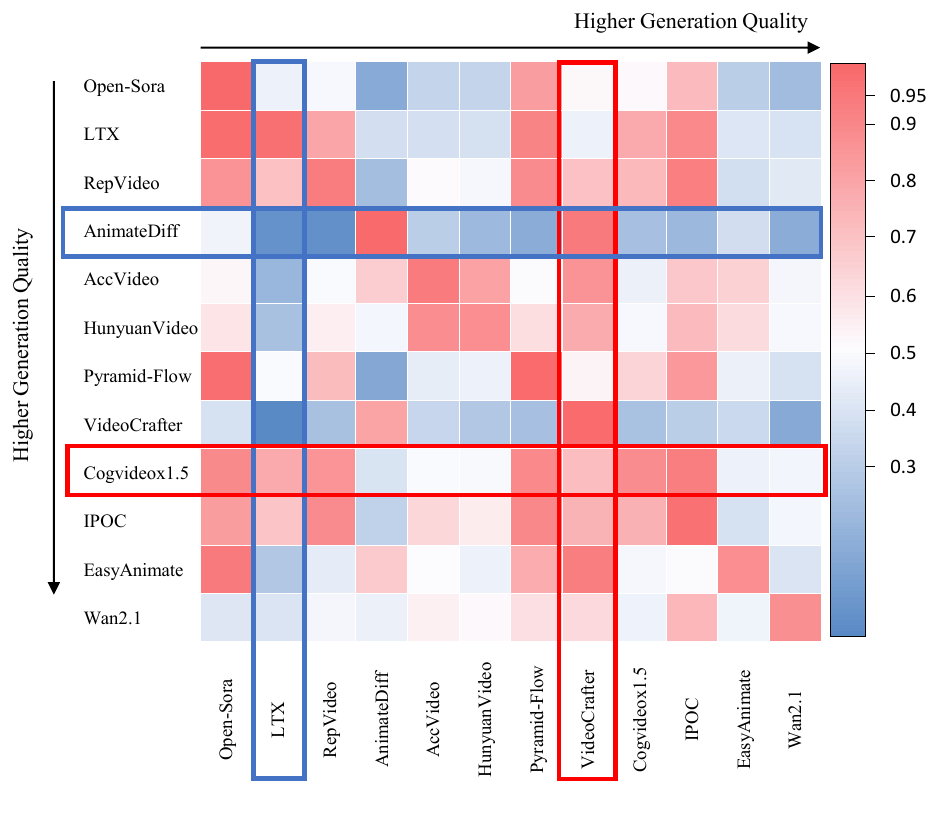}
   \scriptsize(a) DeCoF
    \includegraphics[width=\textwidth]{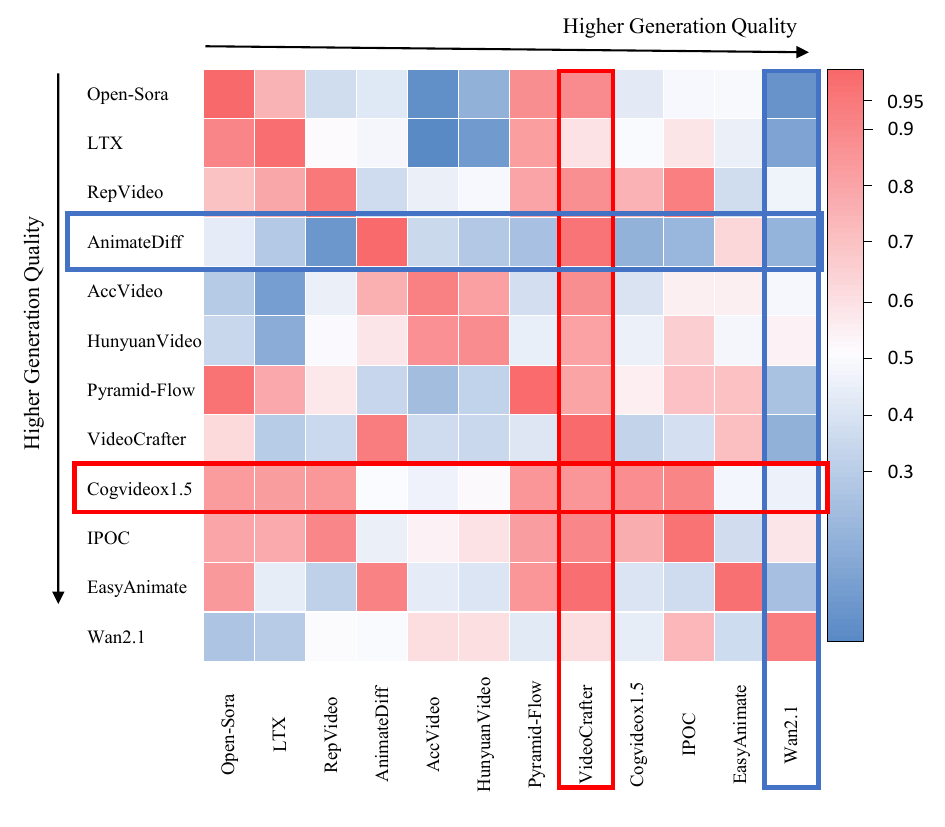}
    \scriptsize(b) UnivFD
    \includegraphics[width=\textwidth]{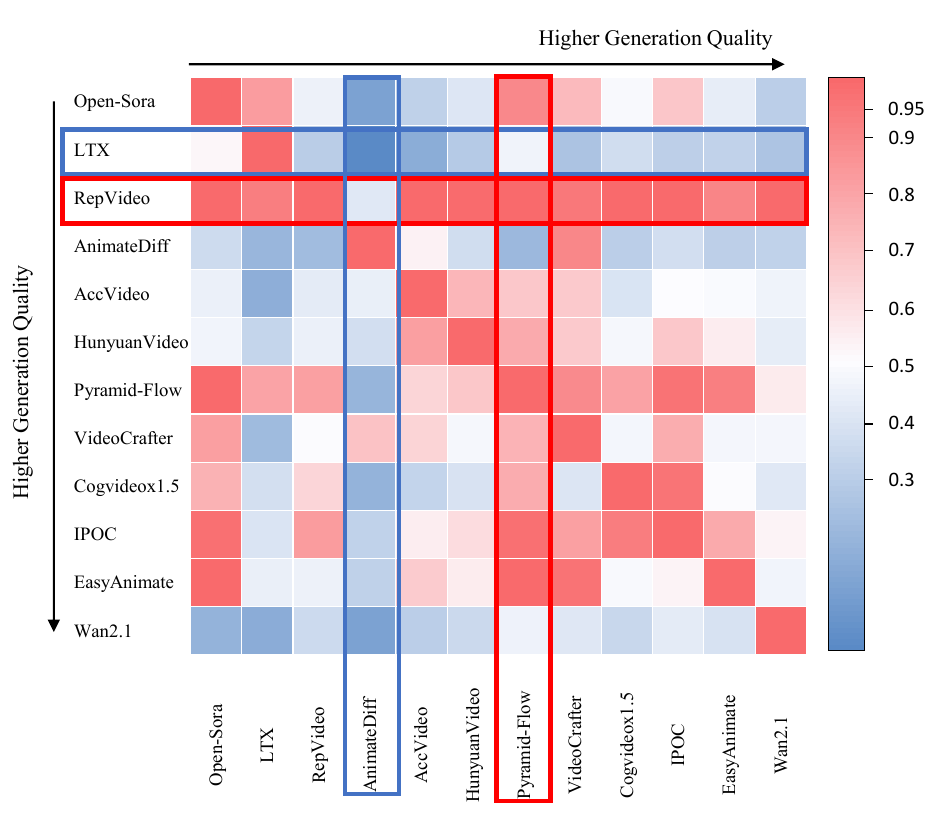}
     \scriptsize(c) I3D
     \vspace{-0.2cm}
    \caption{Each heatmap displays the AUC scores of a detector trained on one generative model (rows) and evaluated across others (columns).}
    \label{fig:heatmaps}
  \end{minipage}
\end{figure*}
\section{Evaluation and Analysis}
\subsection{Benchmark Setup}
We preprocess each video by uniformly sampling 32 frames from its first 128 frames (or all available frames if the total is fewer than 128). Each frame is center-cropped along the shorter side and resized to 256×256 resolution. To maintain format consistency, all frames are saved in PNG format. In model training and inference, only the first 8 of the 32 sampled frames are used.
Our experimental setup follows prior work on AI-generated video detection~\cite{decof,DeMamba}, in which a model trained on videos from one generator is evaluated across multiple other generators. 

We choose Open-Sora~\cite{opensorav2} as the training source given its significance as a milestone in open-source video generation. Additional details regarding the benchmark configuration are provided in the Appendix.
To systematically evaluate AI-generated video detection methods, we group existing approaches into four categories: Video Classification Models~\cite{i3d,MViTv2,slowfast,videomae,timesformer,lin2019tsm,uniformer,uniformerv2,videoswin,x3d},
Generated Image Detection Models~\cite{CNN-generated,d3,effort,ForgeLens,freqdet,Gram-Net,fusing,NPR,Universal},
Generated Video Detection Models~\cite{decof,DeMamba}, and
Multimodal Large Language Models (MLLMs)~\cite{Qwen2.5-VL,zhu2025internvl3,Kimi-vl,fastvlm,DeepSeek-VL2,lu2024deepseekvl,liu2023improvedllava}.
For each category, we select several representative models. The detailed model descriptions can be found in the Appendix.
\subsection{Findings and Analyses}

To investigate the field of AI-generated video detection, we employ AIGVDBench and the VBench evaluation framework. We specifically address the following key questions, which have not yet been systematically explored:

\vspace{0.3cm}
\noindent\textbf{Finding-1: AI-generated video detection remains fraught with challenges, making the deep investigation and optimization of these four paradigms a crucial direction for future research. }

As indicated by our evaluation results~\cref{tab:video_detection_benchmark,tab:vision_language_video_detection,tab:closed_source_auc_benchmark,tab:closed_source_acc_benchmark}, we identify four promising research directions for detecting AI-generated videos. First, adapting classic video classification networks remains highly viable. While their efficacy on closed-source models is currently limited, there is significant potential in designing specialized architectures, optimizing training processes, or creating tailored data augmentation methods for this task.
Furthermore, advanced image detection algorithms can be effectively adapted for frame-level video detection. The strong performance of Effort~\cite{effort} and ForgenLens~\cite{ForgeLens} on both open-source and closed-source models demonstrates the efficacy of their strategies to discard irrelevant features and reduce data dependency. Meanwhile, the significant performance degradation observed in other image detection algorithms on closed-source models provides indirect support for the spatial-temporal artifact hypothesis proposed in DeCoF~\cite{decof}. 

Third, although  DeCoF and Demamba have not achieved state-of-the-art performance on current benchmarks, their pivotal contribution lies in the core idea of inhibiting spatial artifacts while leveraging temporal artifacts for detection. We speculate that their performance limitations may stem from constraints in the scale and quality of previous available data. We anticipate that the dataset constructed in this study will facilitate advancements in this direction, promoting the development of more sophisticated video detection algorithms.

Finally, Vision-Language Models (VLMs) currently lag behind traditional models on this task. A key limitation is their output of only labels rather than probabilities, complicating accuracy assessment. The excellent performance of Deepseek-VL-2 on closed source models is analyzed further in the ~\cref{FQ1} of Appendix. Despite this, VLMs remain a highly promising direction for boosting both interpretability and accuracy.

In summary, all four detection approaches have room for improvement. The most effective strategy guides models to ignore irrelevant variables and focus on key real-vs-generated distinctions. Other viable paths include controlling training data volume and leveraging temporal artifacts. For the long term, developing interpretable algorithms with VLMs is the most promising direction, followed by enhancing traditional video classification models. 

Building upon \textbf{Finding-1}, the subsequent exploration is organized into three distinct dimensions: further analyses, additional findings, and broader discussions. Details are available in \cref{FQ1} of the Appendix.
\begin{itemize}
     \item \textbf{Analysis-1.1:}  An analysis on the heightened difficulty of detecting closed-source models relative to open-source ones. 

    \item  \textbf{Analysis-1.2:} Further analysis on Effort and ForgeLens.

     \item  \textbf{Finding-1.1:} The type of generation task has a significant impact on detector performance, and the extent of this impact varies considerably across different types of detectors.

    \item \textbf{Finding-1.2:} Current VLMs lack reliable capability for detecting AI-generated videos.
    \item \textbf{Analysis-1.3:} An exploration of whether the performance superiority of DeepSeek-VL signifies a genuine ability to discern authenticity.
\end{itemize}

\vspace{0.2cm}
\noindent\textbf{Finding-2: Improvements in video generation model quality do not ensure reduced detectability or better detector generalization.} 
\vspace{0.2cm}
To evaluate the impact of generation model on detection robustness, we established a comprehensive evaluation framework. Three benchmark detectors (I3D, DeCoF, UnivFD) were each trained on videos generated by one specific model and then tested on outputs from all other models. This cross-detection matrix was designed to assess generalization performance beyond the training domain. To isolate the variable of model architecture, all generation models were operated under a text-to-video (T2V) paradigm.  To precisely characterize the generalization dynamics, we define four key concepts. From the training perspective, the ``Optimal generation model" enables the best generalization performance, while the ``Least favorable generation model" yields the poorest generalization. From the evaluation perspective, the ``Best generation model" is the most challenging to detect (lowest performance), and the ``Worst generation model" is the easiest to detect (highest performance). In Figure~\ref{fig:heatmaps}, red boxes mark the optimal (rows) and worst (columns) models; blue boxes indicate the least favorable (rows) and best (columns) models.

Consistent with the Vbench, we ranked the generation models on the text-to-video (T2V) task. The results indicate that Wan 2.1 achieved the best overall performance, while Open-Sora ranked comparatively lower. The complete ranking is provided in \cref{tab:quality_comparison} in the Appendix.
Our experiments further revealed that although the same detector exhibited significant performance variation when trained on different generation models, its performance was not directly positively correlated with the quality of the generation models. Similarly, the converse was not true: lower-quality generation models did not invariably lead to inferior detector performance. Moreover, the "Optimal generation model" differed across detectors. For instance, RepVideo was the optimal generation model for the I3D detector, but not for DeCoF and UnivFD . For a given detector, higher model quality did not consistently make detection more difficult, and its "Worst generation model" also varied by detector type. For example, VideoCrafter was the worst  generation model for UnivFD and DeCoF, but not for I3D.

In summary, although detector performance generally tends to decrease when confronted with "higher-quality" generation models, the definition of "higher quality" should not be based solely on the generation capability of the models themselves. For a detector, higher-quality training data can indeed improve its performance; however, the standard for what constitutes "high-quality" training data must be assessed in conjunction with the specific characteristics of the detector and cannot be simplistically equated with the objective performance of the generation model.

Building upon \textbf{Finding-2}, the subsequent exploration is organized into three distinct dimensions: further analyses and  findings. Details are available in \cref{FQ2} of the Appendix.
\begin{itemize}
    \item \textbf{Analysis-2.1}: Analysis of variations in the detectability of generative models.
    \item  \textbf{Analysis-2.2}:  Key characteristics of generative models for producing high-quality training data and their relationship to detector type.

    \item \textbf{Analysis-2.3}: The effects of sampling steps  on detector performance.
    \item \textbf{Analysis-2.4}: The effects of guidance scale on detector performance.
    \item \textbf{Analysis-2.5}: The impact of video type on detection difficulty.
\end{itemize}

\section{Conclusion}
In this work, we have systematically addressed the core challenges in the field of AI-generated video detection by introducing a standardized dataset construction pipeline and establishing a large-scale, high-quality benchmark dataset, thereby laying a solid foundation for this important area of research. The proposed attribute balancing algorithm and model selection strategy ensure the diversity and representativeness of the dataset, which covers 31 generative models and over 440,000 videos, making it the highest-quality benchmark resource currently available in the field of generated video detection. 

Comprehensive experimental evaluation reveals the performance of four categories of detection methods on this task, revealing previously unexplored yet critical research questions. We hope that this work will promote the rapid iteration of generated content detection technologies, thereby collectively advancing the development of trustworthy digital media authentication and contributing to the construction of a secure digital ecosystem.
{
    \small
    \bibliographystyle{ieeenat_fullname}
    \bibliography{main}
}
\let\twocolumn\oldtwocolumn
\clearpage
\maketitlesupplementary
\definecolor{lightcoral}{RGB}{240,128,128}
\definecolor{lightblue}{RGB}{173,216,230}
\section{Content Structure in Appendix}
We have organized additional important content in the Appendix due to the limited space. We present a brief outline of the content structure of the Appendix to facilitate readers to find the corresponding content, as follows:

\begin{itemize}[leftmargin=10pt] 
    \item \textbf{~\cref{6}: Further Findings and Analysis}
    \begin{itemize}
        \item  ~\cref{6.1}: Definition of Evaluation Metrics;
        \item ~\cref{FQ1}:  Further Findings and Analyses of Finding-1;
        \item ~\cref{FQ2}: Further Findings and Analyses of Finding-2;
    \end{itemize}
    \item \textbf{\cref{7}: Details of AIGVDBench:}
    \begin{itemize}
        \item  ~\cref{7.1}: Brief Introduction of Generation Models;
        \item  ~\cref{7.2}: Introduction of the Used Detection Methods; 
        \item  ~\cref{7.3}: Introduction of Original Data;
        \item  ~\cref{7.4}: Details of Prompts Categorization;
        \item  ~\cref{7.5}: Balance Comparison with Prior Datasets;
        \item  ~\cref{7.6}: Discussion on Video Encoding Methods;
        \item  ~\cref{7.7}: Training details;
        \item  ~\cref{7.8}: Timeline of AIGVDBench Construction;
        \
    \end{itemize}
    \item \textbf{~\cref{8}: Limitation;}
        

\end{itemize}

\section{Further Findings and Analysis}
\label{6}
\subsection{Definition of Evaluation Metrics.}
\label{6.1}
The VBench benchmark suite systematically evaluates the performance of video generation models across six core dimensions, with each metric quantifying a distinct key characteristic of the generated content. The metrics are defined and explained below:\\
\begin{itemize}

 \item \textbf{Aesthetic Quality}: This metric assesses the artistic value and visual appeal of individual video frames, encompassing compositional harmony, colour richness and coordination, photographic realism, naturalness, and overall artistic merit. It reflects human perceptual standards for visual content, analogous to criteria used in evaluating professional photography.\\
 \item \textbf{Background Consistency}: This measures the temporal coherence of the background scene throughout a video sequence. It focuses on the stability and consistency of non-subject environmental elements (e.g., landscapes, buildings) across frames, aiming to detect illogical mutations or structural distortions in the background.\\
 \item \textbf{Dynamic Degree}: This quantifies the intensity and magnitude of motion within a video. It evaluates whether the generated content exhibits significant motion characteristics (e.g., rapidly moving objects, camera movement) appropriate to the scene context, thereby distinguishing dynamic content from static imagery and reflecting the level of visual vitality.\\
 \item \textbf{Imaging Quality}: This metric detects low-level image attributes and technical artefacts in video frames. It primarily evaluates distortions such as over-exposure, image noise, motion blur, and colour inaccuracies, focusing on objective signal-level quality rather than subjective aesthetic judgement.\\
 \item \textbf{Motion Smoothness}: This analyses the fluency and naturalness of motion in the video. It examines whether motion trajectories adhere to physical principles (e.g., uniform or accelerated motion), with a focus on identifying non-continuous phenomena such as jerky transitions, jitter, or temporal inconsistencies.\\
 \item \textbf{Subject Consistency}: This evaluates the temporal coherence of the appearance of the main subject (e.g., a person, animal, or vehicle) in the video. It requires that the subject's identity and visual attributes remain consistent across frames, preventing implausible changes in colour, shape, or texture.
   
\end{itemize}

\subsection{Further Findings and Analyses of Findings-1}
\label{FQ1}
\vspace{0.1cm}
\textbf{Analysis-1.1: An analysis on the heightened difficulty of detecting closed-source models relative to open-source ones.} 
\vspace{0.1cm}

\begin{figure*}[tbp]
    \centering
    
    \begin{subfigure}[t]{0.28\textwidth}
        \centering
        \includegraphics[width=\linewidth]{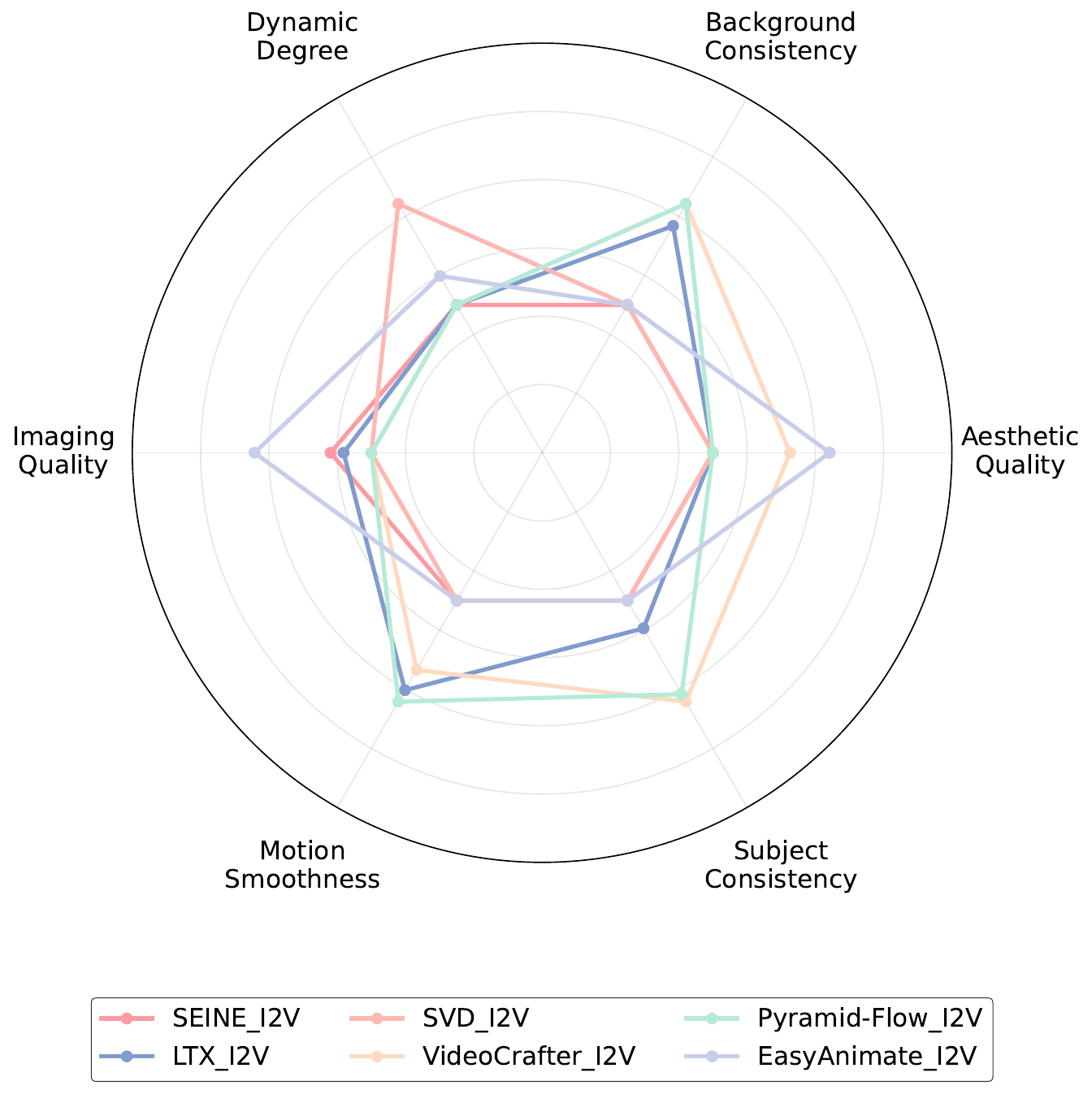}
        \caption{Comparison of image to video models.}
        \label{fig:sub1}
    \end{subfigure}
    \hfill
    \begin{subfigure}[t]{0.38\textwidth}
        \centering
        \includegraphics[width=\linewidth]{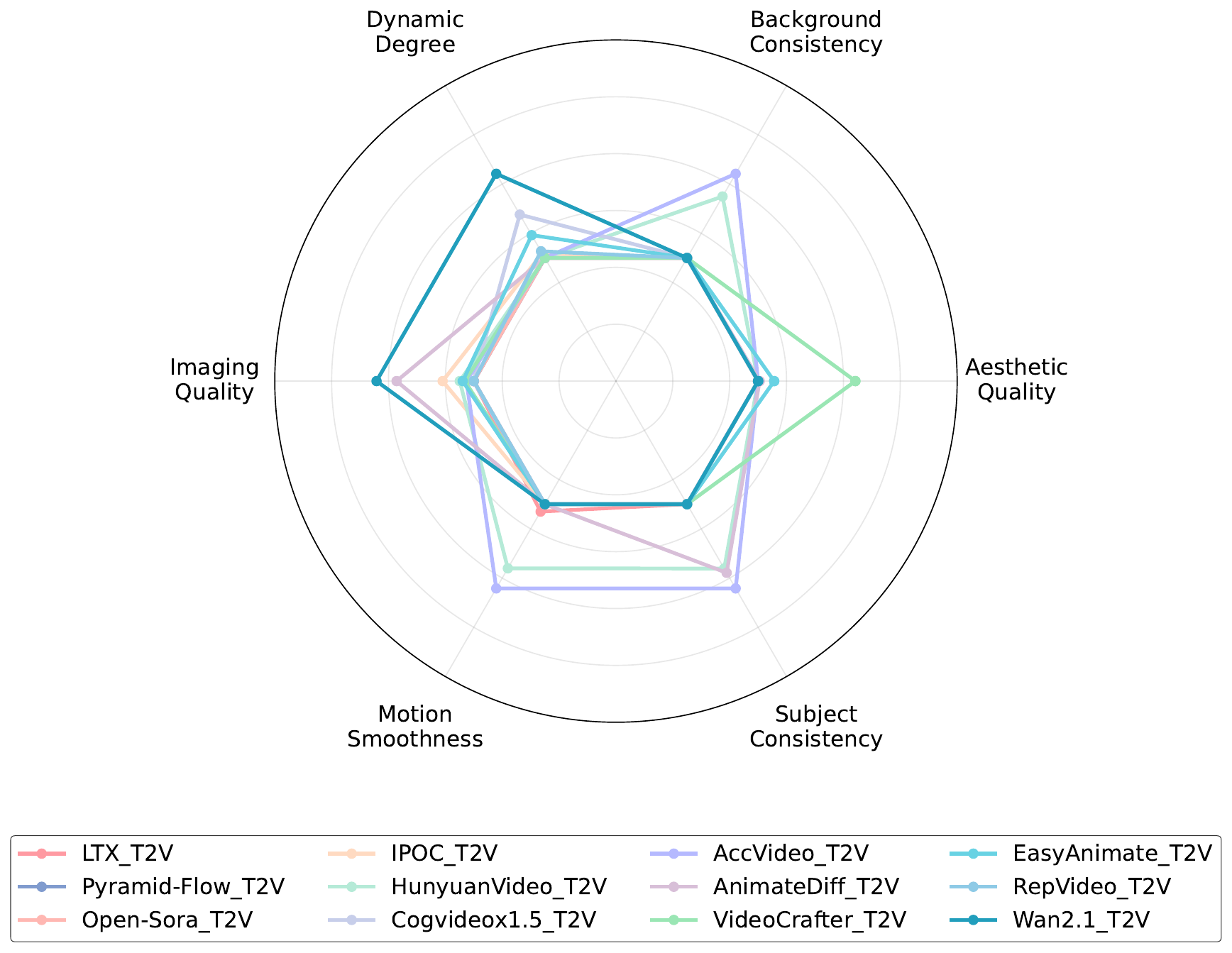}
        \caption{Comparison of text to video models.}
        \label{fig:sub2}
    \end{subfigure}
    \hfill
    \begin{subfigure}[t]{0.28\textwidth}
        \centering
        \includegraphics[width=\linewidth]{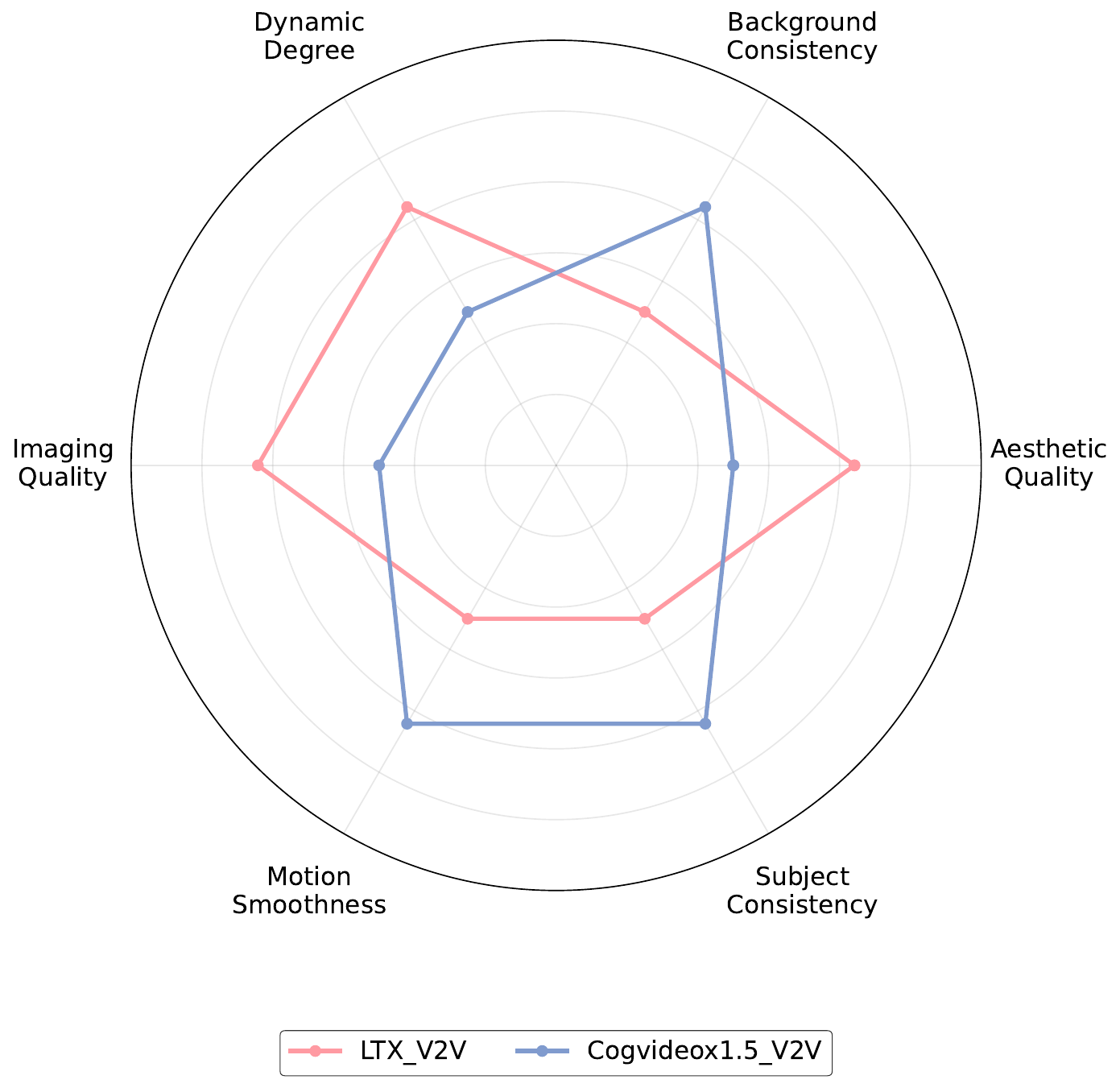}
        \caption{Comparison of video to video models.}
        \label{fig:sub3}
    \end{subfigure}
    
    \vspace{0.8cm}
    
    \begin{subfigure}[t]{0.48\textwidth}
        \centering
        \includegraphics[width=0.7\linewidth]{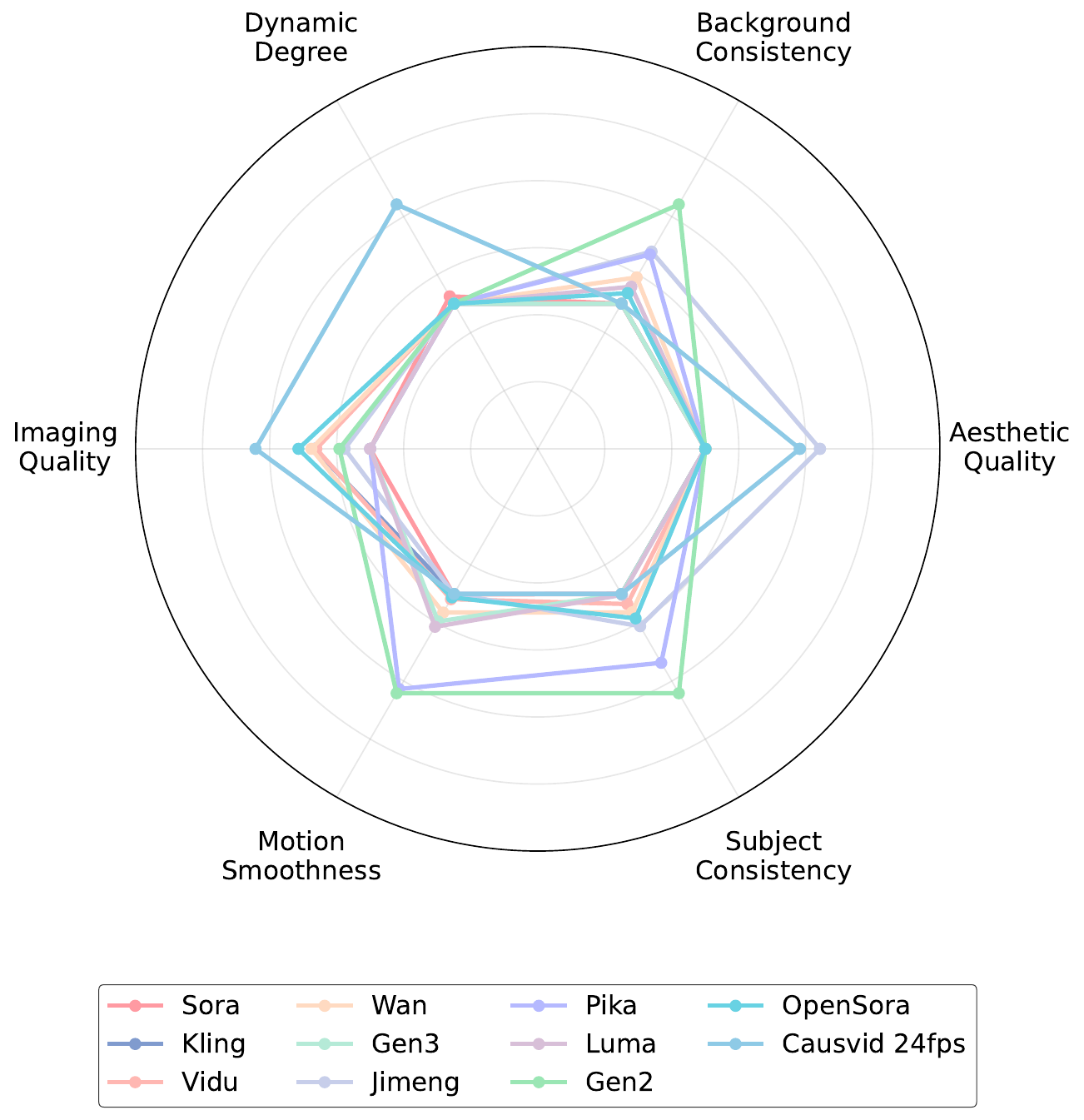}
        \caption{Comparison of closed source models.}
        \label{fig:sub4}
    \end{subfigure}
    \hfill
    \begin{subfigure}[t]{0.48\textwidth}
        \centering
        \includegraphics[width=0.7\linewidth]{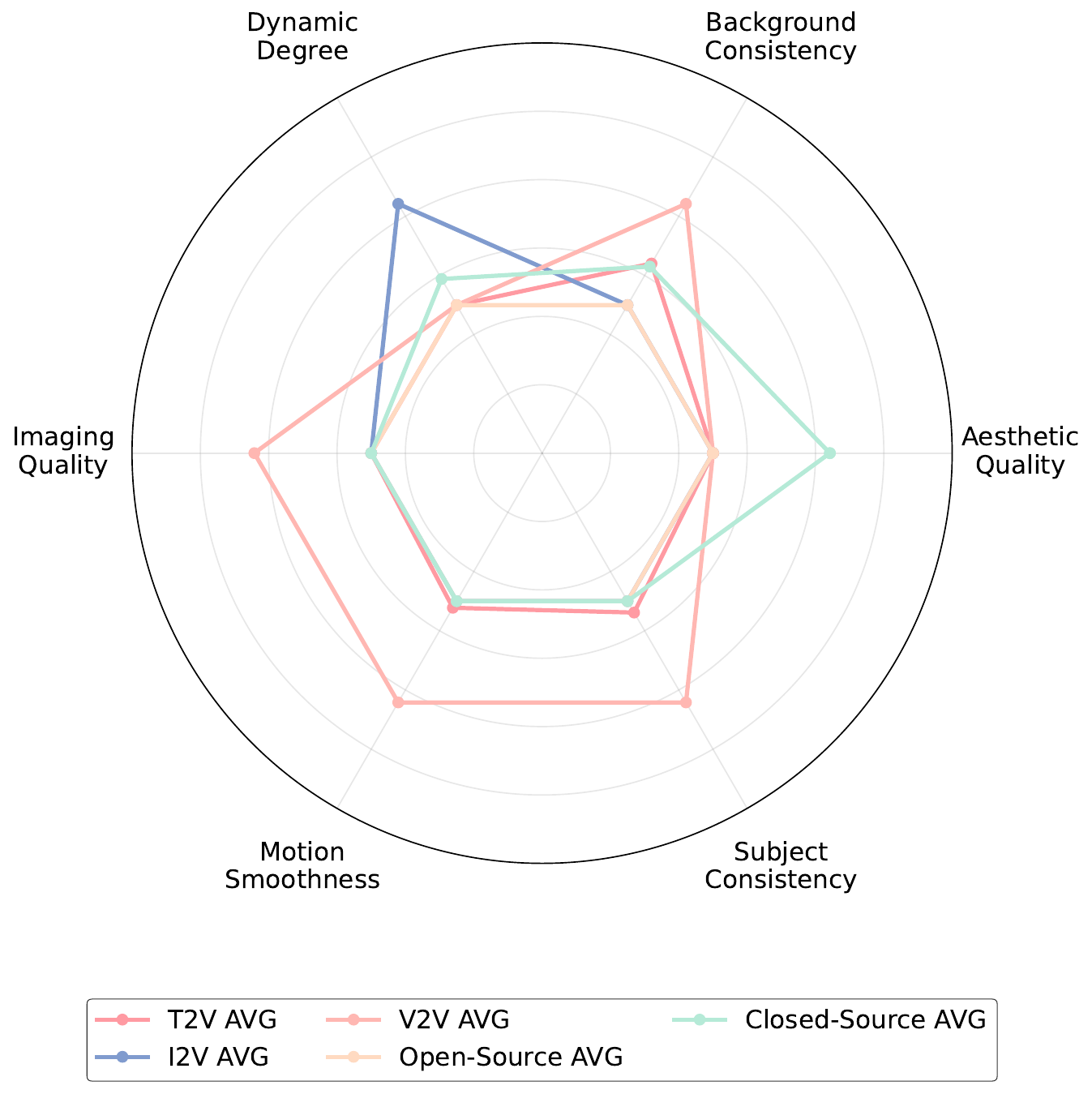}
        \caption{Comparison of different model sources and generation tasks.}
        \label{fig:sub5}
    \end{subfigure}
    
    \caption{Comparison of video generation models on Aesthetic Quality, Background Consistency, Dynamic Degree, Imaging Quality, Motion Smoothness, Subject Consistency, Frame-Level (Image \& Aesthetic Average) and Final Score based on Vbench.~\cite{vbench}}
    \label{fig:radar_comprasion}
\end{figure*}

1. A fundamental contrast exists between open and closed source models, which is clearly reflected in their output quality. Our evaluation based on the VBench framework across six dimensions indicates that closed source models generally achieve significantly higher overall video quality than open source models (~\cref{fig:radar_comprasion,tab:quality_comparison}). This performance disparity aligns with the inherent differences between the two: open source models boast full transparency in design and parameters, often with a clear lineage of iterative refinement, whereas closed source models are opaque, not only in terms of hidden hyperparameters but also regarding potential post processing techniques. Moreover, their commercial orientation typically leads to larger scale designs with vast parameter counts, ultimately contributing to their superior generative performance.

\begin{table*}[t]
\centering
\vspace{0.5cm}
\caption{Comparison of video generation models on Aesthetic Quality, Background Consistency, Dynamic Degree, Imaging Quality, Motion Smoothness, Subject Consistency, Frame-Level (Image \& Aesthetic Average) and Final Score based on Vbench.~\cite{vbench}}
\resizebox{\textwidth}{!}{
\begin{tabular}{ccccccccc}
\toprule
Model Name & \makecell{Aesthetic\\Quality} & \makecell{Imaging\\Quality} & \makecell{Frame-\\Level} & \makecell{Background\\Consistency} & \makecell{Dynamic\\Degree} & \makecell{Motion\\Smoothness} & \makecell{Subject\\Consistency} & \makecell{Final\\Score} \\
\midrule
\multicolumn{9}{c}{\textbf{Open-Source T2V Models}} \\
\midrule
Open-Sora & 54.46 & 53.68 & 54.07 & 97.24 & 7.17 & 95.80 & 97.01 & 73.70 \\
LTX & 54.11 & 58.60 & 56.36 & 97.11 & 22.53 & 96.90 & 96.66 & 77.44 \\
RepVideo & 55.08 & 63.20 & 59.14 & 96.32 & 23.63 & 93.07 & 95.50 & 77.60 \\
AnimateDiff & 60.95 & 67.07 & 64.01 & 97.87 & 5.70 & 96.36 & 98.93 & 77.61 \\
AccVideo & 58.37 & 63.72 & 61.05 & 99.17 & 9.10 & 98.89 & 99.15 & 77.89 \\
HunyuanVideo & 59.31 & 64.04 & 61.68 & 98.92 & 9.67 & 98.37 & 98.87 & 78.03 \\
Pyramid-Flow & 58.32 & 63.16 & 60.74 & 97.16 & 18.58 & 96.23 & 96.86 & 78.24 \\
VideoCrafter & 64.17 & 63.80 & 63.99 & 97.88 & 11.27 & 95.35 & 97.95 & 78.26 \\
Cogvideox1.5 & 54.98 & 63.39 & 59.19 & 96.91 & 27.25 & 94.80 & 96.74 & 78.92 \\
IPOC & 56.14 & 64.86 & 60.50 & 97.39 & 23.43 & 94.86 & 97.88 & 79.01 \\
EasyAnimate & 61.44 & 63.90 & 62.67 & 97.12 & 25.22 & 93.09 & 97.28 & 79.64 \\
Wan2.1 & 59.52 & 68.03 & 63.78 & 97.14 & 31.28 & 92.18 & 96.62 & 80.87 \\
\midrule
\multicolumn{9}{c}{\textbf{Open-Source I2V Models}} \\
\midrule
SVD & 53.52 & 59.57 & 56.55 & 92.34 & 36.52 & 81.10 & 92.22 & 75.50 \\
SEINE & 53.86 & 64.98 & 59.42 & 92.14 & 29.13 & 89.03 & 92.14 & 76.60 \\
LTX & 55.30 & 64.73 & 60.02 & 96.71 & 17.23 & 95.46 & 95.97 & 77.35 \\
VideoCrafter & 57.81 & 64.00 & 60.91 & 97.06 & 15.50 & 94.49 & 97.15 & 77.46 \\
Pyramid-Flow & 56.09 & 64.09 & 60.09 & 97.06 & 20.57 & 96.01 & 97.03 & 78.34 \\
EasyAnimate & 58.34 & 66.43 & 62.39 & 94.85 & 31.65 & 90.02 & 94.04 & 79.15 \\
\midrule
\multicolumn{9}{c}{\textbf{Open-Source V2V Models}} \\
\midrule
Cogvideox1.5 & 55.29 & 66.24 & 60.77 & 98.19 & 7.65 & 97.73 & 98.79 & 77.07 \\
LTX & 56.91 & 67.58 & 62.25 & 98.03 & 9.78 & 97.14 & 98.30 & 77.77 \\
\midrule
\multicolumn{9}{c}{\textbf{Closed-Source Models}} \\
\midrule
Pika & 58.83 & 58.52 & 58.68 & 98.29 & 10.38 & 98.31 & 97.94 & 76.77 \\
Gen3 & 58.30 & 60.19 & 59.25 & 94.72 & 23.50 & 95.27 & 91.62 & 77.02 \\

Gen2 & 63.04 & 64.10 & 63.57 & 98.99 & 3.68 & 98.50 & 98.64 & 77.63 \\
Vidu & 60.67 & 64.58 & 62.63 & 97.00 & 19.88 & 94.27 & 96.59 & 78.72 \\
OpenSora & 62.99 & 64.95 & 63.97 & 97.75 & 18.20 & 94.19 & 96.92 & 79.09 \\
Wan & 60.65 & 64.68 & 62.67 & 97.97 & 20.88 & 94.87 & 96.78 & 79.24 \\
Kling & 62.76 & 64.62 & 63.69 & 97.28 & 22.38 & 94.02 & 96.28 & 79.52 \\
Luma & 62.97 & 63.33 & 63.15 & 97.84 & 23.10 & 95.51 & 96.38 & 79.84 \\
Sora & 60.57 & 63.38 & 61.98 & 97.05 & 29.25 & 94.06 & 95.45 & 79.96 \\
Jimeng & 68.29 & 63.99 & 66.14 & 98.33 & 21.15 & 92.67 & 97.10 & 80.28 \\
Causvid 24fps & 67.72 & 65.84 & 66.78 & 97.18 & 40.25 & 84.76 & 95.68 & 82.08 \\
\midrule
\textbf{I2V AVG}&55.82&	\underline{63.97}&	59.90&	95.03&	\underline{25.10}&	91.02&	94.76&	77.40\\
\textbf{V2V AVG}&\underline{56.10}&	\textbf{66.91}&	\textbf{61.51}&	\textbf{98.11}&	8.72&	\textbf{97.44}&	\textbf{98.54}&	\underline{77.42}\\
\textbf{T2V AVG}&\textbf{58.07}&63.12&\underline{60.60}&\underline{97.52}&\underline{17.90}&\underline{95.49}&\underline{97.45}&	\textbf{78.10}\\

\midrule
\textbf{Open-Source AVG} & \underline{57.20} & \textbf{63.75} & \underline{60.48} & \underline{96.83} & \underline{19.14} & \textbf{94.34} & \textbf{96.75} & \underline{77.82} \\
\textbf{Closed-Source AVG} & \textbf{62.44} & \underline{63.47} & \textbf{62.96} & \textbf{97.49} & \textbf{21.15} & \underline{94.22} & \underline{96.31} & \textbf{79.10} \\
\bottomrule
\end{tabular}}
\label{tab:quality_comparison}
\end{table*}

\begin{figure}[t]
  \centering
\includegraphics[width=\linewidth]{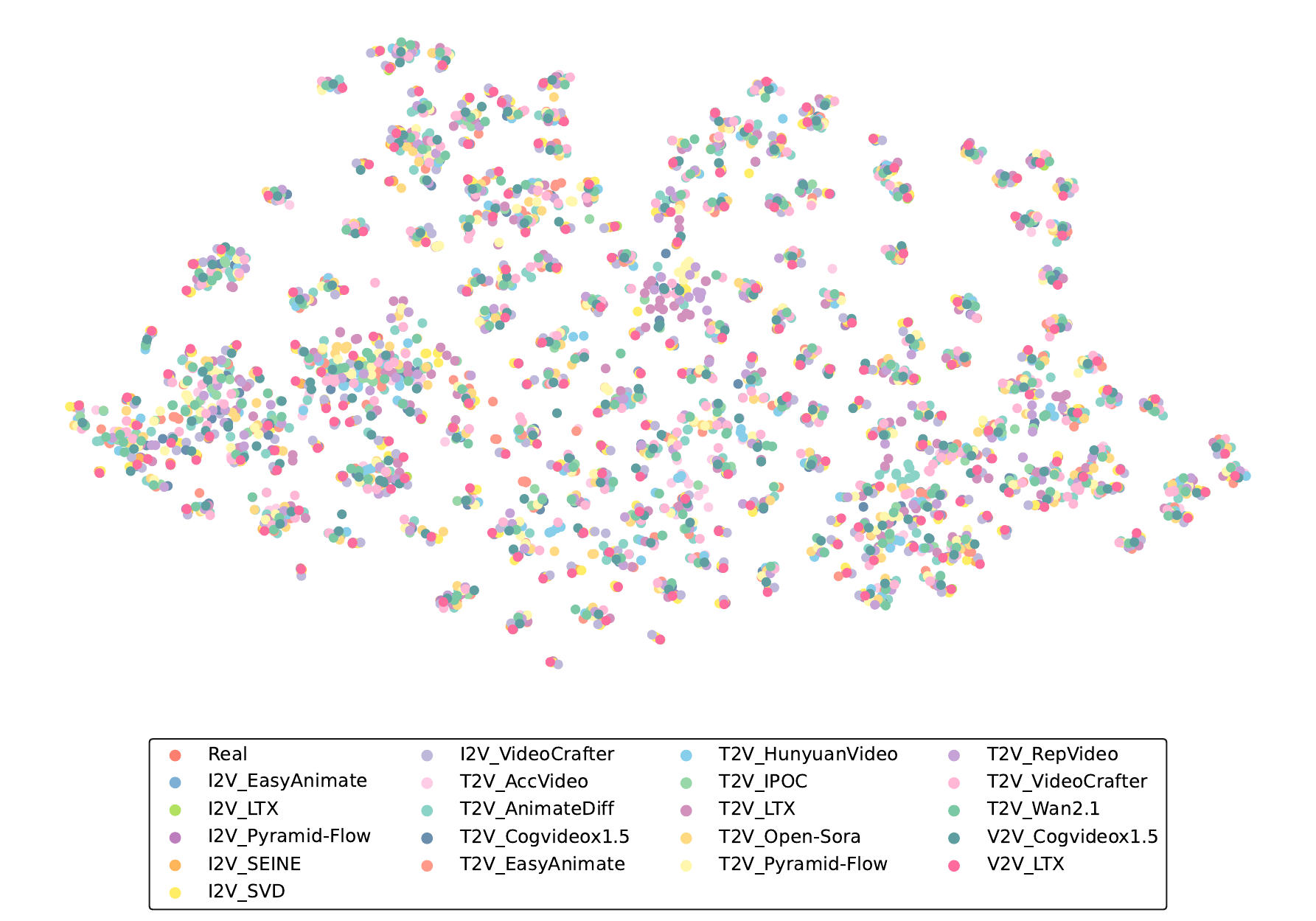}
   \caption{t-SNE~\cite{maaten2008visualizing} visualization of feature distribution for videos generated by open source models on CLIP-ViT-L/14 embeddings.}
   \label{fig:open_source_tsne}
\end{figure}

2. A critical difference also lies in the data itself. Our open source dataset was constructed with strict controls to guarantee homogeneity in semantic content. This is evidenced by CLIP-ViT-L/14 feature projections (~\cref{fig:open_source_tsne}), where videos form tight clusters containing outputs from all generative models. No such constraints apply to the closed source data, which is a heterogeneous collection of various tasks (primarily text to video). Thus, from a detection standpoint, the open source data constitutes a single domain benchmark, while the closed source data presents a far more challenging cross domain scenario, leading to a substantial increase in detection difficulty.  

\begin{figure}[t]
  \centering
\includegraphics[width=\linewidth]{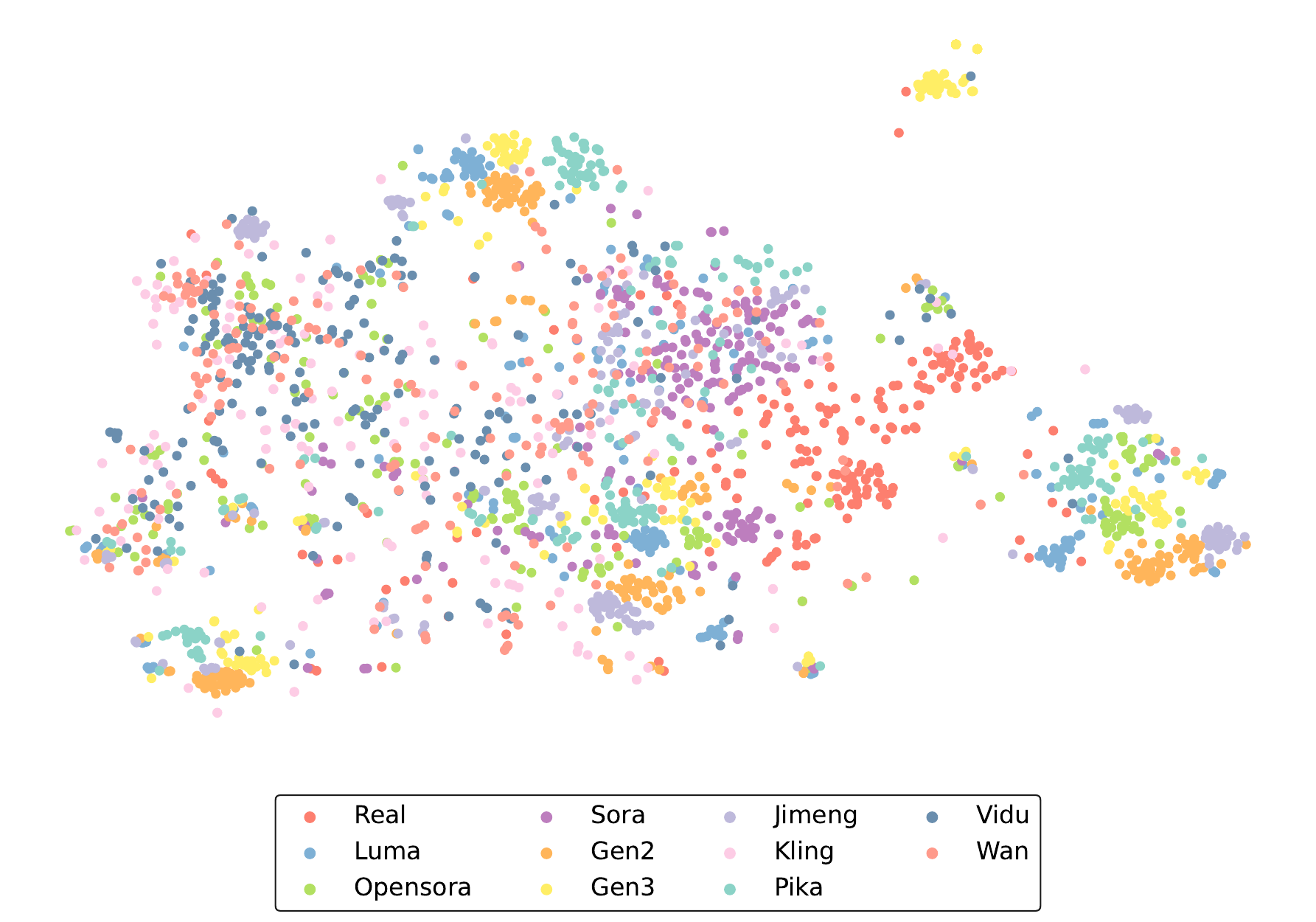}
   \caption{t-SNE~\cite{maaten2008visualizing} visualization of feature distribution for videos generated by closed source models on CLIP-ViT-L/14 embeddings.}
   \label{fig:close_source_tsne}
\end{figure}

A noteworthy and interesting phenomenon observed in \cref{fig:close_source_tsne} is that, under this open set setting, the features of closed source models do not cluster by semantic content as seen in open source models, but instead exhibit a model wise hierarchical clustering pattern, which aligns with the findings reported in~\cite{Universal}. This observation leads us to hypothesize that when real and fake videos in a dataset lack content alignment, it may induce such model driven clustering behavior. This type of pattern could potentially simplify the detection task, thereby explaining the anomalous performance of models on such datasets.

\begin{table}[t]
\centering
\caption{Performance of TimeSformer, UniformerV2, and Effort Across Video Generation Tasks and Model Sources}
\label{tab:performance of there models}
\resizebox{\linewidth}{!}{
\begin{tabular}{cccccc}
\toprule
\multirow{2}{*}{Model} & \multirow{2}{*}{Close Source} & \multicolumn{4}{c}{Open Source} \\
 \cmidrule(lr){3-6}
&  & I2V & T2V & V2V & AVG \\
\midrule
TimeSformer~\cite{timesformer} & 86.50 & 75.46 & 86.84 & 67.59 & 81.50 \\
UniformerV2~\cite{uniformerv2} & 87.07 & 81.27 & 85.94 & 73.30 & 83.27 \\
Effort~\cite{effort} & 94.05 & 86.32 & 90.20 & 74.74 & 87.49 \\
\bottomrule
\end{tabular}}
\end{table}
\textbf{Abnormal Phenomenon.} A notable exception to this performance degradation is observed in models such as TimeSformer~\cite{timesformer}, UniformerV2~\cite{uniformerv2}, and Effort~\cite{effort}, which maintain stable performance on closed source data. To further investigate this anomalous behavior, we compared the performance of these three detection models across various generative models, with a particular focus on their results on different tasks within the open source models. As shown in the ~\cref{tab:performance of there models}, all three models perform significantly better on Text to Video (T2V) tasks than on Image to Video (I2V) or Video to Video (V2V) tasks, a trend that aligns with their overall performance on closed source models. This outcome is expected, since closed source video content is mainly generated through T2V, a format favored by users for its convenience, and also because the VBench benchmark primarily uses T2V based evaluation tasks.
Furthermore, although closed source models generally achieve higher overall quality scores than open source models, their performance on certain dimensions, such as imaging quality, motion smoothness, and subject consistency, is comparable to that of open source T2V models. This indicates that the strong T2V performance of the three detection models is the main reason for their anomalous stability on closed source data. This conclusion is further supported by their notably high detection accuracy observed in T2V models.
This analysis leads to a new question: is there an underlying pattern that determines how different detection models perform across different generation tasks? 
\vspace{0.2cm}

\begin{figure}[t]
  \centering
   \includegraphics[width=\linewidth]{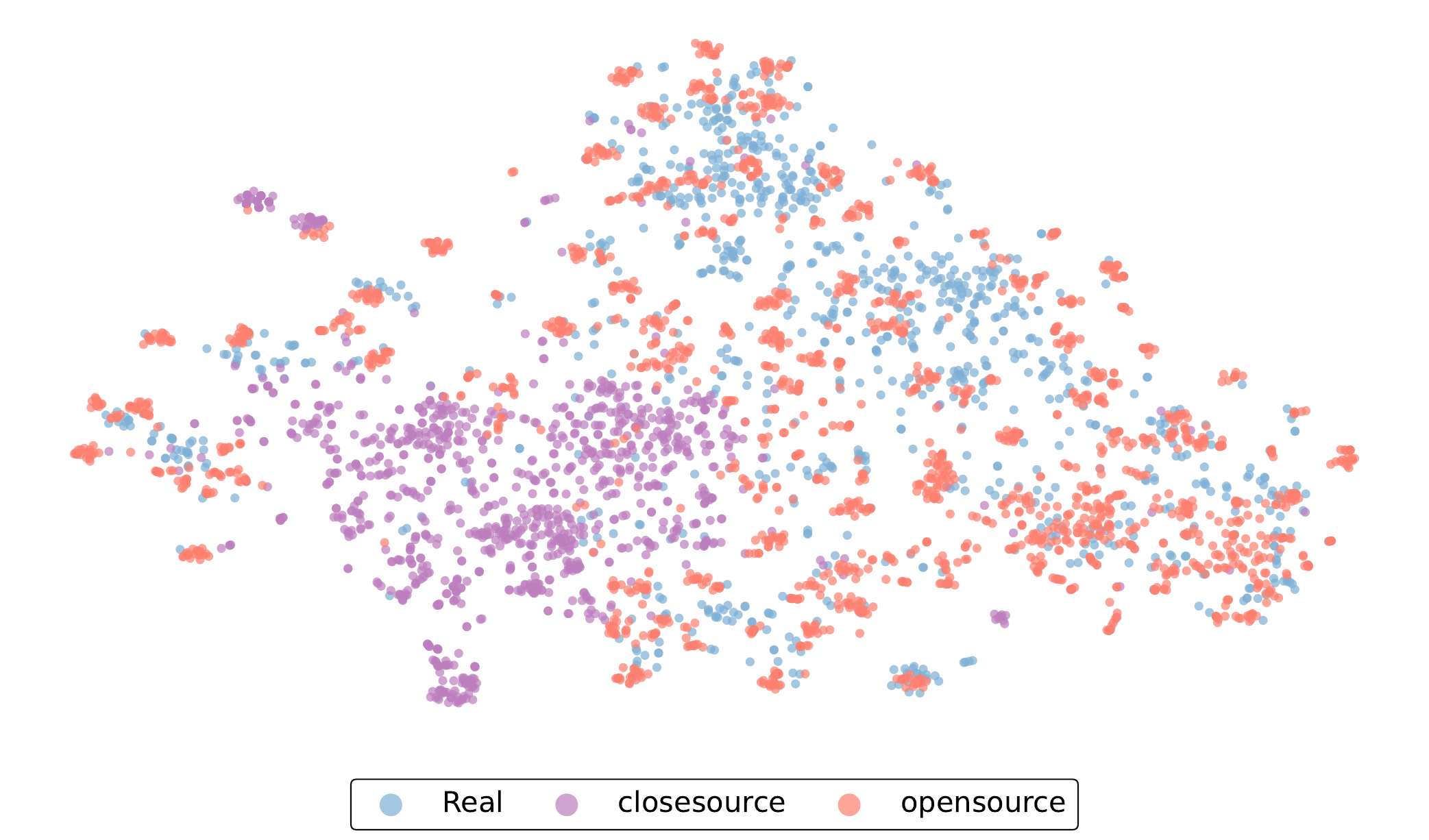}
   \caption{t-SNE~\cite{maaten2008visualizing} visualization of feature distribution from real, open-source models generated, and closed-source models generated videos on UnivFD.}
   \label{fig:tsen_uni}
\end{figure}

\begin{figure}[t]
  \centering
   \includegraphics[width=\linewidth]{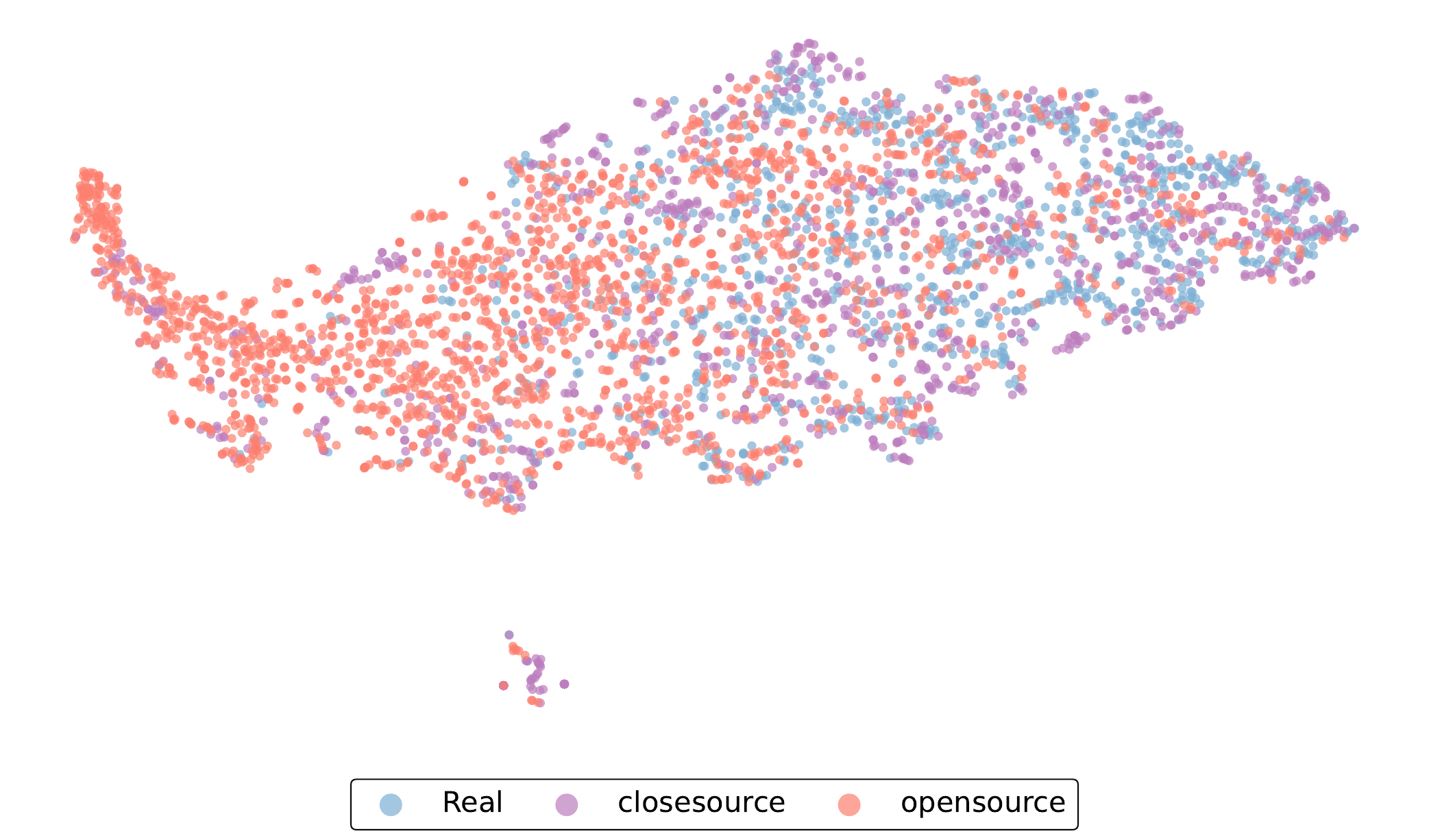}
   \caption{t-SNE~\cite{maaten2008visualizing} visualization of feature distribution from real, open-source models generated, and closed-source  models generated videos on CNNSpot. }
   \label{fig:tsen_resnet}
\end{figure}

\begin{figure}[t]
  \centering
   \includegraphics[width=\linewidth]{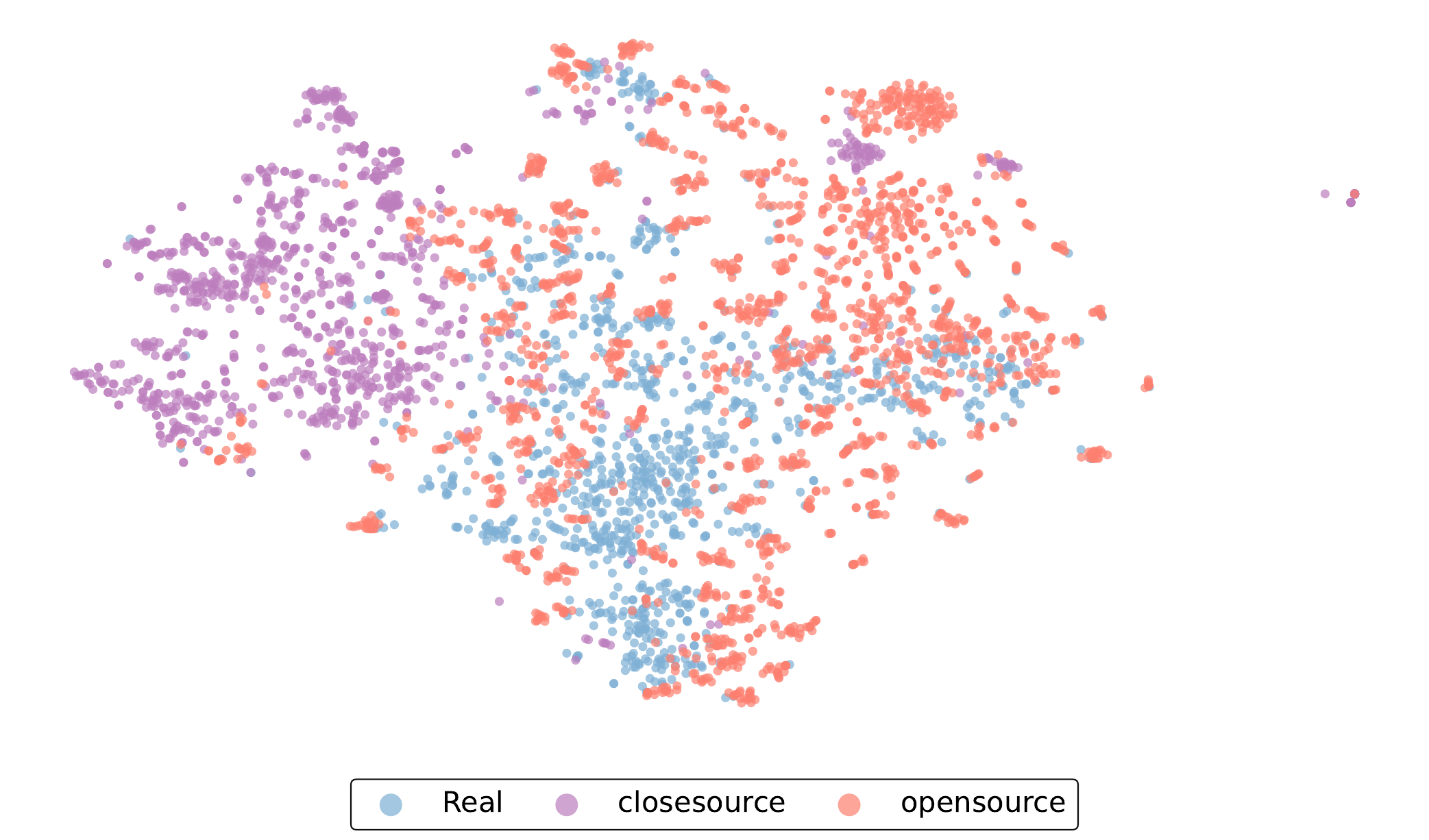}
   \caption{t-SNE~\cite{maaten2008visualizing} visualization of feature distribution from real, open-source generated, and closed-source generated videos on Effort.}
   \label{fig:tsen_effrot}
\end{figure}

\noindent\textbf{Analysis-1.2: Further analysis on Effort and ForgeLens.}

\cref{tab:video_detection_benchmark,tab:closed_source_auc_benchmark} demonstrates the potential of advanced generated image detection algorithms for frame level detection in generated videos, with the Effort~\cite{effort} and ForgeLens~\cite{ForgeLens} models exhibiting particularly outstanding performance. A noticeable trend in generated image detection is the transition of backbone networks from architectures such as ResNet to CLIP-ViT-L/14 following the introduction of UnivFD~\cite{Universal}.

To investigate the rationale behind this shift, we conducted a feature visualization analysis comparing the off the shelf CLIP-ViT-L/14 without additional training and a ResNet model trained on the Open-Sora dataset. The analysis included features from real videos and those generated by both open source and close source models. As shown in \cref{fig:tsen_resnet,fig:tsen_uni} and \cref{tab:video_detection_benchmark}, the untrained CLIP-ViT-L/14 exhibits a feature discrimination capability comparable to the trained ResNet. However, its performance is slightly inferior when handling videos generated by close source models (~\cref{tab:closed_source_auc_benchmark}). This is corroborated by the greater separability between features of close source model generated videos and real videos  observed in \cref{fig:tsen_resnet}.

In summary, CLIP-ViT-L/14 shows promise as a backbone network for generated video detection, a point noted in previous work~\cite{decof,DeMamba}. Nevertheless, as observed from ~\cref{fig:tsen_uni} and ~\cref{tab:closed_source_auc_benchmark}, there remains room for improvement. The fine tuning strategy proposed by Effort significantly enhances the detection performance of the base model (\cref{fig:tsen_effrot}), improvements of 1.8\% on open source models and 28.79\% on closed source models. Furthermore, the training strategy employed by ForgeLens, which utilizes a small dataset over more training epochs (ForgeLens1 in \cref{tab:video_detection_benchmark,tab:closed_source_auc_benchmark}), proves more effective compared to using large datasets with fewer training epochs(ForgeLens3 in \cref{tab:video_detection_benchmark,tab:closed_source_auc_benchmark}), gains of 2.22\% and 8.16\% were achieved on open-source and closed source models, respectively.

\vspace{0.3cm}
\definecolor{maxcolor}{RGB}{231, 76, 60}    
\definecolor{midcolor}{RGB}{52, 152, 219}   
\definecolor{mincolor}{RGB}{46, 204, 113}   

\begin{table}[t]
\centering
\caption{Comprehensive Performance Evaluation of Video Detection Models on Text-to-Video, Image-to-Video, and Video-to-Video Generation Tasks.}
\label{tab:performance_color}
\begin{tabular}{ccccc}
\toprule
\textbf{Method} & \textbf{I2V} & \textbf{T2V} & \textbf{V2V} & \textbf{Closed} \\
\midrule
I3D & \cellcolor{maxcolor!20}89.05 & \cellcolor{midcolor!20}80.78 & \cellcolor{mincolor!20}78.07 & 61.18 \\
MVITv2 & \cellcolor{midcolor!20}65.60 & \cellcolor{maxcolor!20}66.96 & \cellcolor{mincolor!20}63.57 & 56.94 \\
SlowFast & \cellcolor{mincolor!20}72.34 & \cellcolor{maxcolor!20}82.31 & \cellcolor{midcolor!20}78.18 & 62.71 \\
VideoSwin & \cellcolor{maxcolor!20}74.97 & \cellcolor{midcolor!20}67.93 & \cellcolor{mincolor!20}62.01 & 52.47 \\
Timesformer & \cellcolor{midcolor!20}75.46 & \cellcolor{maxcolor!20}86.84 & \cellcolor{mincolor!20}67.59 & 86.50 \\
TSM & \cellcolor{maxcolor!20}83.90 & \cellcolor{midcolor!20}74.52 & \cellcolor{mincolor!20}67.78 & 56.08 \\
Uniformer & \cellcolor{mincolor!20}53.35 & \cellcolor{maxcolor!20}74.12 & \cellcolor{midcolor!20}72.21 & 57.02 \\
Uniformerv2 & \cellcolor{midcolor!20}81.27 & \cellcolor{maxcolor!20}85.94 & \cellcolor{mincolor!20}73.30 & 87.07 \\
VideoMAE & \cellcolor{mincolor!20}67.56 & \cellcolor{maxcolor!20}78.51 & \cellcolor{midcolor!20}71.40 & 57.66 \\
X3D & \cellcolor{maxcolor!20}90.40 & \cellcolor{midcolor!20}83.65 & \cellcolor{mincolor!20}67.60 & 66.04 \\
\midrule
CNNSpot & \cellcolor{maxcolor!20}89.57 & \cellcolor{midcolor!20}84.80 & \cellcolor{mincolor!20}78.60 & 70.60 \\
D3 & \cellcolor{midcolor!20}75.32 & \cellcolor{maxcolor!20}87.28 & \cellcolor{mincolor!20}75.28 & 76.58 \\
Effort & \cellcolor{midcolor!20}86.32 & \cellcolor{maxcolor!20}90.20 & \cellcolor{mincolor!20}74.74 & 94.05 \\
ForgeLens1 & \cellcolor{maxcolor!20}98.09 & \cellcolor{midcolor!20}90.22 & \cellcolor{mincolor!20}82.63 & 85.03 \\
ForgeLens3 & \cellcolor{maxcolor!20}96.98 & \cellcolor{midcolor!20}87.78 & \cellcolor{mincolor!20}78.39 & 76.87 \\
FreDect & \cellcolor{maxcolor!20}72.52 & \cellcolor{midcolor!20}69.86 & \cellcolor{mincolor!20}58.81 & 55.68 \\
Fusing & \cellcolor{maxcolor!20}86.30 & \cellcolor{midcolor!20}77.36 & \cellcolor{mincolor!20}72.33 & 65.46 \\
Gram-Net & \cellcolor{maxcolor!20}83.48 & \cellcolor{midcolor!20}80.97 & \cellcolor{mincolor!20}69.62 & 65.68 \\
NPR & \cellcolor{maxcolor!20}75.19 & \cellcolor{midcolor!20}66.85 & \cellcolor{mincolor!20}61.65 & 52.36 \\
UnivFD & \cellcolor{maxcolor!20}90.30 & \cellcolor{midcolor!20}83.97 & \cellcolor{mincolor!20}82.20 & 65.26 \\
\midrule
DeCoF & \cellcolor{mincolor!20}76.65 & \cellcolor{maxcolor!20}85.68 & \cellcolor{midcolor!20}81.59 & 72.90 \\
DeMamba & \cellcolor{maxcolor!20}82.09 & \cellcolor{midcolor!20}81.87 & \cellcolor{mincolor!20}72.47 & 69.43 \\
\bottomrule
\end{tabular}

\vspace{0.5em}
\footnotesize
\textcolor{maxcolor!100}{$\blacksquare$ Maximum} \quad 
\textcolor{midcolor!100}{$\blacksquare$ Middle} \quad 
\textcolor{mincolor!100}{$\blacksquare$ Minimum} \\
Color coding applies only to I2V, T2V and V2V columns.
\end{table}

\begin{table}[t]
\centering
\caption{ Comparison of video generation tasks on Aesthetic Quality, Background Consistency, Dynamic Degree, Imaging Quality, Motion Smoothness, Subject Consistency, Frame-Level (Image \& Aesthetic Average) and Final Score based on Vbench.~\cite{vbench}}
\label{tab:dimension_performance}
\begin{tabular}{lccc}
\toprule
\textbf{Dimension} & \textbf{I2V} & \textbf{T2V} & \textbf{V2V} \\
\midrule
Aesthetic Quality & 
\cellcolor{maxcolor!20}55.82 & 
\cellcolor{mincolor!20}58.07 & 
\cellcolor{midcolor!20}56.10 \\

Imaging Quality & 
\cellcolor{midcolor!20}63.97 & 
\cellcolor{maxcolor!20}63.12 & 
\cellcolor{mincolor!20}66.91 \\

Frame-Level &
\cellcolor{maxcolor!20}59.90&
 \cellcolor{midcolor!20}60.60&
\cellcolor{mincolor!20}61.51\\
\midrule

Background Consistency & 
\cellcolor{maxcolor!20}95.03 & 
\cellcolor{midcolor!20}97.52 & 
\cellcolor{mincolor!20}98.11 \\

Dynamic Degree (DD)  & 
\cellcolor{mincolor!20}25.10 & 
\cellcolor{midcolor!20}17.90 & 
\cellcolor{maxcolor!20}8.72 \\

Motion Smoothness & 
\cellcolor{maxcolor!20}91.02 & 
\cellcolor{midcolor!20}95.49 & 
\cellcolor{mincolor!20}97.44 \\

Subject Consistency & 
\cellcolor{maxcolor!20}94.76 & 
\cellcolor{midcolor!20}97.45 & 
\cellcolor{mincolor!20}98.54 \\
Video-Level(w/o DD)&
\cellcolor{maxcolor!20}93.60 &	
\cellcolor{midcolor!20}96.82& 	
\cellcolor{mincolor!20}98.03 \\

\midrule
Final Score & 
\cellcolor{maxcolor!20}77.40 & 
\cellcolor{mincolor!20}78.10 & 
\cellcolor{midcolor!20}77.42 \\
\bottomrule
\end{tabular}

\vspace{0.5em}
\footnotesize
\textcolor{mincolor!100}{$\blacksquare$ Best} \quad 
\textcolor{midcolor!100}{$\blacksquare$ Middle} \quad 
\textcolor{maxcolor!100}{$\blacksquare$ Worst} \\
Color coding applied per row (each dimension independently).
\end{table}
\noindent\textbf{Finding-1.1: The type of generation task has a significant impact on detector performance, and the extent of this impact varies considerably across different types of detectors.} 
\vspace{0.1cm}

~\cref{tab:performance_color}  presents the performance evaluation of various detection models across multiple video generation tasks, with the following key observations:
\definecolor{maxcolor}{RGB}{231, 76, 60}    
\definecolor{midcolor}{RGB}{52, 152, 219}   
\definecolor{mincolor}{RGB}{46, 204, 113}   

\begin{table*}[t]
\centering
\caption{Performance of detection models on videos from T2V, I2V, and V2V tasks of a single generation model.}
\label{tab:table11}
\begin{tabular}{lccc|cc|cc|cc|cc}
\toprule
& \multicolumn{3}{c|}{\textbf{LTX Dataset}} & \multicolumn{2}{c|}{\textbf{EasyAnimate}} & \multicolumn{2}{c|}{\textbf{Pyramid-Flow}} & \multicolumn{2}{c|}{\textbf{VideoCrafter}} & \multicolumn{2}{c}{\textbf{Cogvideox1.5}} \\
\cmidrule(lr){2-4} \cmidrule(lr){5-6} \cmidrule(lr){7-8} \cmidrule(lr){9-10} \cmidrule(lr){11-12}
\textbf{Method} & I2V & T2V & V2V & I2V & T2V & I2V & T2V & I2V & T2V & T2V & V2V \\
\midrule
I3D & \cellcolor{midcolor!20}81.68 & \cellcolor{maxcolor!20}94.62 & \cellcolor{mincolor!20}76.40 & \cellcolor{maxcolor!20}{81.00} & \cellcolor{midcolor!20}{78.40} & \cellcolor{midcolor!20}{94.01} & \cellcolor{maxcolor!20}{96.75} & \cellcolor{midcolor!20}{84.79} & \cellcolor{maxcolor!20}{91.49} & \cellcolor{maxcolor!20}{83.49} & \cellcolor{midcolor!20}{79.74} \\
MVITv2 & \cellcolor{midcolor!20}67.74 & \cellcolor{maxcolor!20}69.52 & \cellcolor{mincolor!20}64.93 & \cellcolor{midcolor!20}{52.26} & \cellcolor{maxcolor!20}{60.65} & \cellcolor{midcolor!20}{65.93} & \cellcolor{maxcolor!20}{70.02} & \cellcolor{maxcolor!20}{70.45} & \cellcolor{midcolor!20}{68.61} & \cellcolor{maxcolor!20}{63.61} & \cellcolor{midcolor!20}{62.21} \\
SlowFast & \cellcolor{mincolor!20}71.62 & \cellcolor{midcolor!20}76.22 & \cellcolor{maxcolor!20}77.50 & \cellcolor{midcolor!20}{63.92} & \cellcolor{maxcolor!20}{74.91} & \cellcolor{midcolor!20}{83.83} & \cellcolor{maxcolor!20}{92.75} & \cellcolor{midcolor!20}{68.36} & \cellcolor{maxcolor!20}{91.33} & \cellcolor{maxcolor!20}{80.23} & \cellcolor{midcolor!20}{78.86} \\
VideoSwin & \cellcolor{maxcolor!20}70.88 & \cellcolor{midcolor!20}70.08 & \cellcolor{mincolor!20}66.52 & \cellcolor{midcolor!20}{60.43} & \cellcolor{maxcolor!20}{62.99} & \cellcolor{maxcolor!20}{65.63} & \cellcolor{midcolor!20}{63.88} & \cellcolor{maxcolor!20}{77.10} & \cellcolor{midcolor!20}{76.33} & \cellcolor{maxcolor!20}{65.64} & \cellcolor{midcolor!20}{57.49} \\
Timesformer & \cellcolor{midcolor!20}70.38 & \cellcolor{maxcolor!20}85.51 & \cellcolor{mincolor!20}65.55 & \cellcolor{midcolor!20}{66.33} & \cellcolor{maxcolor!20}{84.34} & \cellcolor{midcolor!20}{80.32} & \cellcolor{maxcolor!20}{96.17} & \cellcolor{midcolor!20}{65.38} & \cellcolor{maxcolor!20}{96.42} & \cellcolor{maxcolor!20}{84.55} & \cellcolor{midcolor!20}{69.63} \\
TSM & \cellcolor{midcolor!20}70.38 & \cellcolor{maxcolor!20}80.06 & \cellcolor{mincolor!20}66.67 & \cellcolor{maxcolor!20}{79.41} & \cellcolor{midcolor!20}{77.64} & \cellcolor{midcolor!20}{85.88} & \cellcolor{maxcolor!20}{87.01} & \cellcolor{midcolor!20}{84.25} & \cellcolor{maxcolor!20}{94.45} & \cellcolor{midcolor!20}{68.53} & \cellcolor{maxcolor!20}{68.89} \\
Uniformer & \cellcolor{midcolor!20}60.05 & \cellcolor{mincolor!20}52.92 & \cellcolor{maxcolor!20}69.35 & \cellcolor{midcolor!20}{48.46} & \cellcolor{maxcolor!20}{65.95} & \cellcolor{midcolor!20}{66.36} & \cellcolor{maxcolor!20}{84.92} & \cellcolor{midcolor!20}{68.78} & \cellcolor{maxcolor!20}{83.73} & \cellcolor{midcolor!20}{67.57} & \cellcolor{maxcolor!20}{75.06} \\
Uniformerv2 & \cellcolor{midcolor!20}77.59 & \cellcolor{maxcolor!20}82.60 & \cellcolor{mincolor!20}72.65 & \cellcolor{midcolor!20}{73.94} & \cellcolor{maxcolor!20}{84.35} & \cellcolor{midcolor!20}{77.11} & \cellcolor{maxcolor!20}{92.95} & \cellcolor{midcolor!20}{82.28} & \cellcolor{maxcolor!20}{97.14} & \cellcolor{maxcolor!20}{82.95} & \cellcolor{midcolor!20}{73.94} \\
VideoMAE & \cellcolor{mincolor!20}69.09 & \cellcolor{maxcolor!20}75.01 & \cellcolor{midcolor!20}71.54 & \cellcolor{midcolor!20}{57.50} & \cellcolor{maxcolor!20}{66.88} & \cellcolor{midcolor!20}{77.04} & \cellcolor{maxcolor!20}{92.66} & \cellcolor{midcolor!20}{65.40} & \cellcolor{maxcolor!20}{86.68} & \cellcolor{maxcolor!20}{75.33} & \cellcolor{midcolor!20}{71.25} \\
X3D & \cellcolor{midcolor!20}68.43 & \cellcolor{maxcolor!20}84.17 & \cellcolor{mincolor!20}61.67 & \cellcolor{midcolor!20}{90.98} & \cellcolor{maxcolor!20}{93.81} & \cellcolor{midcolor!20}{92.77} & \cellcolor{maxcolor!20}{97.06} & \cellcolor{midcolor!20}{96.56} & \cellcolor{maxcolor!20}{97.48} & \cellcolor{maxcolor!20}{76.92} & \cellcolor{midcolor!20}{73.52} \\
\midrule
CNNSpot & \cellcolor{midcolor!20}81.41 & \cellcolor{maxcolor!20}96.82 & \cellcolor{mincolor!20}71.81 & \cellcolor{maxcolor!20}{88.54} & \cellcolor{midcolor!20}{88.20} & \cellcolor{midcolor!20}{97.33} & \cellcolor{maxcolor!20}{99.21} & \cellcolor{midcolor!20}{73.17} & \cellcolor{maxcolor!20}{91.97} & \cellcolor{maxcolor!20}{89.74} & \cellcolor{midcolor!20}{85.39} \\
D3 & \cellcolor{mincolor!20}77.69 & \cellcolor{maxcolor!20}83.82 & \cellcolor{midcolor!20}77.86 & \cellcolor{midcolor!20}{74.69} & \cellcolor{maxcolor!20}{84.82} & \cellcolor{midcolor!20}{82.45} & \cellcolor{maxcolor!20}{98.37} & \cellcolor{midcolor!20}{42.52} & \cellcolor{maxcolor!20}{95.37} & \cellcolor{maxcolor!20}{90.09} & \cellcolor{midcolor!20}{72.69} \\
Effort & \cellcolor{midcolor!20}79.15 & \cellcolor{maxcolor!20}83.13 & \cellcolor{mincolor!20}78.42 & \cellcolor{midcolor!20}{88.74} & \cellcolor{maxcolor!20}{97.01} & \cellcolor{midcolor!20}{94.99} & \cellcolor{maxcolor!20}{99.87} & \cellcolor{midcolor!20}{58.25} & \cellcolor{maxcolor!20}{99.61} & \cellcolor{maxcolor!20}{89.31} & \cellcolor{midcolor!20}{71.05} \\
ForgeLens1 & \cellcolor{midcolor!20}92.89 & \cellcolor{maxcolor!20}95.63 & \cellcolor{mincolor!20}91.41 & \cellcolor{midcolor!20}{98.21} & \cellcolor{maxcolor!20}{98.68} & \cellcolor{midcolor!20}{98.68} & \cellcolor{maxcolor!20}{99.71} & \cellcolor{midcolor!20}{98.86} & \cellcolor{maxcolor!20}{99.81} & \cellcolor{maxcolor!20}{87.03} & \cellcolor{midcolor!20}{73.85} \\
ForgeLens3 & \cellcolor{midcolor!20}89.67 & \cellcolor{maxcolor!20}93.76 & \cellcolor{mincolor!20}86.79 & \cellcolor{midcolor!20}{95.79} & \cellcolor{maxcolor!20}{96.17} & \cellcolor{midcolor!20}{97.78} & \cellcolor{maxcolor!20}{99.59} & \cellcolor{midcolor!20}{98.76} & \cellcolor{maxcolor!20}{99.50} & \cellcolor{maxcolor!20}{84.34} & \cellcolor{midcolor!20}{69.98} \\
FreDect & \cellcolor{midcolor!20}61.23 & \cellcolor{maxcolor!20}70.06 & \cellcolor{mincolor!20}52.62 & \cellcolor{midcolor!20}{59.70} & \cellcolor{maxcolor!20}{67.01} & \cellcolor{midcolor!20}{69.19} & \cellcolor{maxcolor!20}{73.32} & \cellcolor{midcolor!20}{77.62} & \cellcolor{maxcolor!20}{81.54} & \cellcolor{maxcolor!20}{71.51} & \cellcolor{midcolor!20}{65.00} \\
Fusing & \cellcolor{midcolor!20}73.76 & \cellcolor{maxcolor!20}90.64 & \cellcolor{mincolor!20}64.38 & \cellcolor{maxcolor!20}{81.69} & \cellcolor{midcolor!20}{76.79} & \cellcolor{midcolor!20}{93.07} & \cellcolor{maxcolor!20}{95.98} & \cellcolor{midcolor!20}{77.91} & \cellcolor{maxcolor!20}{87.98} & \cellcolor{maxcolor!20}{81.90} & \cellcolor{midcolor!20}{80.28} \\
Gram-Net & \cellcolor{midcolor!20}71.57 & \cellcolor{maxcolor!20}92.95 & \cellcolor{mincolor!20}59.62 & \cellcolor{midcolor!20}{77.22} & \cellcolor{maxcolor!20}{81.26} & \cellcolor{midcolor!20}{88.61} & \cellcolor{maxcolor!20}{95.48} & \cellcolor{midcolor!20}{77.04} & \cellcolor{maxcolor!20}{93.49} & \cellcolor{maxcolor!20}{84.00} & \cellcolor{midcolor!20}{79.61} \\
NPR & \cellcolor{midcolor!20}67.60 & \cellcolor{maxcolor!20}80.67 & \cellcolor{mincolor!20}58.58 & \cellcolor{maxcolor!20}{69.58} & \cellcolor{midcolor!20}{65.39} & \cellcolor{midcolor!20}{77.80} & \cellcolor{maxcolor!20}{85.35} & \cellcolor{midcolor!20}{59.54} & \cellcolor{maxcolor!20}{70.85} & \cellcolor{maxcolor!20}{75.61} & \cellcolor{midcolor!20}{64.72} \\
UnivFD & \cellcolor{midcolor!20}89.91 & \cellcolor{maxcolor!20}94.70 & \cellcolor{mincolor!20}89.38 & \cellcolor{midcolor!20}{87.39} & \cellcolor{maxcolor!20}{88.73} & \cellcolor{midcolor!20}{89.57} & \cellcolor{maxcolor!20}{97.39} & \cellcolor{midcolor!20}{81.12} & \cellcolor{maxcolor!20}{97.55} & \cellcolor{maxcolor!20}{84.69} & \cellcolor{midcolor!20}{75.02} \\
\midrule
DeCoF & \cellcolor{midcolor!20}81.53 & \cellcolor{maxcolor!20}87.74 & \cellcolor{mincolor!20}81.34 & \cellcolor{midcolor!20}{70.64} & \cellcolor{maxcolor!20}{78.03} & \cellcolor{midcolor!20}{83.74} & \cellcolor{maxcolor!20}{96.57} & \cellcolor{midcolor!20}{45.21} & \cellcolor{maxcolor!20}{90.77} & \cellcolor{maxcolor!20}{90.71} & \cellcolor{midcolor!20}{81.84} \\
DeMamba & \cellcolor{midcolor!20}77.59 & \cellcolor{maxcolor!20}85.28 & \cellcolor{mincolor!20}77.46 & \cellcolor{maxcolor!20}{76.37} & \cellcolor{midcolor!20}{76.59} & \cellcolor{midcolor!20}{92.84} & \cellcolor{maxcolor!20}{98.90} & \cellcolor{midcolor!20}{52.02} & \cellcolor{maxcolor!20}{88.01} & \cellcolor{maxcolor!20}{82.17} & \cellcolor{midcolor!20}{67.48} \\
\bottomrule
\end{tabular}

\vspace{0.5em}
\footnotesize
\textcolor{maxcolor!100}{$\blacksquare$ Maximum} \quad 
\textcolor{midcolor!100}{$\blacksquare$ Middle} \quad 
\textcolor{mincolor!100}{$\blacksquare$ Minimum} \\

\end{table*}

\begin{table*}[t]
\centering
\label{table12}
\caption{Comparison of Aesthetic Quality, Background Consistency, Dynamic Degree, Imaging Quality, Motion Smoothness, Subject Consistency, Frame-Level (Image \& Aesthetic Average), and Final Score  across video generation tasks of a single generation model.~\cite{vbench}}
\resizebox{\textwidth}{!}{
\begin{tabular}{l|ccc|cc|cc|cc|cc}
\toprule
& \multicolumn{3}{c|}{\textbf{LTX}} & \multicolumn{2}{c|}{\textbf{EasyAnimate}} & \multicolumn{2}{c|}{\textbf{Pyramid-Flow}} & \multicolumn{2}{c|}{\textbf{VideoCrafter}} & \multicolumn{2}{c}{\textbf{Cogvideox1.5}} \\
\cmidrule(lr){2-4} \cmidrule(lr){5-6} \cmidrule(lr){7-8} \cmidrule(lr){9-10} \cmidrule(lr){11-12}
\textbf{Metric} & I2V & T2V & V2V & I2V & T2V & I2V & T2V & I2V & T2V & T2V & V2V \\
\midrule
Aesthetic Quality & 
\cellcolor{midcolor!20}55.30 & 
\cellcolor{maxcolor!20}54.11 & 
\cellcolor{mincolor!20}56.91 & 
\cellcolor{maxcolor!20}58.34 & 
\cellcolor{midcolor!20}61.44 & 
\cellcolor{maxcolor!20}56.09 & 
\cellcolor{midcolor!20}58.32 & 
\cellcolor{maxcolor!20}57.81 & 
\cellcolor{midcolor!20}64.17 & 
\cellcolor{maxcolor!20}54.98 & 
\cellcolor{midcolor!20}55.29 \\
Imaging Quality & 
\cellcolor{midcolor!20}64.73 & 
\cellcolor{maxcolor!20}58.60 & 
\cellcolor{mincolor!20}67.58 & 
\cellcolor{midcolor!20}66.43 & 
\cellcolor{maxcolor!20}63.90 & 
\cellcolor{midcolor!20}64.09 & 
\cellcolor{maxcolor!20}63.16 & 
\cellcolor{midcolor!20}64.00 & 
\cellcolor{maxcolor!20}63.80 & 
\cellcolor{maxcolor!20}63.39 & 
\cellcolor{midcolor!20}66.24 \\
Frame-Level
&\cellcolor{midcolor!20}60.02 	&\cellcolor{maxcolor!20}56.36 	&\cellcolor{mincolor!20}62.25 	&\cellcolor{maxcolor!20}62.39 	&\cellcolor{midcolor!20}62.67 	&\cellcolor{maxcolor!20}60.09 	&\cellcolor{midcolor!20}60.74 	&\cellcolor{maxcolor!20}60.91 	&\cellcolor{midcolor!20}63.99 	&\cellcolor{maxcolor!20}59.19 	&\cellcolor{midcolor!20}60.77 
 \\
\midrule

Background Consistency & 
\cellcolor{maxcolor!20}96.71 & 
\cellcolor{midcolor!20}97.11 & 
\cellcolor{mincolor!20}98.03 & 
\cellcolor{maxcolor!20}94.85 & 
\cellcolor{midcolor!20}97.12 & 
\cellcolor{maxcolor!20}97.06 & 
\cellcolor{midcolor!20}97.16 & 
\cellcolor{maxcolor!20}97.06 & 
\cellcolor{midcolor!20}97.88 & 
\cellcolor{maxcolor!20}96.91 & 
\cellcolor{midcolor!20}98.19 \\

Dynamic Degree & 
\cellcolor{midcolor!20}17.23 & 
\cellcolor{mincolor!20}22.53 & 
\cellcolor{maxcolor!20}9.78 & 
\cellcolor{midcolor!20}31.65 & 
\cellcolor{maxcolor!20}25.22 & 
\cellcolor{midcolor!20}20.57 & 
\cellcolor{maxcolor!20}18.58 & 
\cellcolor{midcolor!20}15.50 & 
\cellcolor{maxcolor!20}11.27 & 
\cellcolor{midcolor!20}27.25 & 
\cellcolor{maxcolor!20}7.65 \\

Motion Smoothness & 
\cellcolor{maxcolor!20}95.46 & 
\cellcolor{midcolor!20}96.90 & 
\cellcolor{mincolor!20}97.14 & 
\cellcolor{maxcolor!20}90.02 & 
\cellcolor{midcolor!20}93.09 & 
\cellcolor{maxcolor!20}96.01 & 
\cellcolor{midcolor!20}96.23 & 
\cellcolor{maxcolor!20}94.49 & 
\cellcolor{midcolor!20}95.35 & 
\cellcolor{maxcolor!20}94.80 & 
\cellcolor{midcolor!20}97.73 \\

Subject Consistency & 
\cellcolor{maxcolor!20}95.97 & 
\cellcolor{midcolor!20}96.66 & 
\cellcolor{mincolor!20}98.30 & 
\cellcolor{maxcolor!20}94.04 & 
\cellcolor{midcolor!20}97.28 & 
\cellcolor{midcolor!20}97.03 & 
\cellcolor{maxcolor!20}96.86 & 
\cellcolor{maxcolor!20}97.15 & 
\cellcolor{midcolor!20}97.95 & 
\cellcolor{maxcolor!20}96.74 & 
\cellcolor{midcolor!20}98.79 \\
\midrule
Final Score & 
\cellcolor{maxcolor!20}77.35 & 
\cellcolor{midcolor!20}77.44 & 
\cellcolor{mincolor!20}77.77 & 
\cellcolor{maxcolor!20}79.15 & 
\cellcolor{midcolor!20}79.64 & 
\cellcolor{midcolor!20}78.34 & 
\cellcolor{maxcolor!20}78.24 & 
\cellcolor{maxcolor!20}77.46 & 
\cellcolor{midcolor!20}78.26 & 
\cellcolor{midcolor!20}78.92 & 
\cellcolor{maxcolor!20}77.07 \\
\bottomrule
\end{tabular}}

\vspace{0.5em}

\footnotesize
\textcolor{mincolor!100}{$\blacksquare$ Best} \quad 
\textcolor{midcolor!100}{$\blacksquare$ Middle} \quad 
\textcolor{maxcolor!100}{$\blacksquare$ Worst} \\
\end{table*}
1. \textbf{ Video to Video (V2V) consistently proves to be the most challenging task to detect across all models.} This result is expected, as V2V utilizes a real video as reference, leading to generated content that generally outperforms other tasks in various metrics (as also corroborated by~\cref{tab:dimension_performance}). The lower Dynamic Degree observed in V2V outputs is primarily due to the high frame rate of the input real video, which restricts the dynamic variation range of the generated video.

2.\textbf{ Image-level detectors show a clear task-dependent performance trend.} For image-level detectors, the influence of the generation task type on detection performance exhibits a clear and consistent pattern. Most models achieve the best detection results on Image to Video (I2V), followed by Text to Video (T2V), with V2V being the most difficult. It is noteworthy that this detection performance ranking is inversely correlated with the overall image-level quality ranking of the three tasks: V2V achieves the highest visual quality, followed by T2V, while I2V scores the lowest. Although the performances of Effort and D3 differ from the majority of models, their results align with the ranking of tasks based on the Imaging Quality metric. Therefore, it can be inferred that for image-level detectors, the generation task type does influence detection effectiveness, and this influence is negatively correlated with the image-level quality of the generated content higher quality leads to greater detection difficulty. However, this relationship likely stems from the combined effect of multiple quality metrics rather than being dictated by any single indicator.

3.\textbf{Video-level detectors exhibit more complex behavior influenced by both video and image attributes.} For video-level detectors, the impact of task type is more complex, influenced not only by video-level metrics but also by image-level indicators. According to the video-level metrics in~\cref{tab:dimension_performance} (disregarding Dynamic Degree), V2V shows the best overall performance, followed by T2V, with I2V being the weakest. A straightforward interpretation is that although image-to-video (I2V) generation benefits from richer input information compared to text-to-video (T2V) generation, the reference image may, to some extent, constrain the temporal coherence and overall consistency of the generated video in certain generative models.  We hypothesize that, under the condition of using a consistent generative model, the output quality would exhibit a progressive improvement from T2V to I2V, and further to video-to-video (V2V) generation. This hypothesis will be specifically validated in point 4.

4.  \textbf{For the same generative model, detectors typically underperform on V2V tasks compared to I2V tasks, while I2V tasks present greater challenges than T2V tasks. This trend aligns with the output quality of generative models across different tasks.} As shown in \cref{tab:table11}, for the majority of detectors, the detection difficulty of V2V content generated by the same model is higher than that of I2V content, which in turn is higher than T2V content. Although a few exceptions exist, these are attributed to inherent differences among the detectors. From the perspective of generative models, the same model generally achieves higher generation quality in V2V tasks than in I2V tasks, and higher in I2V than in T2V tasks, as shown in \cref{table12}. However, certain generative models, such as EasyAnimate, Pyramid-Flow and VideoCrafter do not strictly follow this pattern in I2V and T2V tasks. Notably, these anomalies in generation quality do not significantly affect detector performance; detectors consistently reflect an increasing order of difficulty: V2V $>$ I2V $>$ T2V. Further analysis reveals that although I2V generation quality is lower than T2V across most metrics, it outperforms T2V in terms of "imaging quality" and "dynamic extent." This suggests that these two metrics may play a critical role in detection difficulty. 

However, different video-level detection models exhibit considerable variation in performance across tasks, without forming a consistent pattern. While some models align with the trend indicated by video-level metrics, others show significant heterogeneity. This complexity motivates further investigation, leading to the research question raised in \cref{FQ2}.

\vspace{0.3cm}
\noindent\textbf{Finding-1.2: Current VLMs lack reliable capability for detecting AI-generated videos.}
\begin{figure}[H]
    \centering
    \includegraphics[width=\linewidth]{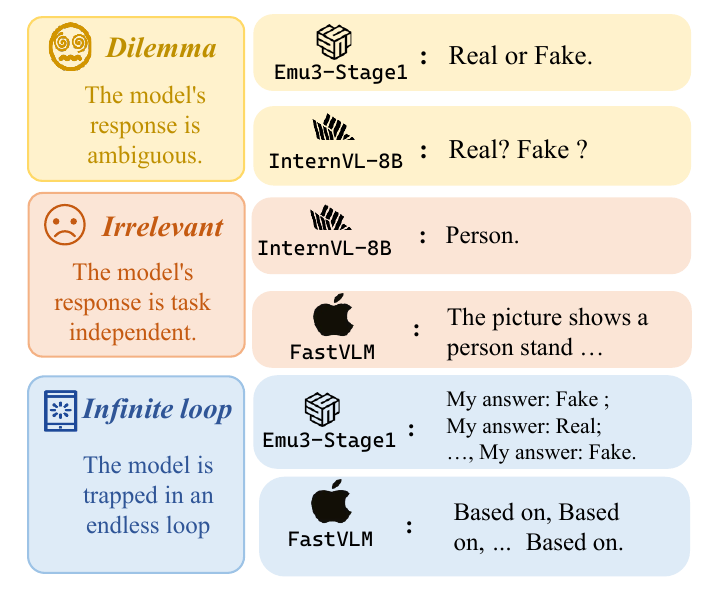}
    \caption{Anomalies in VLM for AI generated video detection.} 
    \label{fig:abnormal_vlm}
\end{figure}
\vspace{0.3cm}
In~\cref{7.7}, we detail the experimental setup for VLMs: given an input video, the model is required to determine whether the content is AI-generated or authentic and return a binary Boolean output. For VLMs lacking multi-frame reasoning capability, we adopt a frame-by-frame evaluation strategy. Specifically, if all frames are classified as "Fake", the video is considered "Fake"; if at least one frame is judged as "Real", the video is labeled as "Real". Videos that do not elicit a valid response from the model are categorized as "No Answer".

\begin{table*}[t]
  \centering
  \caption{ACC Performance Comparison of Vision-Language Models on Open-Source Generative Models: a video is labeled as Fake if at least one frame is classified as Fake.}
  \vspace{0.2cm}
  
  \resizebox{\textwidth}{!}{
    \footnotesize
    \begin{tabular}{>{\centering\arraybackslash}m{2.8cm}|*{20}{c|}c}
    \toprule
    \multirow{3}{*}{\textbf{Method}} & \multicolumn{6}{c|}{\textbf{I2V}} & \multicolumn{12}{c|}{\textbf{T2V}} & \multicolumn{2}{c|}{\textbf{V2V}} & \multirow{3}{*}{\textbf{AVG}} \\
    \cmidrule(lr){2-7} \cmidrule(lr){8-19} \cmidrule(lr){20-21}
    & \makecell[c]{\scriptsize Easy\\\scriptsize Animate} & \makecell[c]{\scriptsize LTX} & \makecell[c]{\scriptsize Pyramid\\\scriptsize Flow} & \makecell[c]{\scriptsize SEINE} & \makecell[c]{\scriptsize SVD} & \makecell[c]{\scriptsize Video\\\scriptsize Crafter} & \makecell[c]{\scriptsize Acc\\\scriptsize Video} & \makecell[c]{\scriptsize Animate\\\scriptsize Diff} & \makecell[c]{\scriptsize Cogvideo\\\scriptsize x1.5} & \makecell[c]{\scriptsize Easy\\\scriptsize Animate} & \makecell[c]{\scriptsize Hunyuan} & \makecell[c]{\scriptsize IPOC} & \makecell[c]{\scriptsize LTX} & \makecell[c]{\scriptsize Open\\\scriptsize Sora} & \makecell[c]{\scriptsize Pyramid\\\scriptsize Flow} & \makecell[c]{\scriptsize Rep\\\scriptsize Video} & \makecell[c]{\scriptsize Video\\\scriptsize Crafter} & \makecell[c]{\scriptsize Wan\\\scriptsize 2.1} & \makecell[c]{\scriptsize Cogvideo\\\scriptsize x1.5} & \makecell[c]{\scriptsize LTX} & \\
    \midrule
    Emu3-Stage1~\cite{wang2024emu3} &33.19 & 33.53 & 33.01 & 32.24 & 28.67 & 32.78 & 31.01 & 40.57 & 31.78 & 32.29 & 30.03 & 31.25 & 33.64 & 27.63 & 30.62 & 31.64 & 42.84 & 31.92 & 34.63 & 35.29 & 32.93 \\
    FastVLM-7B~\cite{fastvlm} &46.61 & 47.74 & 47.38 & 47.68 & 50.63 & 49.69 & 47.61 & 49.03 & 48.30 & 46.85 & 47.58 & 48.63 & 48.78 & 48.58 & 46.89 & 48.91 & 51.14 & 46.72 & 48.21 & 47.54 & 48.22 \\
   DeepseekVL2-S~\cite{DeepSeek-VL2}& 49.02 & 51.35 & 48.60 & 49.37 & 54.82 & 48.90 & 48.38 & 48.27 & 48.62 & 48.42 & 48.37 & 48.47 & 49.35 & 48.57 & 49.08 & 49.45 & 48.32 & 48.43 & 48.37 & 48.62 & 49.14  \\
   LLaVA-v1.5-7B~\cite{liu2023improvedllava} &50.00 & 50.00 & 50.00 & 50.00 & 50.00 & 50.00 & 50.00 & 50.00 & 50.00 & 50.00 & 50.00 & 50.00 & 50.00 & 50.00 & 50.00 & 50.00 & 50.00 & 50.00 & 50.00 & 50.00 & 50.00 \\
   Kimi-VL-A3B~\cite{Kimi-vl} &50.27 & 52.17 & 50.08 & 50.83 & 55.37 & 51.57 & 50.10 & 51.03 & 50.47 & 50.07 & 49.98 & 50.33 & 50.83 & 49.80 & 51.45 & 51.55 & 50.87 & 50.00 & 50.07 & 50.03 & 50.84 \\
    Qwen2.5-VL-3B~\cite{Qwen2.5-VL} & 49.05 & 51.50 & 48.77 & 53.00 & 60.03 & 61.10 & 49.10 & 56.47 & 50.27 & 45.52 & 47.65 & 50.80 & 51.22 & 45.87 & 45.97 & 52.37 & 58.43 & 48.87 & 50.40 & 49.03 & 51.27  \\
    Deepseek-VL-7B~\cite{lu2024deepseekvl} & 51.03 & 55.10 & 50.58 & 53.13 & 59.30 & 52.50 & 50.15 & 49.80 & 51.82 & 50.13 & 50.13 & 50.50 & 54.42 & 50.50 & 50.72 & 53.37 & 50.05 & 49.95 & 50.78 & 50.30 & 51.71  \\
    DeepseekVL2~\cite{DeepSeek-VL2} & 52.18 & 56.42 & 51.28 & 58.43 & 67.07 & 59.77 & 49.22 & 50.08 & 53.72 & 49.17 & 48.98 & 52.22 & 57.80 & 49.88 & 52.83 & 55.30 & 50.58 & 49.23 & 50.20 & 50.43 & 53.24   \\
   
    Qwen2.5-VL-32B~\cite{Qwen2.5-VL} & 49.68 & 51.92 & 49.50 & 51.58 & 61.30 & 59.98 & 51.73 & 62.10 & 53.67 & 48.90 & 49.83 & 53.87 & 50.73 & 49.93 & 50.78 & 54.05 & 64.43 & 50.05 & 52.00 & 50.25 & 53.32    \\

    Qwen2.5-VL-7B~\cite{Qwen2.5-VL} & 52.45 & 54.75 & 51.98 & 55.22 & 61.72 & 59.57 & 53.67 & 59.08 & 55.32 & 51.25 & 52.68 & 55.08 & 54.82 & 53.72 & 53.10 & 58.28 & 63.35 & 53.73 & 53.32 & 52.00 & 55.25 \\

    InternVL-8B~\cite{zhu2025internvl3} &50.33 & 55.39 & 51.28 & 58.18 & 65.45 & 64.44 & 52.11 & 63.59 & 55.57 & 46.73 & 51.89 & 55.98 & 58.12 & 53.72 & 52.98 & 57.65 & 67.32 & 52.85 & 52.34 & 50.99 & 55.85   \\
    \bottomrule
    \end{tabular}%
  }
  \label{tab:VLM fake is fake}%
\end{table*}

\begin{table*}[t]
\centering
\caption{ACC Performance Comparison of Vision-Language Models on Closed-Source Generative Models: a video is labeled as Fake if at least one frame is classified as Fake.}
\vspace{0.2cm}
\resizebox{\textwidth}{!}{
\begin{tabular}{lccccccccccccc}
\toprule
\textbf{Method} & \textbf{Luma} & \textbf{OpenSora} & \textbf{Sora} & \textbf{Causvid} & \textbf{Gen2} & \textbf{Gen3} & \textbf{Jimeng} & \textbf{Kling} & \textbf{Pika} & \textbf{Vidu} & \textbf{Wan} & \textbf{AVG} \\
\midrule
Emu3-Stage1~\cite{wang2024emu3}& 36.78 & 38.91 & 37.31 & 41.84 & 40.16 & 34.89 & 42.76 & 39.69 & 37.28 & 37.75 & 36.18 & 38.50 \\
DeepseekVL2-S~\cite{DeepSeek-VL2} & 48.63 & 48.58 & 48.40 & 48.28 & 48.18 & 49.25 & 48.23 & 48.58 & 48.78 & 48.83 & 48.45 & 48.56 \\
LLaVA-v1.5-7B~\cite{liu2023improvedllava} & 50.00 & 50.00 & 50.00 & 50.00 & 50.00 & 50.00 & 50.00 & 50.00 & 50.00 & 50.00 & 50.00 & 50.00 \\
Deepseek-VL-7B~\cite{lu2024deepseekvl}& 50.90 & 49.95 & 50.40 & 50.33 & 49.63 & 51.40 & 49.75 & 51.03 & 50.83 & 50.40 & 50.33 & 50.45 \\
Kimi-VL-A3B~\cite{Kimi-vl} & 53.60 & 52.55 & 53.78 & 55.23 & 49.95 & 58.23 & 50.30 & 59.90 & 56.10 & 54.70 & 53.73 & 54.37 \\
FastVLM-7B~\cite{fastvlm} & 51.85 & 52.88 & 50.85 & 53.70 & 48.50 & 52.68 & 51.44 & 60.16 & 57.00 & 52.84 & 53.48 & 53.21 \\
Qwen2.5-VL-7B~\cite{Qwen2.5-VL} & 59.83 & 57.83 & 56.00 & 60.35 & 50.68 & 60.45 & 55.73 & 70.23 & 67.55 & 59.23 & 59.03 & 59.72 \\
Qwen2.5-VL-32B~\cite{Qwen2.5-VL} & 59.85 & 60.83 & 55.60 & 57.93 & 50.65 & 58.40 & 61.63 & 67.75 & 70.38 & 58.38 & 59.90 & 60.12 \\
InternVL3-8B~\cite{zhu2025internvl3} & 62.78 & 58.43 & 53.18 & 60.08 & 47.85 & 59.05 & 56.18 & 62.46 & 68.59 & 52.38 & 56.39 & 57.94  \\
Qwen2.5-VL-3B~\cite{Qwen2.5-VL} & 64.25 & 61.33 & 64.73 & 63.78 & 56.90 & 68.10 & 61.95 & 71.68 & 75.53 & 63.08 & 66.00 & 65.21 \\
DeepseekVL2~\cite{DeepSeek-VL2} & 87.10 & 87.60 & 84.00 & 85.43 & 82.20 & 86.10 & 83.83 & 85.85 & 91.75 & 84.18 & 84.18 & 85.65 \\
\bottomrule
\end{tabular}}
\label{tab:VLM fake is fake on closed source}
\end{table*}

\begin{table*}[t]
  \centering
  \caption{ACC Performance Comparison of Vision-Language Models on Open-Source Generative Models  under the majority voting setting: the video label is determined by the majority classification of its individual frames.}
  \vspace{0.2cm}
  \resizebox{\textwidth}{!}{
    \footnotesize
    \begin{tabular}{>{\centering\arraybackslash}m{2.8cm}|*{20}{c|}c}
    \toprule
    \multirow{3}{*}{\textbf{Method}} & \multicolumn{6}{c|}{\textbf{I2V}} & \multicolumn{12}{c|}{\textbf{T2V}} & \multicolumn{2}{c|}{\textbf{V2V}} & \multirow{3}{*}{\textbf{AVG}} \\
    \cmidrule(lr){2-7} \cmidrule(lr){8-19} \cmidrule(lr){20-21}
    & \makecell[c]{\scriptsize Easy\\\scriptsize Animate} & \makecell[c]{\scriptsize LTX} & \makecell[c]{\scriptsize Pyramid\\\scriptsize Flow} & \makecell[c]{\scriptsize SEINE} & \makecell[c]{\scriptsize SVD} & \makecell[c]{\scriptsize Video\\\scriptsize Crafter} & \makecell[c]{\scriptsize Acc\\\scriptsize Video} & \makecell[c]{\scriptsize Animate\\\scriptsize Diff} & \makecell[c]{\scriptsize Cogvideo\\\scriptsize x1.5} & \makecell[c]{\scriptsize Easy\\\scriptsize Animate} & \makecell[c]{\scriptsize Hunyuan} & \makecell[c]{\scriptsize IPOC} & \makecell[c]{\scriptsize LTX} & \makecell[c]{\scriptsize Open\\\scriptsize Sora} & \makecell[c]{\scriptsize Pyramid\\\scriptsize Flow} & \makecell[c]{\scriptsize Rep\\\scriptsize Video} & \makecell[c]{\scriptsize Video\\\scriptsize Crafter} & \makecell[c]{\scriptsize Wan\\\scriptsize 2.1} & \makecell[c]{\scriptsize Cogvideo\\\scriptsize x1.5} & \makecell[c]{\scriptsize LTX} & \\
    \midrule
   Emu3-Stage1~\cite{wang2024emu3} & 33.23 & 33.57 & 33.04 & 32.28 & 28.70 & 32.82 & 31.04 & 40.60 & 31.82 & 32.33 & 30.07 & 31.28 & 33.68 & 27.66 & 30.65 & 31.68 & 42.88 & 31.95 & 34.66 & 35.33 & 32.96 \\
FastVLM-7B~\cite{fastvlm}& 46.56 & 47.74 & 47.38 & 47.59 & 50.53 & 49.59 & 47.58 & 48.98 & 48.15 & 46.73 & 47.57 & 48.50 & 48.71 & 48.61 & 46.88 & 48.73 & 51.04 & 46.57 & 48.14 & 47.54 & 48.15 \\
DeepseekVL2-S~\cite{DeepSeek-VL2} & 49.20 & 49.58 & 49.23 & 49.27 & 50.92 & 49.35 & 49.18 & 49.17 & 49.22 & 49.17 & 49.15 & 49.22 & 49.37 & 49.15 & 49.50 & 49.22 & 49.15 & 49.15 & 49.20 & 49.22 & 49.33 \\
LLaVA-v1.5-7B~\cite{liu2023improvedllava} & 50.00 & 50.00 & 50.00 & 50.00 & 50.00 & 50.00 & 50.00 & 50.00 & 50.00 & 50.00 & 50.00 & 50.00 & 50.00 & 50.00 & 50.00 & 50.00 & 50.00 & 50.00 & 50.00 & 50.00 & 50.00 \\
Kimi-VL-A3B~\cite{Kimi-vl} & 49.98 & 50.08 & 50.00 & 50.20 & 52.02 & 50.88 & 50.22 & 50.87 & 50.17 & 50.05 & 50.12 & 50.23 & 50.28 & 49.85 & 50.92 & 50.50 & 50.58 & 49.93 & 50.08 & 50.03 & 50.35 \\
Deepseek-VL-7B~\cite{lu2024deepseekvl} & 50.05 & 51.30 & 50.20 & 50.52 & 52.93 & 50.78 & 50.20 & 49.90 & 50.68 & 50.08 & 50.10 & 50.12 & 51.65 & 50.07 & 50.32 & 50.75 & 50.05 & 49.95 & 50.42 & 50.20 & 50.51 \\
Qwen2.5-VL-3B~\cite{Qwen2.5-VL} & 48.98 & 50.62 & 49.63 & 50.08 & 53.78 & 56.33 & 49.87 & 54.88 & 50.30 & 48.02 & 49.32 & 50.47 & 50.78 & 47.92 & 48.30 & 51.22 & 54.48 & 49.17 & 50.78 & 49.83 & 50.74 \\
Qwen2.5-VL-32B~\cite{Qwen2.5-VL} & 49.43 & 50.25 & 49.80 & 49.70 & 53.60 & 55.48 & 51.28 & 58.07 & 52.32 & 49.65 & 50.23 & 52.27 & 50.18 & 49.60 & 50.87 & 52.67 & 58.72 & 49.63 & 51.57 & 50.18 & 51.78 \\
DeepseekVL2~\cite{DeepSeek-VL2} & 50.50 & 53.07 & 51.25 & 53.60 & 59.63 & 57.15 & 50.08 & 50.95 & 52.08 & 49.08 & 49.73 & 51.15 & 54.93 & 50.15 & 52.80 & 53.33 & 50.93 & 49.37 & 50.90 & 50.97 & 52.08 \\
Qwen2.5-VL-7B~\cite{Qwen2.5-VL} & 50.70 & 51.58 & 51.28 & 51.78 & 55.48 & 54.67 & 52.52 & 56.40 & 52.80 & 50.58 & 51.88 & 53.03 & 52.35 & 51.93 & 51.92 & 54.90 & 57.75 & 51.77 & 51.80 & 51.18 & 52.82 \\
InternVL3-8B~\cite{zhu2025internvl3} & 49.02 & 54.43 & 51.75 & 55.18 & 62.70 & 63.56 & 54.14 & 63.63 & 56.07 & 47.67 & 53.43 & 56.27 & 57.40 & 53.82 & 53.65 & 57.70 & 67.83 & 52.80 & 53.49 & 52.11 & 55.83 \\
    \bottomrule
    \end{tabular}%
  }
  \label{tab:VLM majority voting setting}%
\end{table*}

\begin{table*}[t]
\centering
\caption{ACC Performance Comparison of Vision-Language Models on Closed-Source Generative Models  under the strict evaluation setting: the video label is determined by the majority classification of its individual frames.}
\vspace{0.2cm}
\resizebox{\textwidth}{!}{
\begin{tabular}{lccccccccccccc}
\toprule
\textbf{Method} & \textbf{Luma} & \textbf{OpenSora} & \textbf{Sora} & \textbf{Causvid} & \textbf{Gen2} & \textbf{Gen3} & \textbf{Jimeng} & \textbf{Kling} & \textbf{Pika} & \textbf{Vidu} & \textbf{Wan} & \textbf{AVG} \\
\midrule
Emu3-Stage1~\cite{wang2024emu3} & 36.83 & 38.96 & 37.36 & 41.89 & 40.21 & 34.94 & 42.81 & 39.74 & 37.33 & 37.80 & 36.23 & 38.55 \\
DeepseekVL2-S~\cite{DeepSeek-VL2} & 49.05 & 49.00 & 49.03 & 49.00 & 49.00 & 49.05 & 49.00 & 49.00 & 49.18 & 49.08 & 49.08 & 49.04 \\
Deepseek-VL-7B~\cite{lu2024deepseekvl} & 50.28 & 49.90 & 50.18 & 50.08 & 49.90 & 50.33 & 49.93 & 50.15 & 50.35 & 50.05 & 50.08 & 50.11 \\
LLaVA-v1.5-7B~\cite{liu2023improvedllava} & 50.00 & 50.00 & 50.00 & 50.00 & 50.00 & 50.00 & 50.00 & 50.00 & 50.00 & 50.00 & 50.00 & 50.00 \\
Kimi-VL-A3B~\cite{Kimi-vl} & 52.45 & 50.95 & 52.28 & 53.05 & 49.68 & 54.30 & 50.03 & 56.45 & 53.65 & 52.75 & 52.40 & 52.54 \\
FastVLM-7B~\cite{fastvlm} & 51.83 & 52.88 & 50.85 & 53.73 & 48.58 & 52.63 & 51.49 & 60.01 & 57.00 & 52.89 & 53.53 & 53.22 \\
Qwen2.5-VL-7B~\cite{Qwen2.5-VL} & 56.00 & 54.58 & 53.18 & 56.05 & 49.88 & 54.90 & 52.43 & 66.53 & 61.73 & 56.83 & 56.08 & 56.20 \\
Qwen2.5-VL-32B~\cite{Qwen2.5-VL} & 56.93 & 58.35 & 53.80 & 55.50 & 50.20 & 55.48 & 58.50 & 65.28 & 66.20 & 57.03 & 58.05 & 57.75 \\
InternVL3-8B~\cite{zhu2025internvl3} & 60.85 & 57.83 & 52.40 & 58.88 & 48.45 & 56.93 & 55.73 & 63.31 & 68.86 & 52.83 & 55.86 & 57.45 \\
Qwen2.5-VL-3B~\cite{Qwen2.5-VL} & 58.60 & 57.43 & 59.58 & 57.63 & 55.33 & 62.38 & 56.70 & 70.10 & 70.45 & 59.98 & 61.68 & 60.89 \\
DeepseekVL2~\cite{DeepSeek-VL2} & 83.13 & 84.13 & 77.70 & 79.28 & 76.93 & 80.73 & 78.58 & 82.20 & 91.35 & 77.85 & 79.70 & 81.05 \\
\bottomrule
\end{tabular}}
\label{tab:VLM majority voting setting on closed source}

\end{table*}

As illustrated in ~\cref{fig:abnormal_vlm}, we classify these "No Answer" cases into three main types:
\begin{itemize}
    \item Dilemma: The model produces an ambiguous response containing both "Real" and "Fake" labels in a single judgment. This differs from the case where a model lacks multi-frame reasoning ability; here, the ambiguity arises within a single inference step.
    \item Irrelevant Response: The model provides answers unrelated to the detection task, behaving as though it is performing a video question-answering task, such as describing the video content.
    \item Infinite Loop: The model enters an infinite repetition cycle, generating either meaningless strings of characters or continuously revising its own responses.
    
\end{itemize}

Although such outputs may occur occasionally under normal circumstances, we observe that "No Answer" responses constitute a considerable proportion of the total outputs for many VLMs. To ensure a fair evaluation, these invalid responses are assigned a 50\% accuracy rate in our calculations.

Furthermore, as indicated in As shown in~\cref{tab:video_detection_benchmark,tab:closed_source_acc_benchmark}, even after excluding "No Answer" cases, the performance of VLMs that provide normal responses remains inadequate for the generated video detection task. It should be noted that the evaluation criteria applied to VLMs without temporal reasoning capabilities may be particularly stringent: a model must correctly identify every frame in a video as fake to classify the video as "Fake". To examine the potential influence of this experimental design, we recalculated the accuracy using two alternative strategies in~\cref{tab:VLM fake is fake,tab:VLM fake is fake on closed source,tab:VLM majority voting setting,tab:VLM majority voting setting on closed source}: majority voting, and a more lenient approach where a video is labeled "Fake" if at least one frame is detected as fake. However, even under these adjusted evaluation protocols, VLM performance remains insufficient for reliably performing the generated video detection task. Moreover, for VLMs within the same architecture family, increasing the number of model parameters does not lead to significant performance gains.

\vspace{0.2cm}

\noindent\textbf{Analysis-1.3: An exploration of whether the performance superiority of DeepSeek-VL2 signifies a genuine ability to discern authenticity.}
\begin{figure}[htbp]
    \centering
    \includegraphics[width=\linewidth]{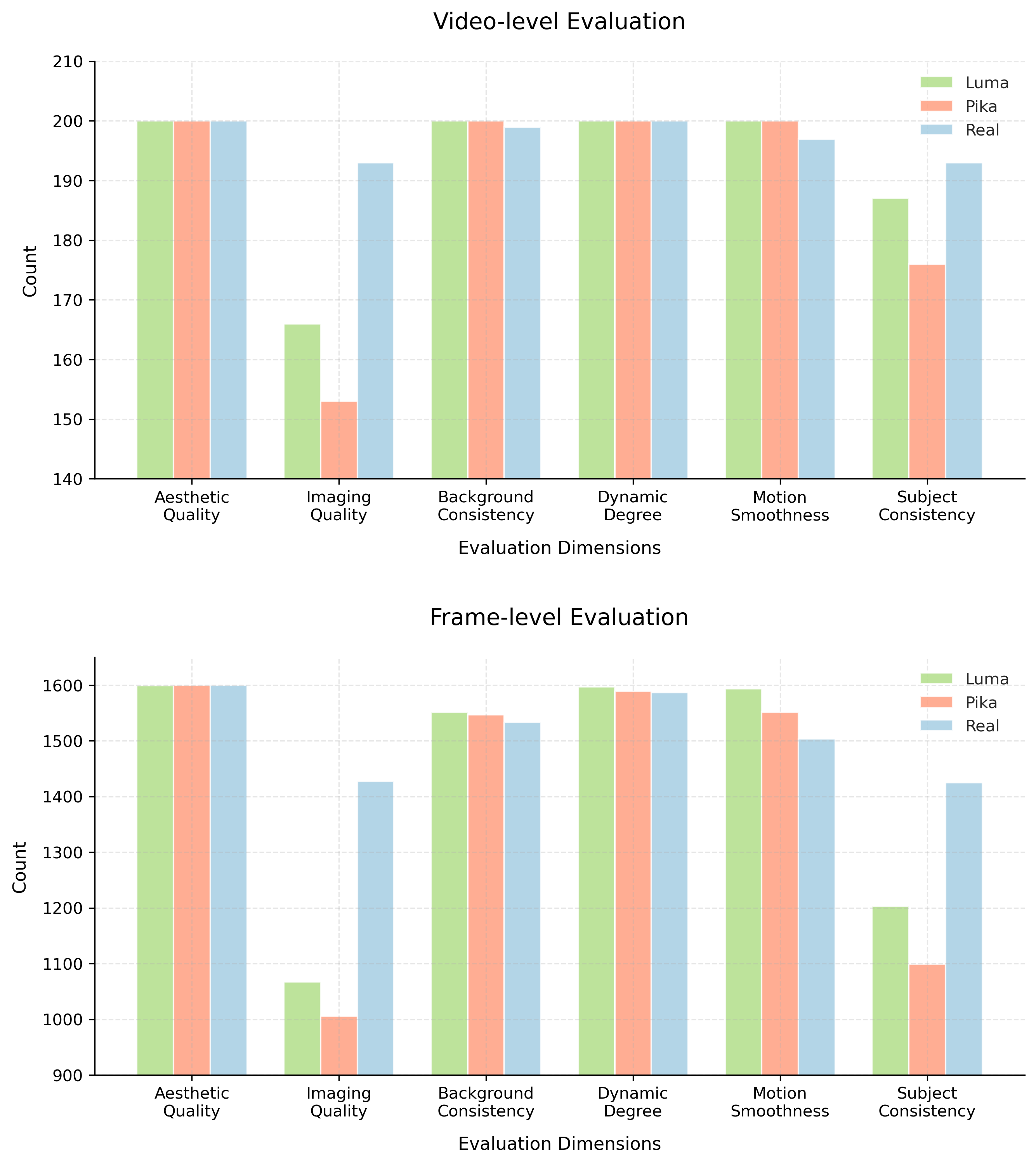}
    \caption{DeepSeek-VL-2's Preference for Evaluation Metrics in AI-Generated Video Detection.} 
    \label{fig:multi_lables_s}
\end{figure}

\begin{table}[htbp]
\centering
\caption{Performance Comparison of DeepSeek-VL2 Across Different Experimental Tasks.}
\label{tab:reson}
\resizebox{\linewidth}{!}{
\begin{tabular}{lccc
}
\toprule
\textbf{Model} & \textbf{F\_ACC} & \textbf{R\_ACC} & \textbf{ACC} \\
\midrule
DeepseekVL2 & 58.00 & 95.50 & 76.75 \\
DeepseekVL2+SVA & 0.00 & 99.50 & 49.75 \\
DeepseekVL2+MAS  & 99.50 & 0.00 & 49.75 \\
\bottomrule
\end{tabular}}
\end{table}

To further investigate the strong performance of DeepSeek-VL2 on closed-source models, we conducted two experiments using the first 200 generated videos from the Luma test set in AIGVDbench and their corresponding real videos:
\begin{itemize}
 
   \item  Synthetic Video Attribution (SVA): The model was asked to provide open-ended responses to determine video authenticity and explain its reasoning.
   \item  Metric Attribution Selection (MAS): The model was required to select potential factors from six evaluation metrics for generated videos that influenced its judgment on video authenticity.

\end{itemize}

\begin{figure*}[h]
    \centering
    \includegraphics[width=0.85\linewidth]{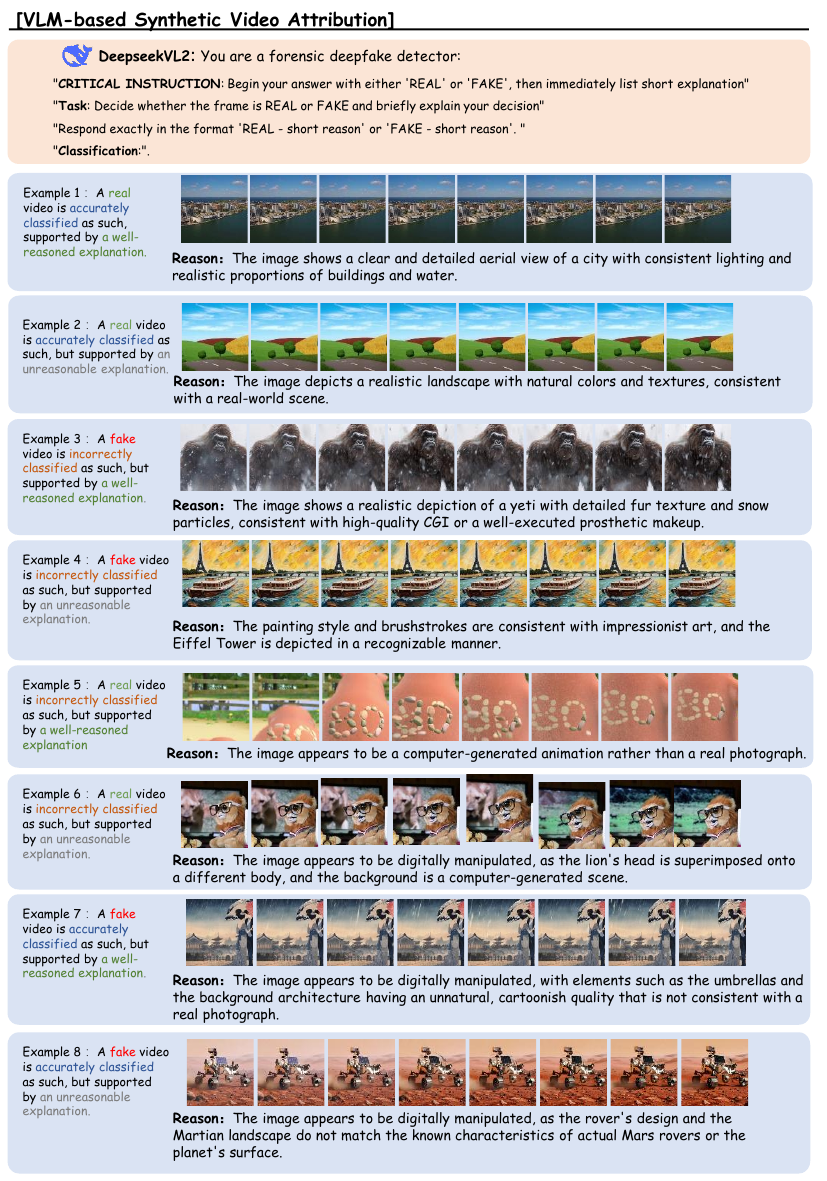}
    \caption{Prompt and Model Responses for Synthetic Video Attribution.} 
    \label{fig:reason}
\end{figure*}

\begin{figure*}[t]
    \centering
    \vspace{2cm}
    \includegraphics[width=\linewidth]{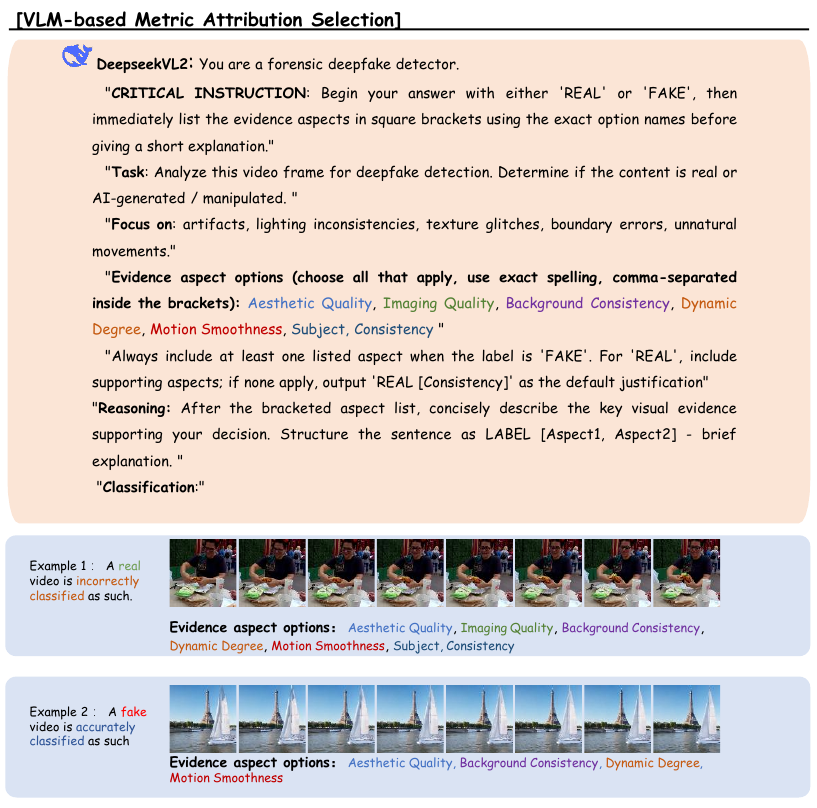}
    \caption{Prompt and Model Responses for Metric Attribution Selection.} 
     \vspace{2cm}
    \label{fig:multi_lables}
\end{figure*}

The prompts and examples for these two experiments are shown in ~\cref{fig:reason,fig:multi_lables}. Notably, despite the prompts being designed without apparent bias, the model's judgments in these two experiments diverged significantly. As shown in ~\cref{tab:reson}, in Experiment 1, the model classified the vast majority of videos as real, while Experiment 2 demonstrated the opposite tendency.
~\cref{fig:reason,fig:multi_lables} revealed that in the open-ended responses of Experiment 1, the model's justifications were notably inconsistent. Whether the classification was correct or not, its explanations ranged from reasonable to illogical. In Experiment 2, however, statistical analysis of frame-level and video-level selection results in ~\cref{fig:multi_lables_s} revealed certain patterns in the model's behavior. For instance, in the imaging quality dimension, the model exhibited a preference of Real \> Luma \> Pika. Combined with its judgment results, this suggests that DeepSeek-VL2 considers the image quality of real videos to be inferior to Luma, which in turn is inferior to Pika. However, according to our previous evaluations of generative models, Luma's generation quality in AIGVDbench should be superior to Pika.

Based on these findings, we conclude that although DeepSeek-VL2's performance on closed-source models is encouraging, it remains inadequate for the task of AI-generated video detection. This conclusion is supported by both its performance on open-source models and the two experiments designed to probe its explainability. Furthermore, the reason for its strong performance on closed-source models remains unclear. Based on comparative analysis of the experiments, we hypothesize that a potential reason could be that during the training of the closed-source model, DeepSeek-VL2 might have been exposed to simple classification tasks or training data for other tasks that included a certain proportion of videos generated by closed-source models, leading to specific recognition biases toward such videos.
\subsection{Further Findings and Analyses of Findings-2}
\label{FQ2}

\begin{figure*}[t]
  \centering
  \vspace{1cm}
  \begin{subfigure}{0.92\linewidth}
    \centering
    \includegraphics[width=\linewidth]{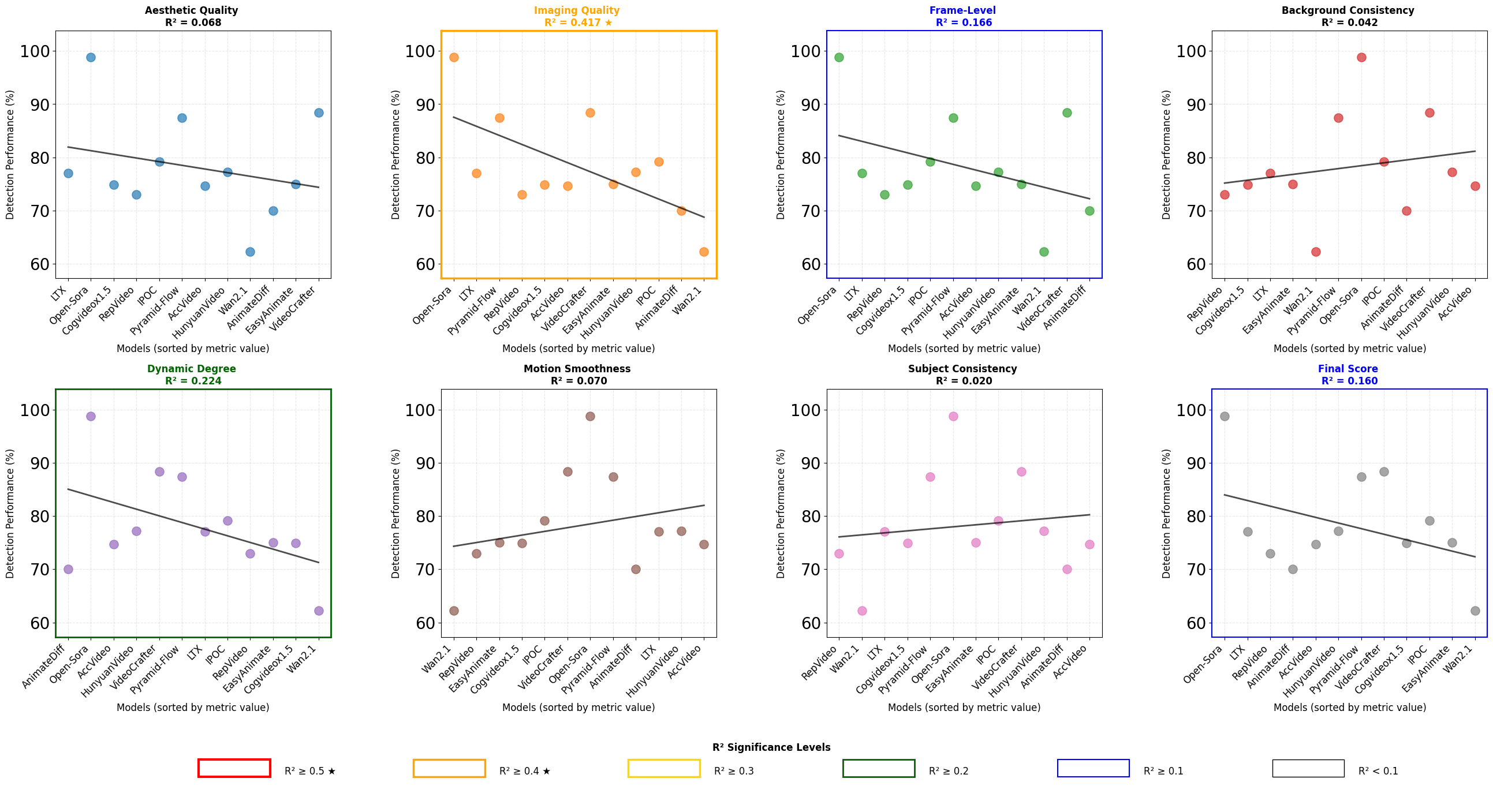}  
    \caption{Text-to-Video (T2V) Task.}
  \end{subfigure}
  
  \vspace{0.5cm}
  
  \begin{subfigure}{0.92\linewidth}
    \centering
    \includegraphics[width=\linewidth]{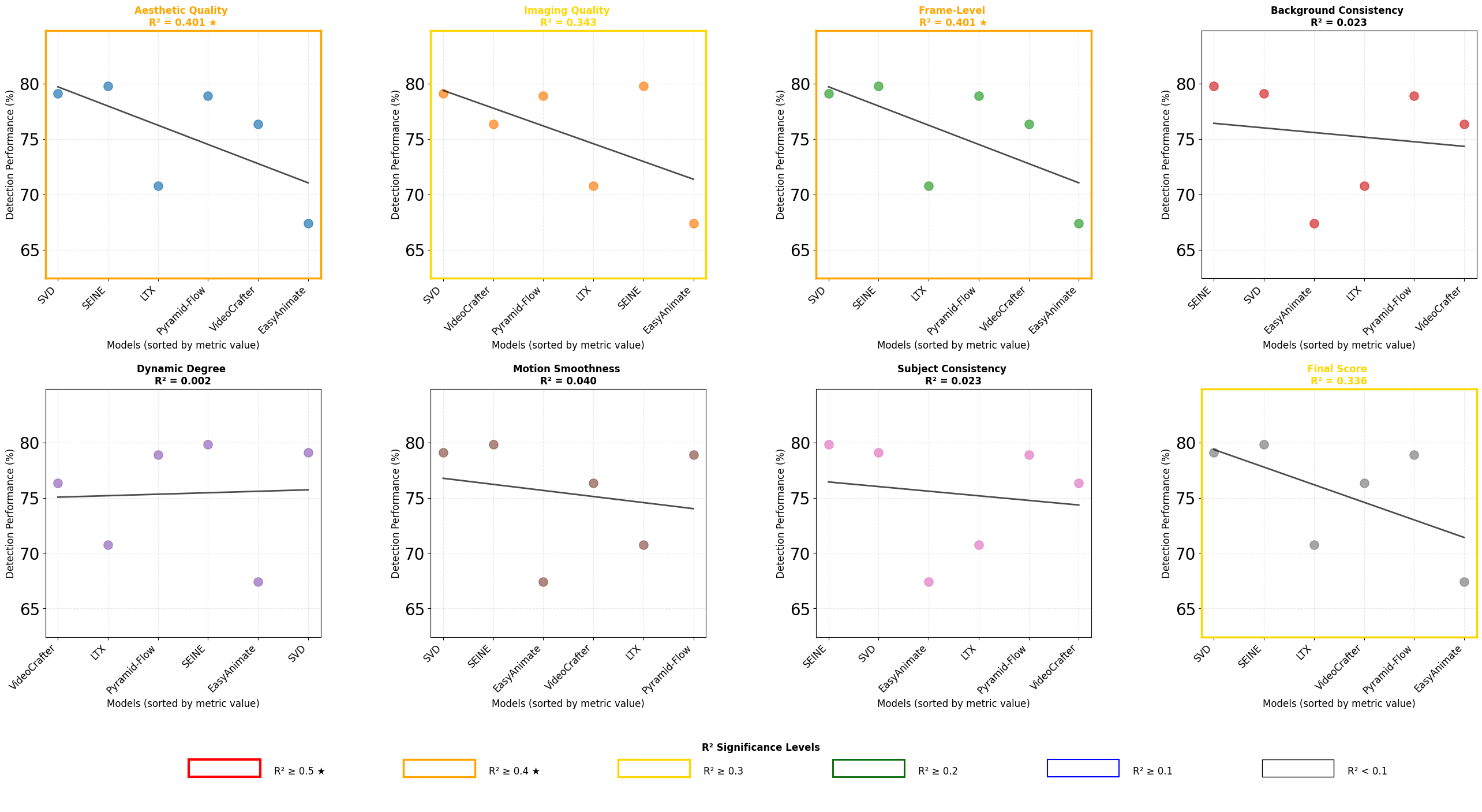}
    \caption{Image-to-Video (I2V) Task.}
  \end{subfigure}
  
  \caption{Correlations between the performance of video classification models and quality metrics of video generation models. For each scatter plot, the metric value increases along the x-axis. The coefficient of determination ($R^2$) quantifies the strength of the linear relationship; a value closer to 1 indicates that the quality metric effectively explains the variance in detection performance (AUC).}
  \label{fig:correlation_video_classification} 
\end{figure*}

\begin{figure*}[t]
  \centering
  \vspace{1cm}
  \begin{subfigure}{0.92\linewidth}
    \centering
    \includegraphics[width=\linewidth]{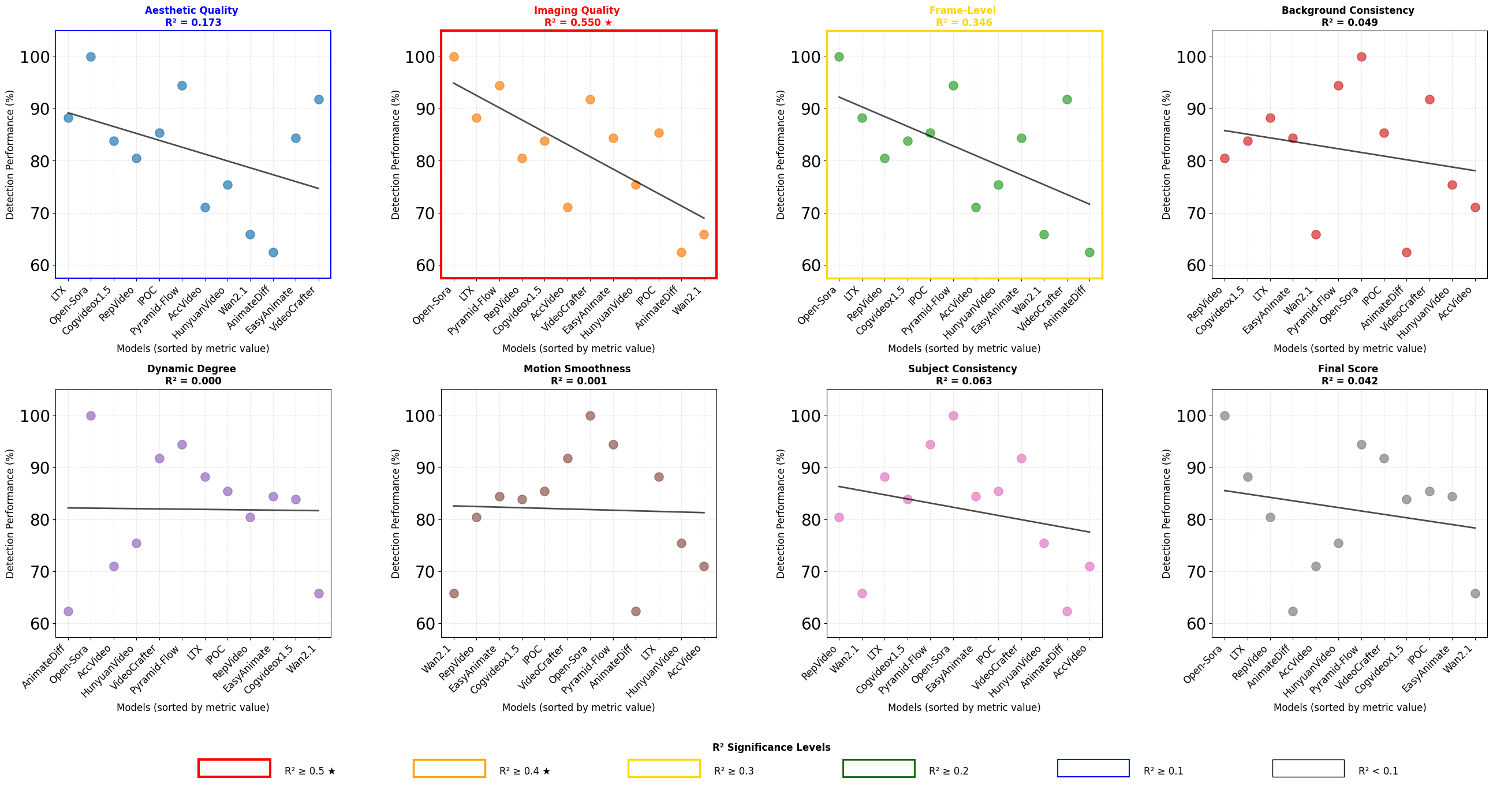}
    \caption{Text-to-Video (T2V) Task.}
  \end{subfigure}
  
  \vspace{0.5cm}
  
  \begin{subfigure}{0.92\linewidth}
    \centering
\includegraphics[width=\linewidth]{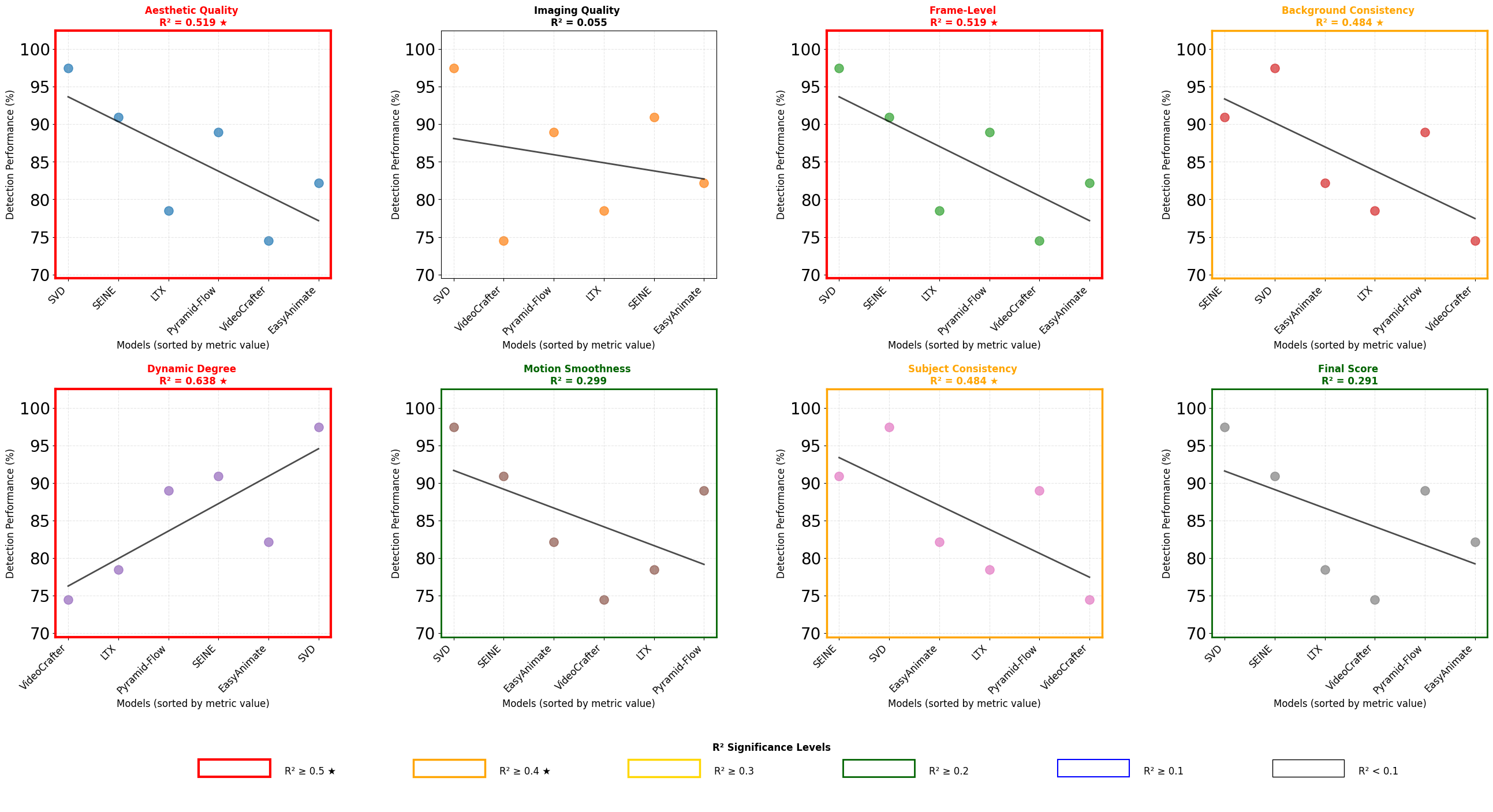}
    \caption{Image-to-Video (I2V) Task.}
  \end{subfigure}
  \caption{Correlations between the performance of AI-generated image detection Models and quality metrics of video generation models. For each scatter plot, the metric value increases along the x-axis. The coefficient of determination ($R^2$) quantifies the strength of the linear relationship; a value closer to 1 indicates that the quality metric effectively explains the variance in detection performance (AUC).}
  \label{fig:correlation_AGID}
\end{figure*}

\begin{figure*}[t]
  \centering
  \vspace{1cm}
  \begin{subfigure}{0.92\linewidth}
    \centering
    \includegraphics[width=\linewidth]{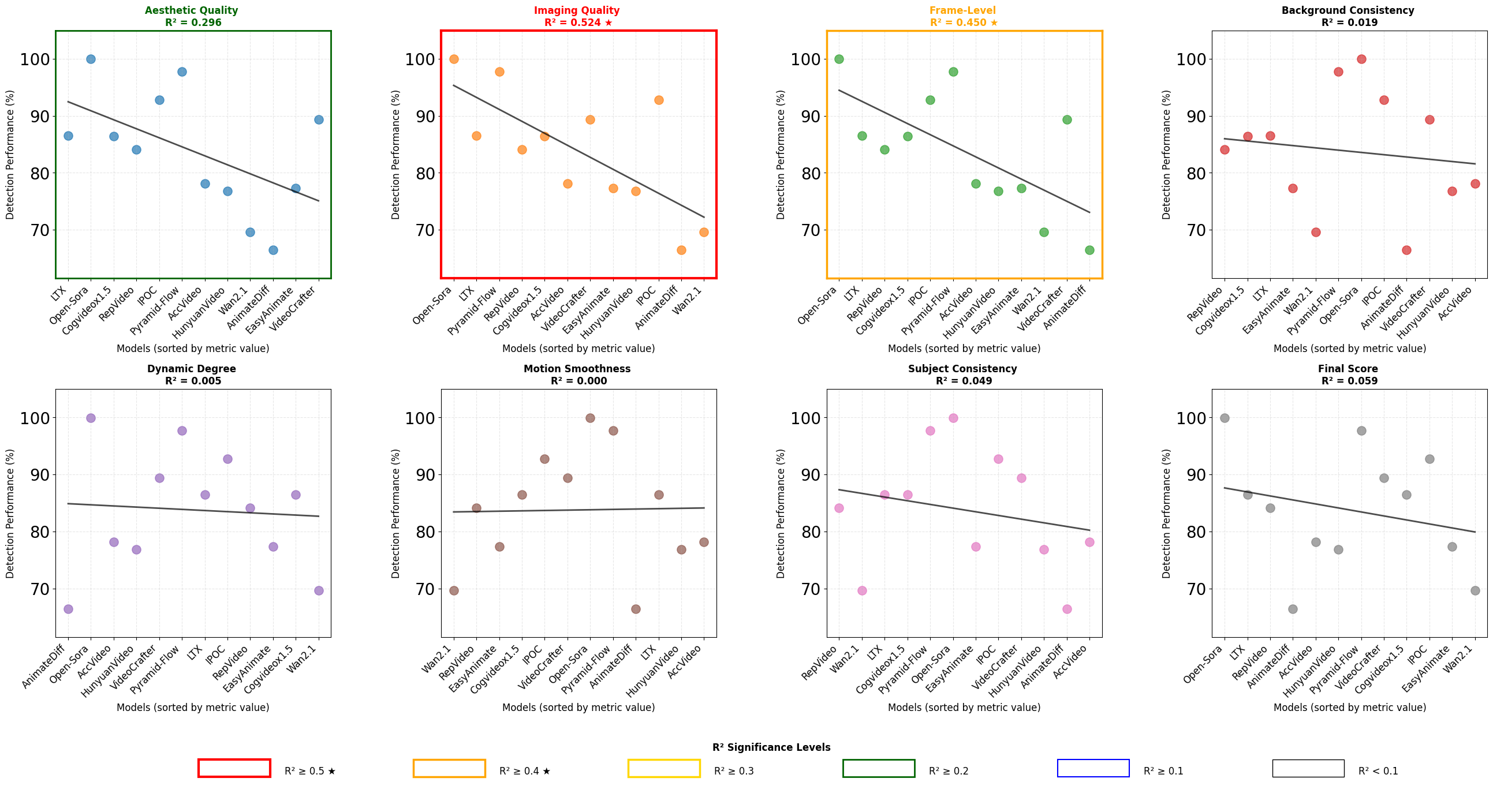}
    \caption{Text-to-Video (T2V) Task.}
  \end{subfigure}
  
  \vspace{0.5cm}
  
  \begin{subfigure}{0.92\linewidth}
    \centering
    \includegraphics[width=\linewidth]{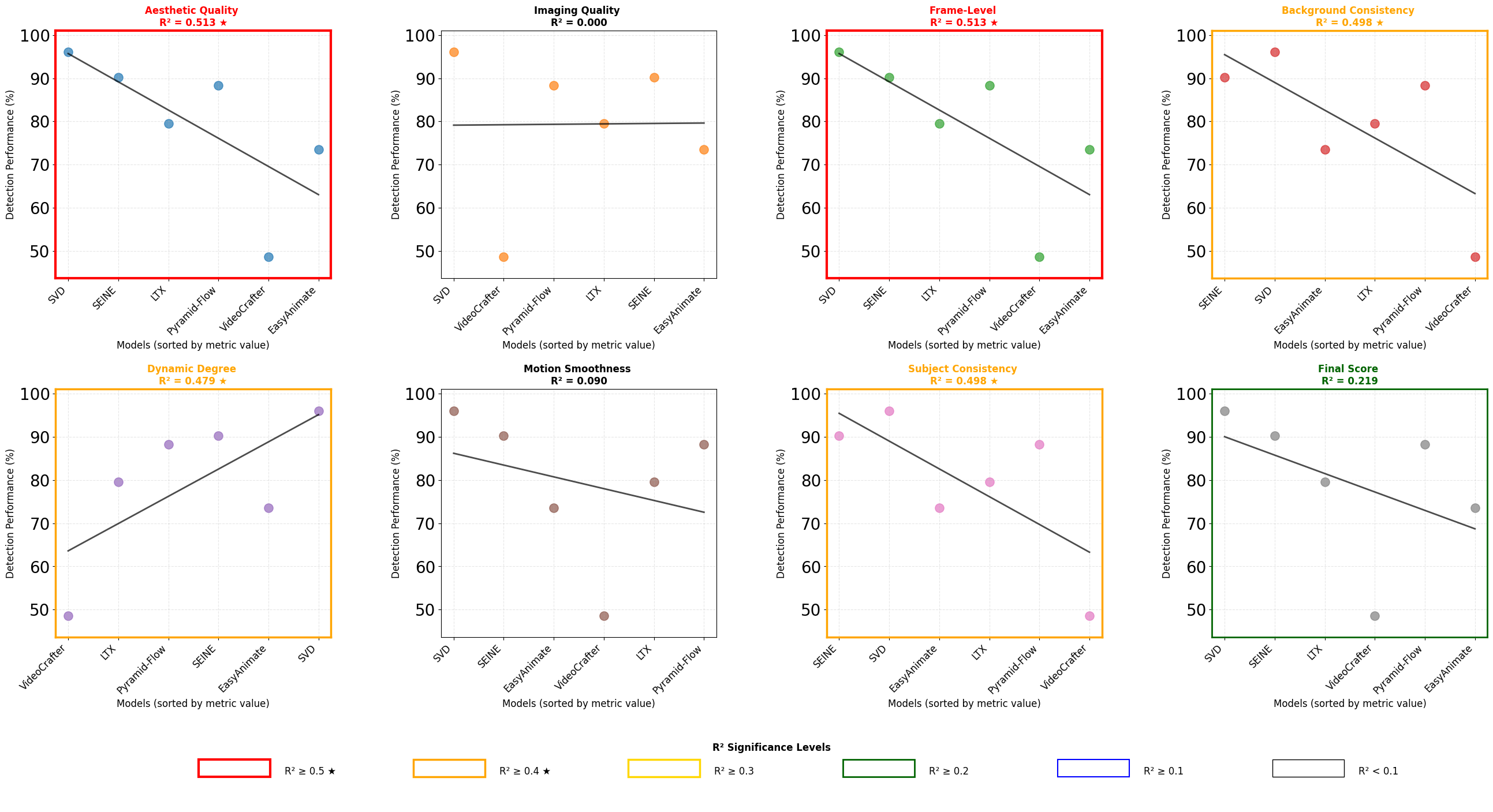}
    \caption{Image-to-Video (I2V) Task.}
  \end{subfigure}
  
  \caption{Correlations between the performance of AI-generated video detection models and quality metrics of video generation models. For each scatter plot, the metric value increases along the x-axis. The coefficient of determination ($R^2$) quantifies the strength of the linear relationship; a value closer to 1 indicates that the quality metric effectively explains the variance in detection performance (AUC).}
  \label{fig:correlation_AGVD}
\end{figure*}

\begin{figure*}[t]
  \centering
  \vspace{1cm}
  \begin{subfigure}{0.92\linewidth}
    \centering
    \includegraphics[width=\linewidth]{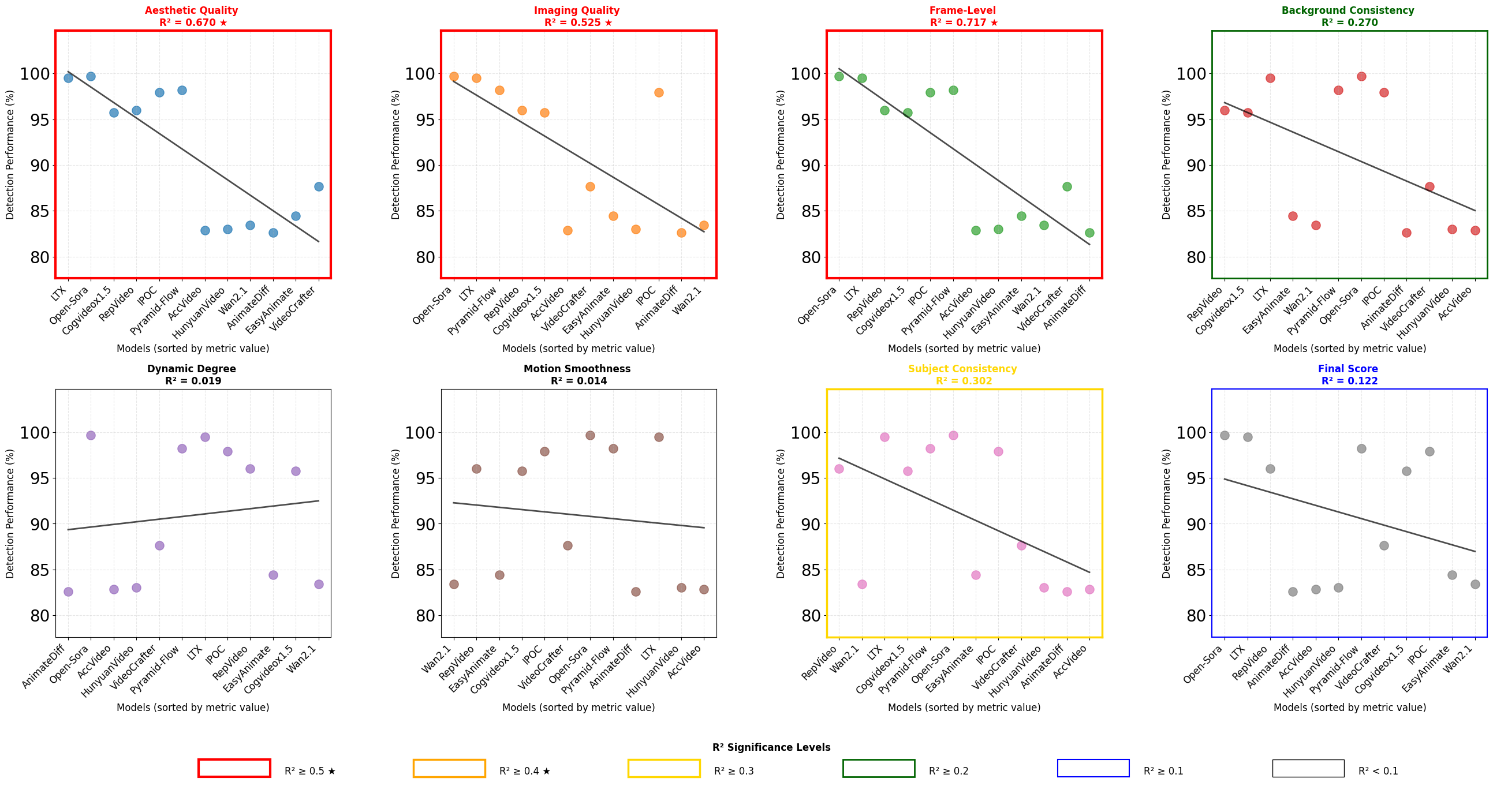}
    \caption{Trained on LTX.}
  \end{subfigure}
  
  \vspace{1cm}
  
  \begin{subfigure}{0.92\linewidth}
    \centering
    \includegraphics[width=\linewidth]{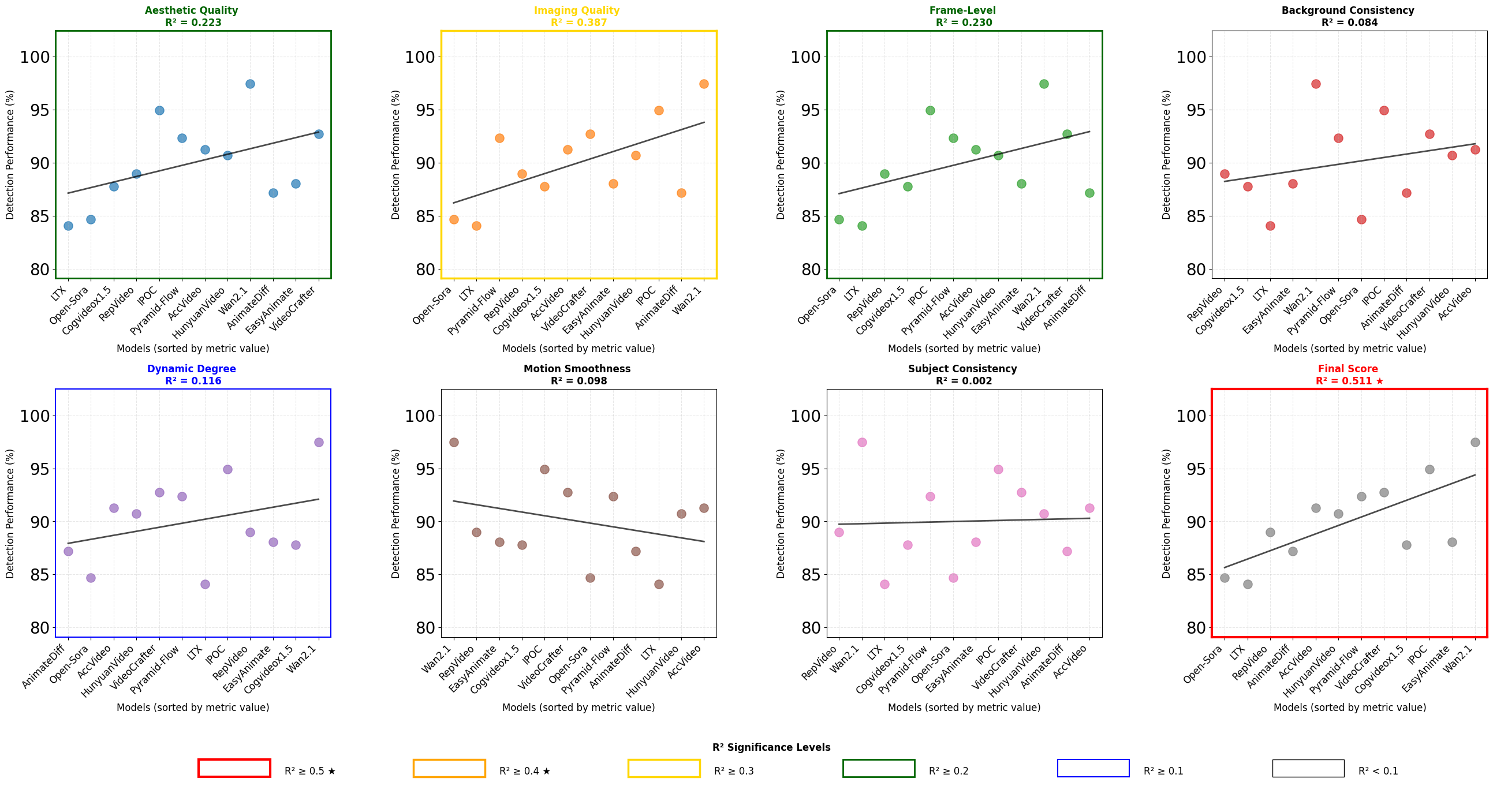}
    \caption{Trained on Wan2.1.}
  \end{subfigure}
  
  \caption{Correlations between the performance of DeCoF and quality metrics of video generation models. For each scatter plot, the metric value increases along the x-axis. The coefficient of determination ($R^2$) quantifies the strength of the linear relationship; a value closer to 1 indicates that the quality metric effectively explains the variance in detection performance (AUC).}
  \label{fig:correlation_DeCoF_trained}
\end{figure*}

\begin{figure*}[t]
  \centering
  \vspace{1cm}
  \begin{subfigure}{0.92\linewidth}
    \centering
    \includegraphics[width=\linewidth]{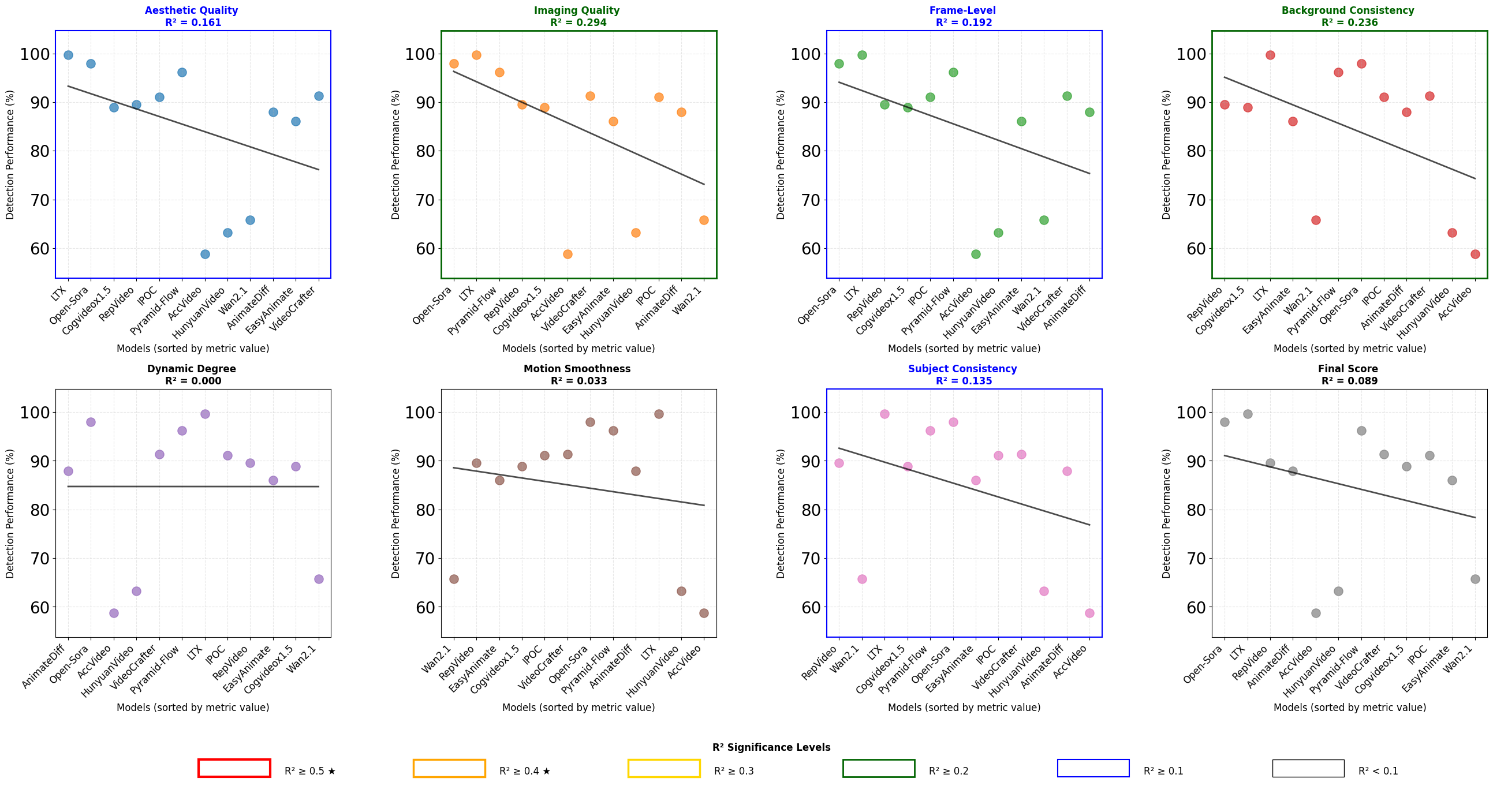}
    \caption{Trained on LTX.}
  \end{subfigure}
  
  \vspace{1cm}
  
  \begin{subfigure}{0.92\linewidth}
    \centering
    \includegraphics[width=\linewidth]{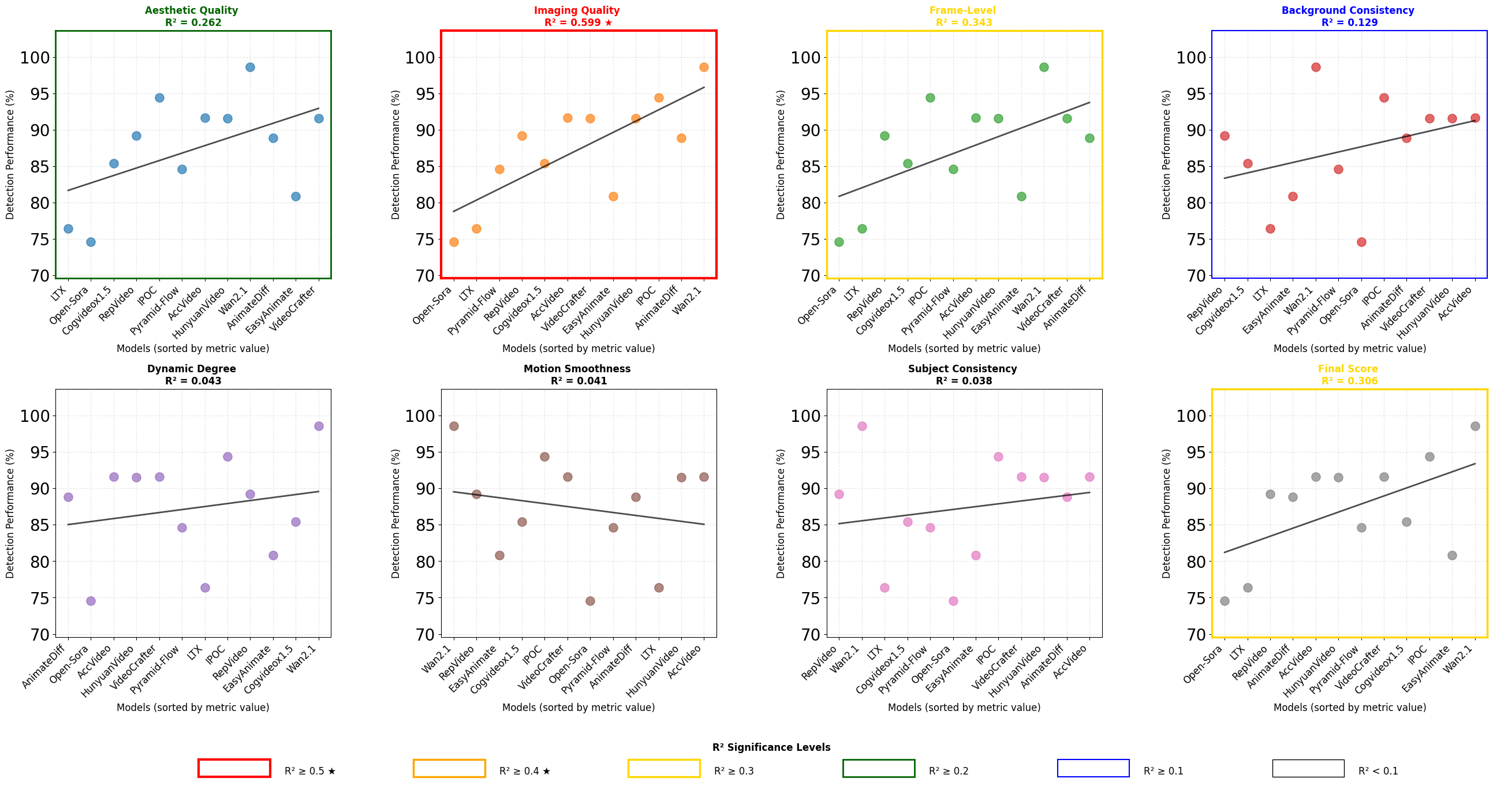}
    \caption{Trained on Wan2.1.}
  \end{subfigure}
  
  \caption{Correlations between the performance of UnivFD and quality metrics of video generation models. For each scatter plot, the metric value increases along the x-axis. The coefficient of determination ($R^2$) quantifies the strength of the linear relationship; a value closer to 1 indicates that the quality metric effectively explains the variance in detection performance (AUC).}
  \label{fig:correlation_UnivFD_trained}
\end{figure*}

\begin{figure*}[t]
  \centering
  \vspace{1cm}
  \begin{subfigure}{0.92\linewidth}
    \centering
    \includegraphics[width=\linewidth]{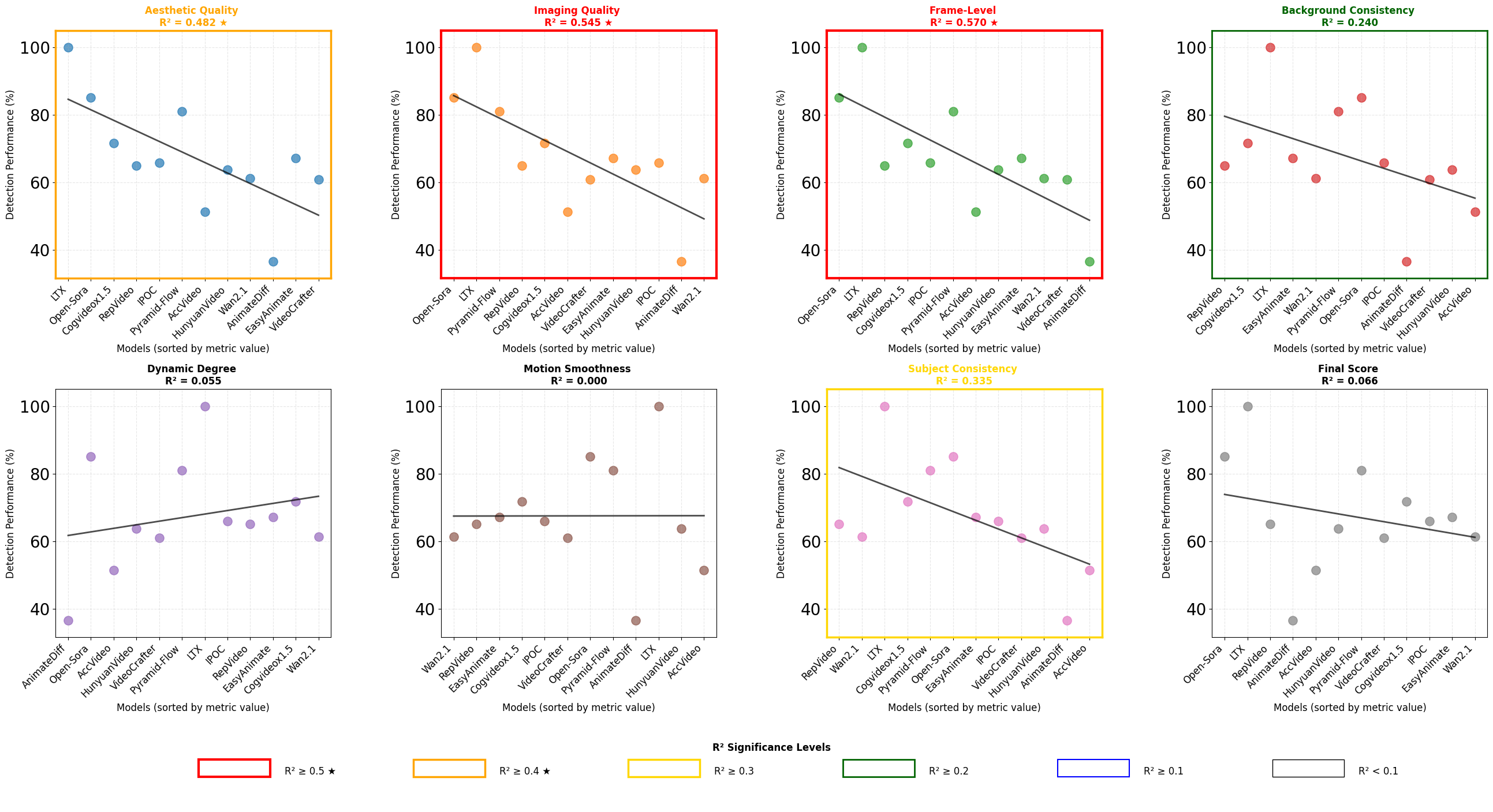}
    \caption{Trained on LTX.}
  \end{subfigure}
  
  \vspace{1cm}
  
  \begin{subfigure}{0.92\linewidth}
    \centering
    \includegraphics[width=\linewidth]{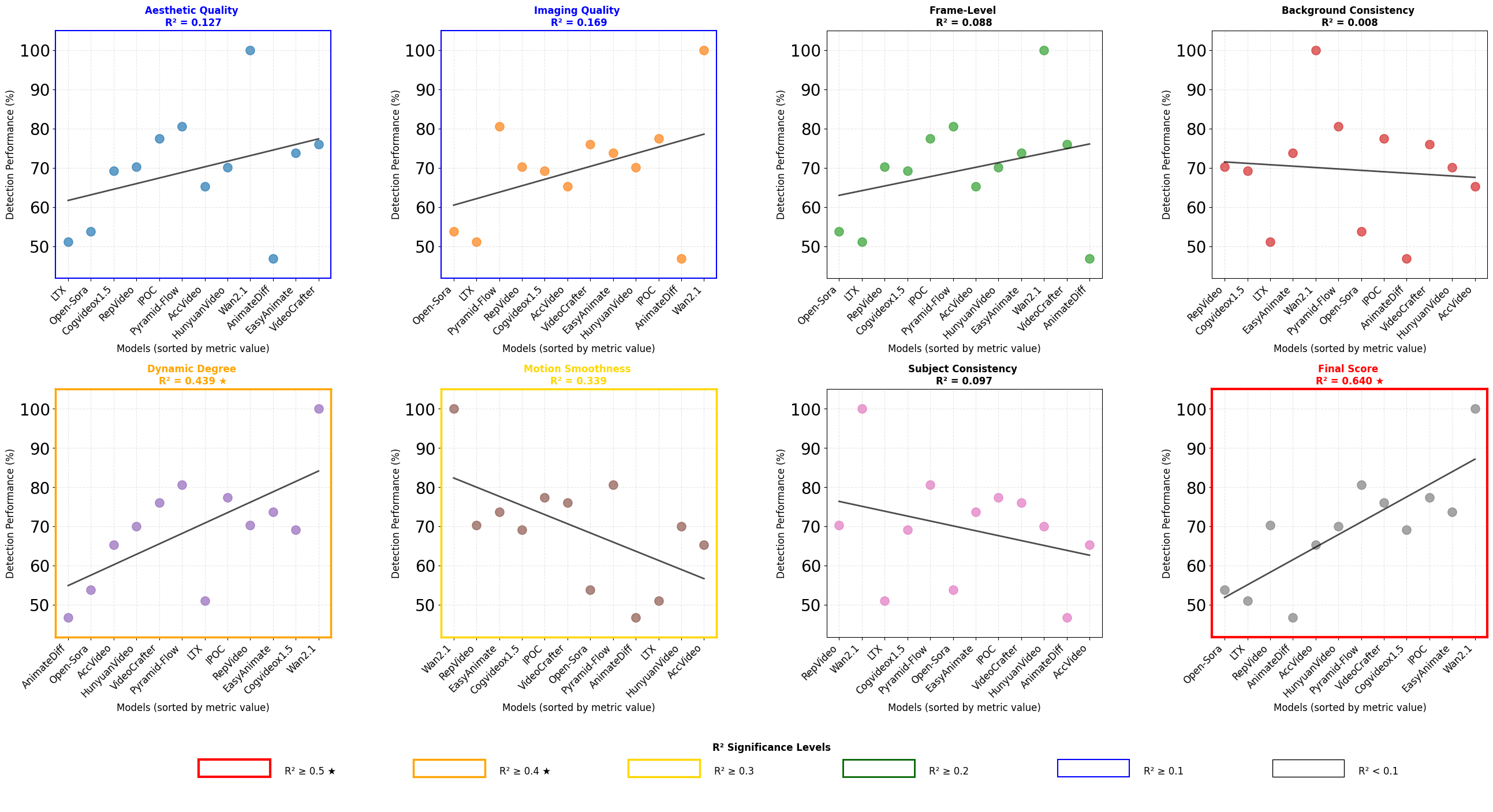}
    \caption{Trained on Wan2.1.}
  \end{subfigure}
  
  \caption{Correlations between the performance of I3D and quality metrics of video generation models. For each scatter plot, the metric value increases along the x-axis. The coefficient of determination ($R^2$) quantifies the strength of the linear relationship; a value closer to 1 indicates that the quality metric effectively explains the variance in detection performance (AUC).}
  \label{fig:correlation_I3D_trained}
\end{figure*}

\begin{figure*}[t]
  \centering
  \vspace{1cm}
  \begin{subfigure}{0.92\linewidth}
    \centering
    \includegraphics[width=\linewidth]{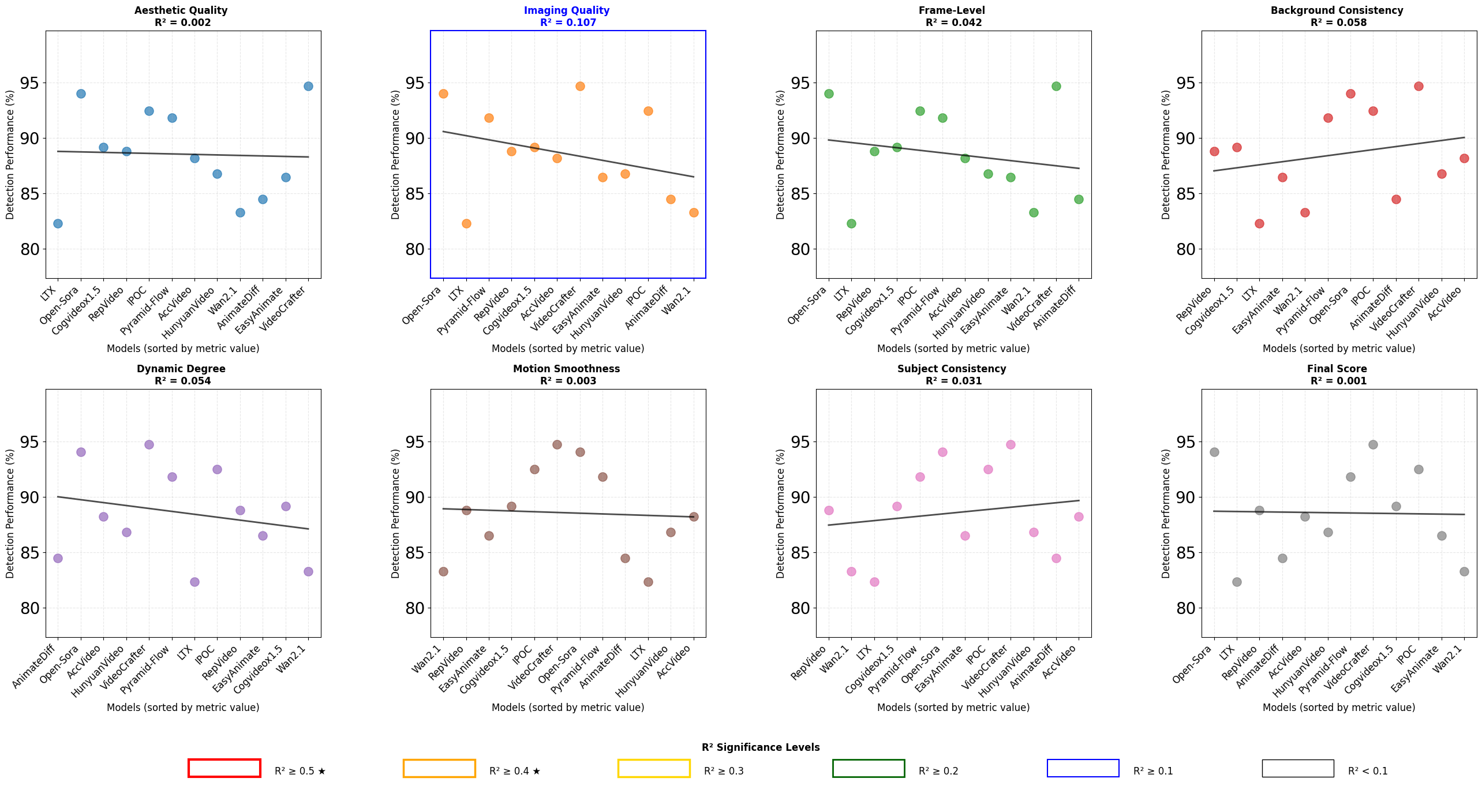}
    \caption{Higher Generative Quality Does Not Increase Detection Difficulty.}
    \label{deocf_test}
  \end{subfigure}
  
  \vspace{1cm}
  
  \begin{subfigure}{0.92\linewidth}
    \centering
    \includegraphics[width=\linewidth]{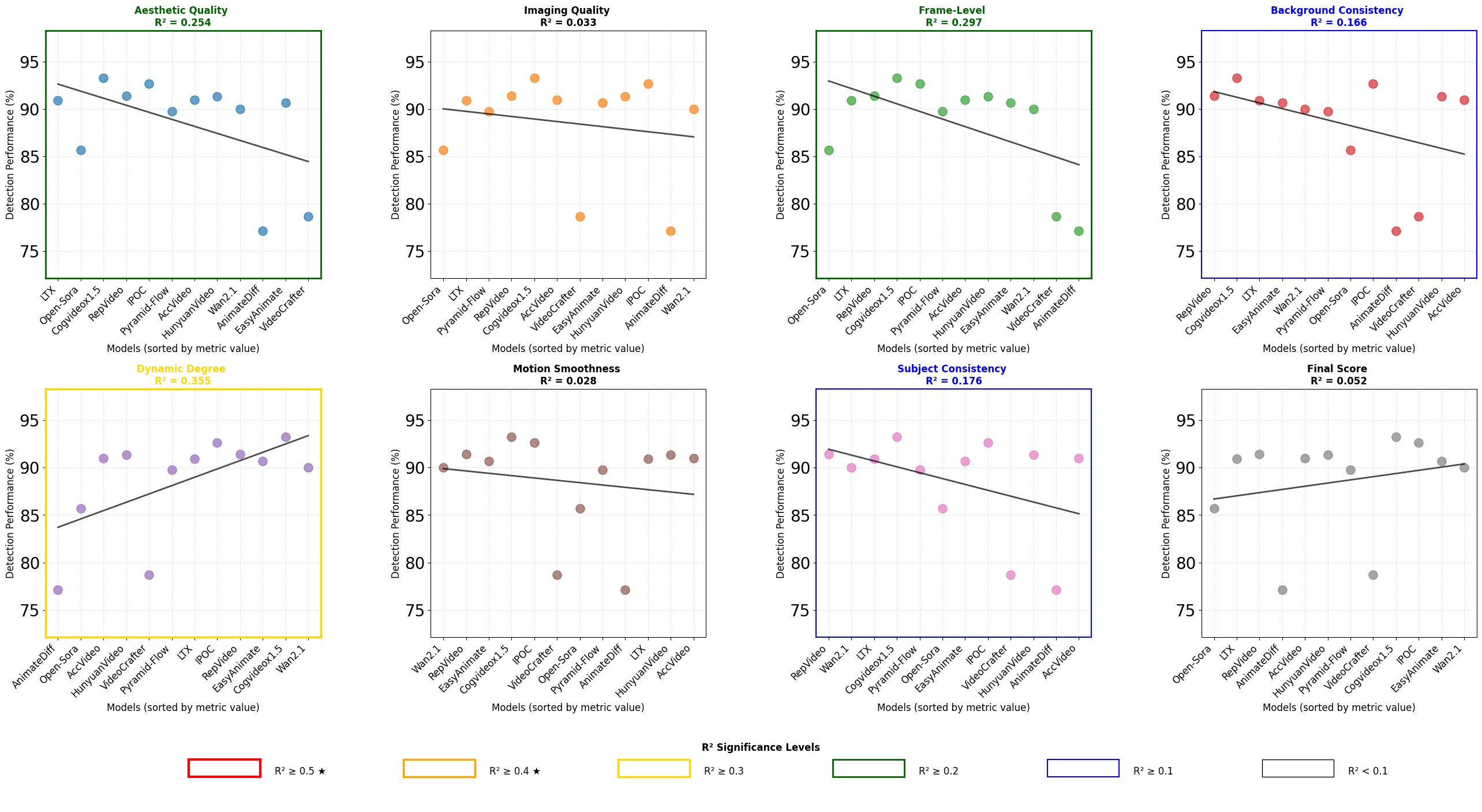}
    \caption{Better Generative Models Do Not Yield Better Detectors.}
    \label{deocf_train}
  \end{subfigure}
    
  \caption{Validation Through DeCoF: On the Relationship between Generative Model Quality and Detection Efficacy.}
\end{figure*}

\begin{figure*}[t]
  \centering
  \vspace{1cm}
  \begin{subfigure}{0.92\linewidth}
    \centering
    \includegraphics[width=\linewidth]{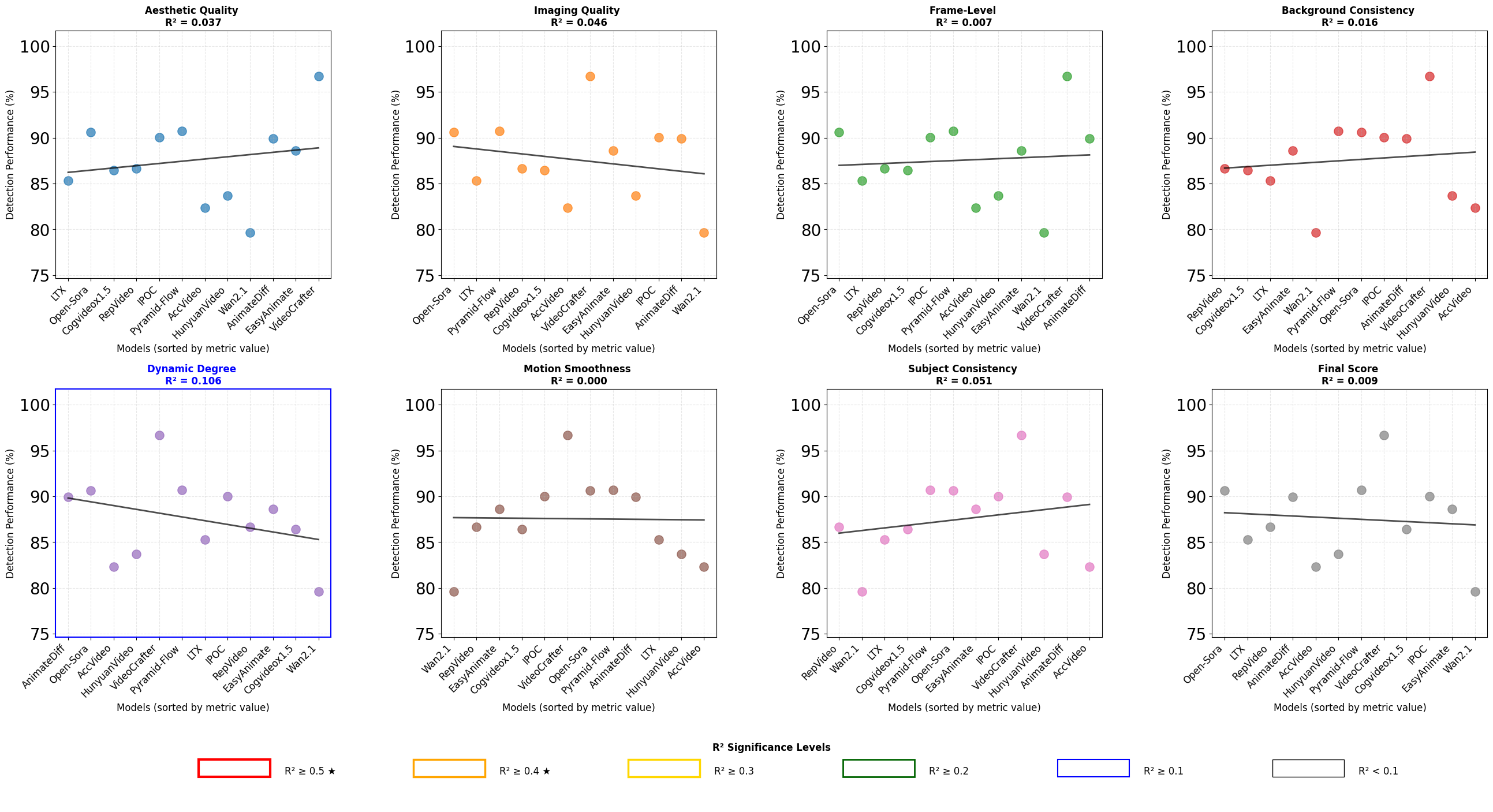}
    \caption{Higher Generative Quality Does Not Increase Detection Difficulty.}
    \label{uni_test}
  \end{subfigure}
  
  \vspace{1.2cm}
  
  \begin{subfigure}{0.92\linewidth}
    \centering
    \includegraphics[width=\linewidth]{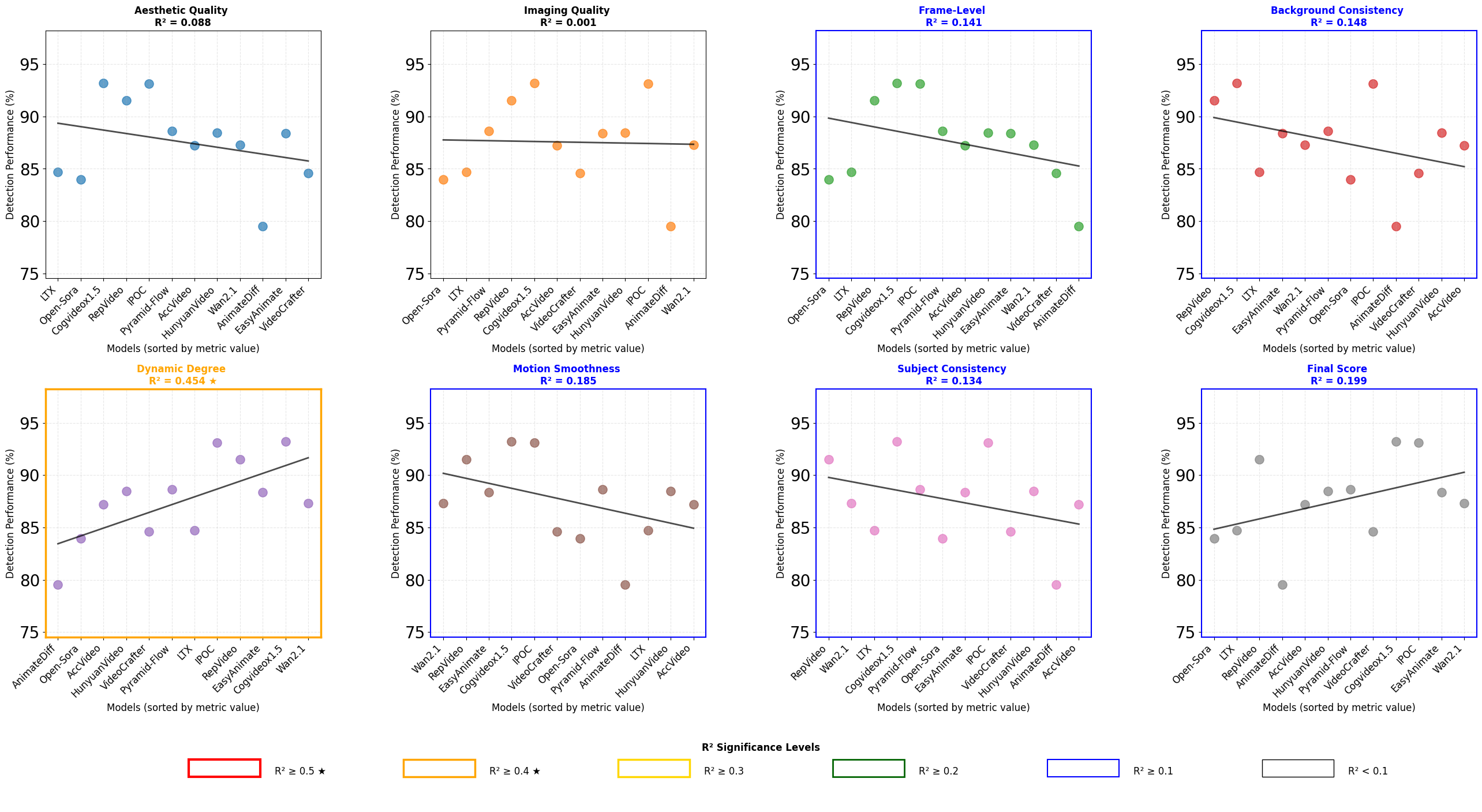}
    \caption{Better Generative Models Do Not Yield Better Detectors.}
    \label{uni_train}
  \end{subfigure}
     
  \caption{Validation Through UnivFD: On the Relationship between Generative Model Quality and Detection Efficacy.}
\end{figure*}

\begin{figure*}[t]
  \centering
  \vspace{1cm}
  \begin{subfigure}{0.92\linewidth}
    \centering
    \includegraphics[width=\linewidth]{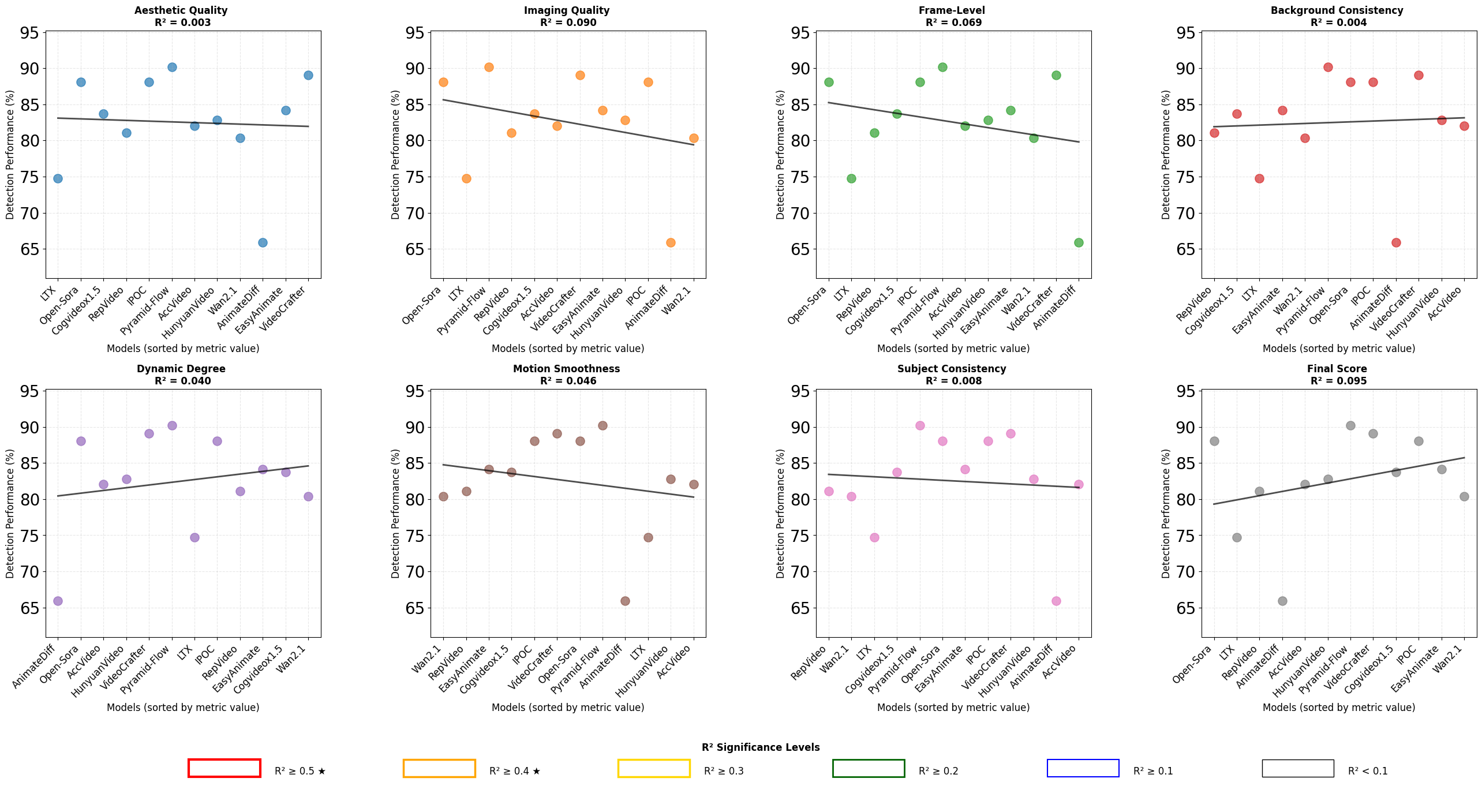}
    \caption{Higher Generative Quality Does Not Increase Detection Difficulty.}
    \label{i3d_test}
  \end{subfigure}
  
  \vspace{1.2cm}
  
  \begin{subfigure}{0.92\linewidth}
    \centering
    \includegraphics[width=\linewidth]{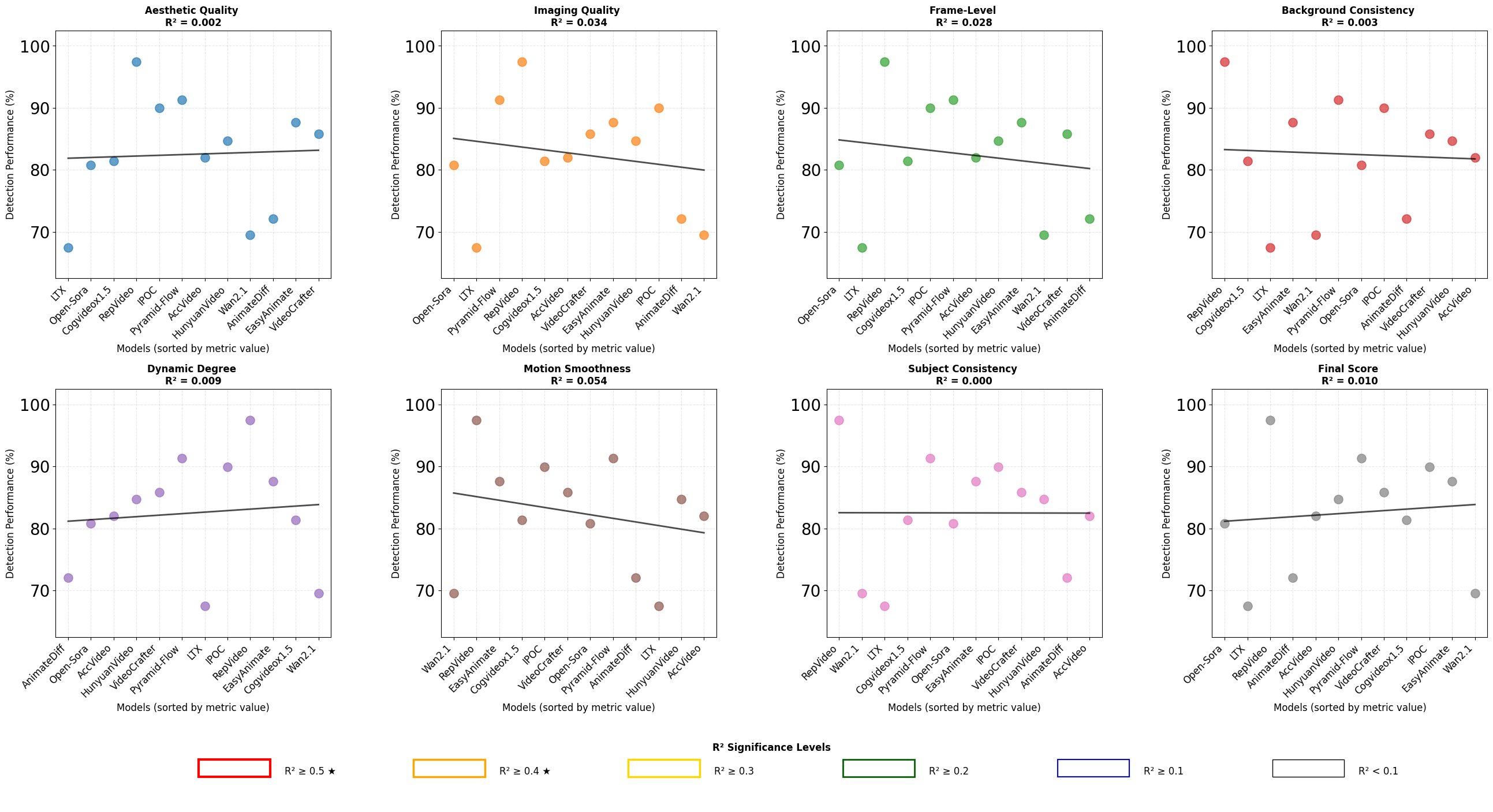}
    \caption{Better Generative Models Do Not Yield Better Detectors.}
    \label{i3d_train}
  \end{subfigure}
   
  \caption{Validation Through I3D: On the Relationship between Generative Model Quality and Detection Efficacy.}
\end{figure*}

\begin{table*}[tbp]
\centering
\small
\caption{Impact of Guidance Scale and Sampling Steps on Detectors.}
\label{tab:guidance and steps}
 \resizebox{\textwidth}{!}{
\begin{tabular}{@{}c| l *{7}{c} *{5}{c}@{}}
\toprule
\multirow{3}{*}{\makecell{Detector\\Type}} & \multirow{3}{*}{Model} & \multicolumn{7}{c}{Guidance Scale} & \multicolumn{5}{c}{Sampling Steps} \\
\cmidrule(lr){3-9} \cmidrule(l){10-14}
 & & 1.0 & 2.0 & 4.0 & 6.0 & 8.0 & 10.0 & 12.0 & 30 & 40 & 50 & 60 & 70 \\
\midrule

\multirow{12}{*}{\rotatebox[origin=c]{90}{DecoF}}
& AccVideo & 43.98 & 70.88 & 89.71 & 93.47 & 95.41 & 96.75 & 97.53 & 89.23 & 86.86 & 93.47 & 87.24 & 88.54 \\
& AnimateDiff & 85.47 & 87.39 & 85.77 & 83.85 & 80.22 & 78.68 & 77.45 & 56.37 & 59.88 & 83.85 & 77.02 & 72.74 \\
& Cogvideox1.5 & 56.02 & 69.07 & 82.29 & 88.22 & 91.49 & 92.43 & 94.29 & 97.06 & 94.41 & 88.22 & 94.36 & 91.93 \\
& EasyAnimate & 77.27 & 92.78 & 96.85 & 98.90 & 97.80 & 98.45 & 98.66 & 85.22 & 87.54 & 98.90 & 91.67 & 92.24 \\
& HunyuanVideo & 42.79 & 68.91 & 88.60 & 92.14 & 94.64 & 95.01 & 95.88 & 87.03 & 82.24 & 92.14 & 83.12 & 82.78 \\
& IPOC & 35.17 & 49.03 & 72.22 & 82.30 & 88.21 & 91.28 & 94.54 & 95.99 & 93.89 & 82.30 & 91.77 & 90.37 \\
& LTX & 82.82 & 74.33 & 79.71 & 84.24 & 85.93 & 86.57 & 89.28 & 97.38 & 94.71 & 84.24 & 94.57 & 92.62 \\
& Open-Sora & 28.27 & 47.59 & 67.50 & 73.96 & 79.80 & 81.02 & 82.91 & 87.82 & 84.25 & 73.96 & 81.40 & 74.53 \\
& Pyramid-Flow & 32.08 & 58.78 & 80.40 & 88.23 & 91.89 & 93.28 & 95.64 & 97.41 & 94.59 & 88.23 & 93.47 & 92.68 \\
& RepVideo & 41.89 & 48.94 & 69.62 & 81.91 & 88.51 & 90.93 & 94.53 & 96.81 & 93.04 & 81.91 & 93.48 & 91.67 \\
& VideoCrafter & 66.36 & 71.94 & 77.25 & 80.46 & 80.57 & 82.72 & 85.78 & 50.39 & 53.68 & 80.46 & 67.68 & 71.87 \\
& Wan2.1 & 45.05 & 65.85 & 82.90 & 86.18 & 88.55 & 88.69 & 90.14 & 91.03 & 86.71 & 86.18 & 87.06 & 83.39 \\
\addlinespace
\midrule
\multirow{12}{*}{\rotatebox[origin=c]{90}{UnivFD}}
& AccVideo & 56.50 & 69.46 & 85.74 & 88.69 & 90.17 & 92.04 & 93.39 & 82.37 & 81.42 & 88.69 & 83.28 & 83.09 \\
& AnimateDiff & 92.74 & 90.49 & 91.15 & 90.87 & 89.87 & 89.71 & 89.02 & 76.27 & 78.96 & 90.87 & 88.88 & 88.73 \\
& Cogvideox1.5 & 73.42 & 76.82 & 84.54 & 88.14 & 90.40 & 91.01 & 92.85 & 96.92 & 94.71 & 88.14 & 95.42 & 92.56 \\
& EasyAnimate & 95.99 & 98.68 & 99.42 & 99.75 & 99.68 & 99.88 & 99.80 & 92.60 & 95.11 & 99.75 & 96.35 & 97.54 \\
& HunyuanVideo & 49.68 & 65.71 & 82.98 & 85.85 & 88.35 & 89.34 & 90.32 & 78.45 & 76.43 & 85.85 & 79.47 & 78.42 \\
& IPOC & 50.73 & 55.48 & 72.31 & 80.58 & 86.17 & 88.41 & 91.07 & 96.21 & 94.10 & 80.58 & 93.84 & 91.34 \\
& LTX & 96.68 & 87.23 & 84.79 & 84.97 & 86.86 & 86.66 & 89.73 & 97.50 & 95.58 & 84.97 & 96.57 & 95.16 \\
& Open-Sora & 64.33 & 74.08 & 83.49 & 86.73 & 89.29 & 88.78 & 90.42 & 95.48 & 93.83 & 86.73 & 95.58 & 91.97 \\
& Pyramid-Flow & 69.32 & 81.52 & 91.10 & 94.43 & 95.30 & 96.14 & 97.43 & 98.50 & 97.83 & 94.43 & 97.99 & 97.50 \\
& RepVideo & 66.19 & 62.25 & 72.21 & 80.38 & 85.06 & 88.27 & 93.19 & 96.79 & 93.53 & 80.38 & 94.92 & 92.58 \\
& VideoCrafter & 88.24 & 89.43 & 91.16 & 94.10 & 94.07 & 95.38 & 96.60 & 73.49 & 77.67 & 94.10 & 85.74 & 90.99 \\
& Wan2.1 & 43.77 & 58.53 & 76.07 & 77.29 & 78.28 & 79.60 & 79.18 & 86.12 & 81.06 & 77.29 & 82.62 & 78.95 \\
\addlinespace
\midrule
\multirow{12}{*}{\rotatebox[origin=c]{90}{I3D}}
& AccVideo & 39.54 & 54.15 & 76.70 & 83.42 & 86.70 & 89.67 & 91.34 & 71.97 & 76.99 & 83.42 & 74.36 & 67.67 \\
& AnimateDiff & 69.20 & 70.95 & 70.16 & 64.59 & 60.29 & 55.97 & 52.07 & 26.92 & 34.62 & 64.59 & 42.48 & 40.14 \\
& Cogvideox1.5 & 55.68 & 68.05 & 77.14 & 84.66 & 86.66 & 88.17 & 89.67 & 72.01 & 77.44 & 84.66 & 70.68 & 73.87 \\
& EasyAnimate & 100.00 & 100.00 & 100.00 & 100.00 & 100.00 & 100.00 & 100.00 & 99.48 & 100.00 & 100.00 & 99.29 & 98.58 \\
& HunyuanVideo & 46.15 & 64.96 & 81.80 & 86.25 & 91.68 & 93.57 & 94.88 & 75.47 & 78.36 & 86.25 & 77.88 & 75.51 \\
& IPOC & 47.18 & 68.74 & 86.54 & 92.42 & 96.01 & 96.83 & 98.08 & 71.87 & 82.57 & 92.42 & 73.39 & 78.98 \\
& LTX & 32.05 & 48.50 & 62.45 & 68.13 & 70.93 & 71.99 & 76.81 & 46.48 & 52.51 & 68.13 & 48.78 & 53.19 \\
& Open-Sora & 36.68 & 58.42 & 74.56 & 77.22 & 83.46 & 81.97 & 85.77 & 70.80 & 81.14 & 77.22 & 71.33 & 73.96 \\
& Pyramid-Flow & 63.05 & 83.29 & 94.91 & 97.27 & 98.75 & 99.22 & 99.61 & 91.34 & 96.40 & 97.27 & 90.00 & 94.37 \\
& RepVideo & 58.18 & 75.84 & 92.23 & 96.83 & 98.58 & 98.99 & 99.47 & 97.05 & 97.49 & 96.83 & 91.28 & 93.93 \\
& VideoCrafter & 38.38 & 49.73 & 72.57 & 77.96 & 82.64 & 84.35 & 83.14 & 43.55 & 52.15 & 77.96 & 54.23 & 59.96 \\
& Wan2.1 & 29.23 & 43.01 & 62.87 & 71.71 & 78.49 & 81.10 & 83.38 & 49.34 & 48.98 & 71.71 & 49.10 & 54.26 \\

\bottomrule
\end{tabular}}

\end{table*}

\begin{table*}[t]
\centering
\caption{Investigation of Video Quality Assessment: Impact of Guidance Scale and Sampling Steps.}
\label{tab:guidance stpes quality}

\begin{tabular}{@{}c c c c c c c c c c@{}}
\toprule
\multirow{3}{*}{Parameter} & \multirow{3}{*}{Value} & \multicolumn{6}{c}{Quality Metrics} & \multirow{3}{*}{\makecell{Quality\\Score}} & \multirow{3}{*}{\makecell{Final\\Score}} \\
\cmidrule(lr){3-8}
 & & \makecell{Aesthetic\\Quality} & \makecell{Background\\Consistency} & \makecell{Dynamic\\Degree} & \makecell{Imaging\\Quality} & \makecell{Motion\\Smoothness} & \makecell{Subject\\Consistency} & & \\
\midrule

\multirow{7}{*}{\rotatebox[origin=c]{90}{Guidance Scale}}
& 1.0 & 57.35 & 96.12 & 12.50 & 61.84 & 95.72 & 96.02 & 76.28 & 76.28 \\
& 2.0 & 61.18 & 97.07 & 18.50 & 64.66 & 95.59 & 97.31 & 78.97 & 78.97 \\
& 4.0 & 61.02 & 97.47 & 23.50 & 63.35 & 95.33 & 97.70 & 79.70 & 79.70 \\
& 6.0 & 60.88 & 97.49 & 20.75 & 63.12 & 94.75 & 97.65 & 79.03 & 79.03 \\
& 8.0 & 60.79 & 97.33 & 22.75 & 61.88 & 94.08 & 97.45 & 78.96 & 78.96 \\
& 10.0 & 60.22 & 97.13 & 23.75 & 60.55 & 93.59 & 97.06 & 78.60 & 78.60 \\
& 12.0 & 59.60 & 97.09 & 23.25 & 59.60 & 93.67 & 97.11 & 78.24 & 78.24 \\
\addlinespace
\midrule
\multirow{5}{*}{\rotatebox[origin=c]{90}{\small Sampling Steps}}
& 30 & 45.79 & 94.20 & 14.75 & 53.52 & 79.81 & 91.59 & 69.03 & 69.03 \\
& 40 & 47.99 & 95.16 & 14.75 & 55.66 & 82.37 & 93.10 & 70.73 & 70.73 \\
& 50 & 60.88 & 97.49 & 20.75 & 63.12 & 94.75 & 97.65 & 79.03 & 79.03 \\
& 60 & 49.50 & 93.97 & 9.75 & 47.79 & 88.99 & 92.46 & 69.54 & 69.54 \\
& 70 & 54.96 & 93.93 & 23.00 & 59.75 & 88.35 & 93.15 & 75.12 & 75.12 \\

\bottomrule
\end{tabular}

\end{table*}
\begin{table*}[tbp]
\centering
\caption{Performance Comparison of Detection Models Across Categories.}
\label{tab:category_performance}
\scriptsize
\resizebox{\textwidth}{!}{
\begin{tabular}{@{}c *{15}{c}@{}}
\toprule
\multirow{2}{*}{Detector} & \multicolumn{5}{c}{LTX} & \multicolumn{5}{c}{Pyramid-Flow} & \multicolumn{5}{c}{Wan2.1} \\
\cmidrule(lr){2-6} \cmidrule(lr){7-11} \cmidrule(lr){12-16}
 & Animals & Food/Bev. & Illust. & Vehicles & Range. & Animals & Food/Bev. & Illust. & Vehicles & Range. & Animals & Food/Bev. & Illust. & Vehicles & Range. \\
\midrule
I3D & 91.70 & 97.33 & 94.74 & 92.72 & 5.63 & 96.36 & 99.21 & 97.65 & 94.34 & 4.87 & 54.34 & 59.41 & 68.91 & 66.03 & 14.57 \\
DeCoF & 86.91 & 90.40 & 85.41 & 83.51 & 6.89 & 98.36 & 96.84 & 97.85 & 96.68 & 1.68 & 64.30 & 77.56 & 74.72 & 70.70 & 13.26 \\
CNNSpot & 99.28 & 99.83 & 99.36 & 99.39 & 0.55 & 94.75 & 97.18 & 96.69 & 95.89 & 2.43 & 59.41 & 73.10 & 75.65 & 69.16 & 16.24 \\
UnivFD & 98.83 & 98.55 & 97.78 & 96.92 & 1.90 & 94.46 & 95.45 & 92.92 & 93.90 & 2.53 & 53.16 & 56.41 & 65.56 & 57.75 & 12.40 \\
\bottomrule
\end{tabular}}
\end{table*}
\textbf{Analysis-2.1: Analysis of variations in the detectability of generative models and their relationship to detector type.}

\vspace{0.2cm}
\textbf{EXP A.} The results in ~\cref{fig:heatmaps} indicate that the performance of the three types of detectors does not exhibit a simple linear decline as the overall generation quality of the models improves. To further investigate the underlying patterns, we extended our analysis in~\cref{fig:correlation_video_classification,fig:correlation_AGID,fig:correlation_AGVD} to examine the correlation between detector performance on different generative models and multiple generation metrics. Our analysis reveals that detection difficulty is not directly determined by the “overall generation quality” of the model. Neither in T2V nor I2V tasks did we observe the expected clear relationship between overall quality and detection difficulty—particularly for T2V models, where almost no discernible pattern exists. While a slight trend appears in I2V tasks, we postulate that this may be a statistical artifact due to the limited number of available models.

What proves to be more indicative are the correlations between certain specific dimensions of generation quality and detection performance, as detailed below:

\begin{itemize}
    \item The performance of all three types of detectors shows a significant correlation with the \textbf{imaging quality} and \textbf{frame-level quality} of the generation models.
    \item The performance of video classification models is also correlated with the \textbf{dynamic degree} of the generation models.
    \item  Both AI-generated video detection models and AI-generated image detection models exhibit correlations with \textbf{aesthetic quality}.
    \item  In I2V tasks, the performance of AI-generated video detection models and video classification models is further correlated with \textbf{background consistency} and \textbf{subject consistency}. 
\end{itemize}

Based on these phenomena, a seemingly plausible inference would be that the higher the quality of a generated video in certain specific dimensions, the more difficult it is to detect. However, regarding this viewpoint, two critical limitations must be emphasized:
1. Correlation does not imply causation. The observed associations cannot be directly interpreted as “higher quality leads to greater detection difficulty.” 
2. More importantly, further analysis demonstrates that this inference is directly refuted.

\textbf{EXP B.} To systematically validate the relationship between the quality of generative models and detection difficulty, we conducted a comparative analysis of detectors trained on different generative models (\cref{fig:correlation_DeCoF_trained,fig:correlation_UnivFD_trained,fig:correlation_I3D_trained}). The experimental results reveal a complex phenomenon: the quality of generated content and detection difficulty do not exhibit the intuitively assumed positive correlation. Specifically, the earlier conclusion drawn from studies based on Open-Sora—that "higher quality leads to greater detection difficulty"—lacks generalizability. In tests involving other generative models, we observed the opposite trend: for detectors trained on high-quality generative models, the detection difficulty often decreases as certain generation metrics improve. Based on these findings, we propose that the key factor determining detection performance is not the absolute quality of the generative model, but rather the correlation between the generative pattern and the detector's training data. When the generative pattern of a test sample falls within the distribution of the training data (high correlation), the detector performs well; conversely, under out-of-distribution generalization scenarios (low correlation), detection efficacy decreases significantly. This naturally leads to a further question: within a single generative model, is there a correlation between the quality of its outputs and the difficulty of detecting them?  This question is systematically examined in Analysis-2.4. 

\vspace{0.2cm}
\noindent\textbf{Analysis-2.2: Key characteristics of generative models for producing high-quality training data and their relationship to detector type.}
\vspace{0.2cm}

\cref{deocf_train,uni_train,i3d_train} presents the cross-evaluation results of three detectors, each trained specifically on one of 12 T2V models and evaluated on all of them. This comprehensive assessment reinforces our prior conclusion: a superior generative model does not yield a more effective detector. Notably, detectors trained on certain models, like Wan 2.1, even exhibit performance degradation. Furthermore, higher evaluation metrics of a generative model do not guarantee improved detection performance. 

To further validate the conclusion presented in  Analysis-2.1, we conducted an additional experiment. \cref{deocf_test,uni_test,i3d_test} illustrates the relationship between the overall detection performance of detectors (each trained on one of 12 distinct generative models) on a specific generator and that generator's own performance. This experiment provides further evidence that a higher-quality generative model does not necessarily lead to a greater challenge for detection.

\textbf{Analysis-2.3:  The effects of sampling steps on detector performance.}
\begin{figure}[htbp]
    \centering
    \includegraphics[width=\linewidth]{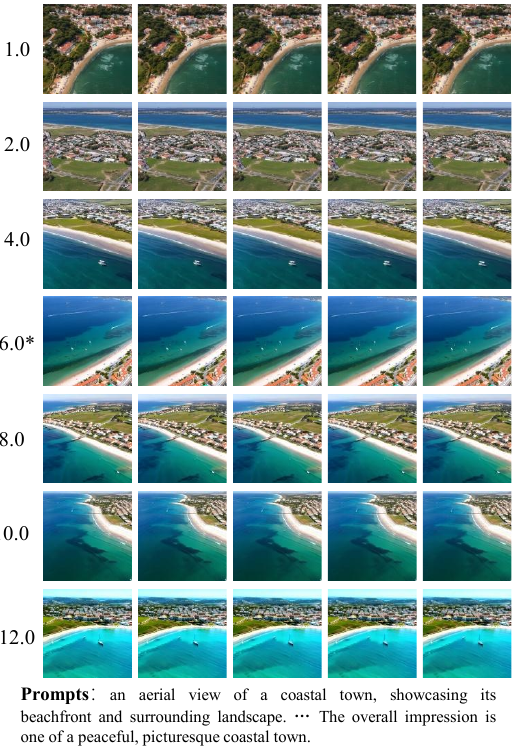}
      \includegraphics[width=\linewidth]{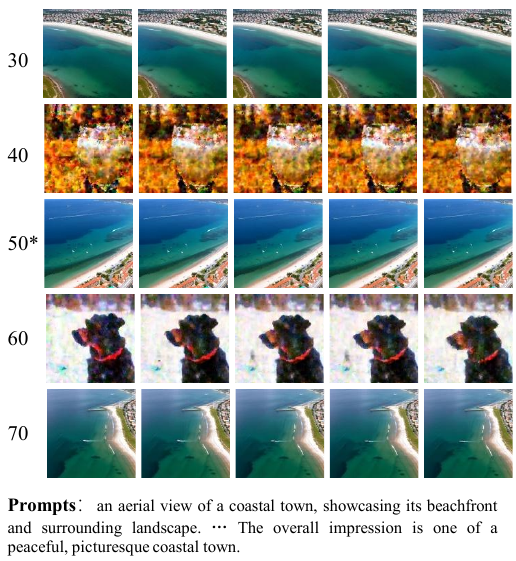}
    \caption{The impact of different sampling steps and guidance scale on videos. * Represents the default parameters of the model.} 
    \label{fig:guidance}
\end{figure}

For a given generative model, sampling steps and guidance scale are two key parameters that influence output quality. In Analysis-2.3 and 2.4, we investigate how these factors affect detector performance and analyze the relationship between such effects and generation quality. All experiments in ~\cref{tab:guidance and steps,tab:guidance stpes quality} are conducted on a newly constructed dataset. We use EasyAnimate as the generative model, with sampling steps set to 30, 40, 50, 60, and 70, and guidance scale set to 1.0, 2.0, 4.0, 6.0, 8.0, 10.0, and 12.0. For each parameter combination, we selected the first 200 samples from the AIGVDBench test set, generating 200 synthetic videos and 200 real videos respectively, resulting in a new test set of 400 videos.

We first examine the effect of sampling steps. As shown in ~\cref{tab:guidance stpes quality}, as the number of sampling steps increases, the quality of the generated videos across both individual dimensions and overall performance first improves and then deteriorates. This pattern is expected, since EasyAnimate recommends 50 sampling steps. ~\cref{fig:guidance} shows a set of videos generated with different step counts. When the step count deviates from 50, the videos exhibit noticeable artifacts: at 40 steps, the frames are dominated by noise, while at 60 steps, the content does not match the text prompt.

In terms of detector performance, we observe two distinct trends. Detectors trained on data with stronger attributes, such as Accvideo, AnimateDiff, EasyAnimate, HunyuanVideo, and VideoCrafter, which emphasize image quality and background \& subject consistency tend to achieve better detection performance as step count increases, up to a point, after which performance drops. In contrast, detectors trained on lower-quality generated data often show the opposite trend: performance first declines and then improves. This reveals two important insights: 

First, the quality of training data is not simply a linear combination of individual scores, but involves a more complex interplay of factors such as visual quality and temporal coherence. Second, the results reaffirm our earlier conclusion that higher-quality generation models are not necessarily harder to detect; rather, detection performance fundamentally depends on the similarity between the test-time generative model and those seen during training.
We also note the impact of detector architecture. For example, frame-based detectors such as UnivFD and DeCoF, which rely on CLIP-ViT-L/14, are more sensitive to image quality and exhibit highly consistent behavior across models. In contrast, video-level detectors such as I3D consistently show a rise-then-fall performance trend. Finally, we observe that results obtained on the 400-sample dataset (~\cref{fig:guidance} for Opensora at 50 steps, and at guidance scale 6.0) differ significantly from the results reported in the main body for the same generative models. This further highlights the importance of benchmark scale in obtaining reliable evaluation outcomes.

\textbf{Analysis-2.4: The effects of  guidance scale on detector performance.}

In contrast to the impact of sampling steps, as the guidance scale increases from 1.0 to 12.0, the generated videos exhibit divergent patterns of change across different quality dimensions. While most metrics follow the expected pattern of initial improvement followed by decline, motion smoothness demonstrates a consistent deterioration throughout this range.
Regarding detector performance, with the increase in guidance scale, all detectors ( except those trained on AnimateDiff ) show a steady improvement in detection capability. 

In stark contrast, detectors based on AnimateDiff exhibit progressive performance degradation as the guidance scale grows.
The underlying reasons for this phenomenon remain unclear, particularly given that motion smoothness is fundamentally a video-level metric, while the frame-level detector UnivFD still conforms to the overall trend observed in video-based detectors. One plausible hypothesis is that higher guidance scales enhance the model's adherence to textual prompts, resulting in greater content consistency across generated videos. This increased uniformity may consequently reduce the challenge of distinguishing between real and synthetic content, thereby improving detection performance for most models.

\textbf{Analysis-2.5: The impact of video type on detection difficulty.}

We have consistently emphasized the importance of dataset scale and diversity. While our previous analysis has demonstrated the significance of scale, we further investigate how different types of videos generated by the same model affect detection difficulty. All videos in this experiment are sourced from the AIGVDBench test set. Since a single video may possess multiple attributes, making clear categorization challenging, we utilize videos with four distinct spatial content types to illustrate this issue. Each category contains 400 videos (200 real and 200 generated), with the spatial content strictly limited to one of the four types. It is also important to note that these videos might still contain spatial elements beyond the four specified categories, as video content is inherently complex, particularly given that we employed detailed textual prompts to control the content generation.

Furthermore, we selected four representative network architectures for evaluation. For instance, CNNSpot employs a ResNet-50 backbone, a architecture widely adopted in generated image detection models. Another example is UnivFD, which utilizes a CLIP-ViT-L/14 framework. For video classification, we included I3D as a representative architecture.
As shown in ~\cref{tab:category_performance}, for lower-performing generative models, the impact of video content variation on detection performance is relatively minimal. However, as the quality of generative models improves, this performance discrepancy becomes increasingly pronounced. Ultimately, the same detector can exhibit performance variations of up to approximately 15\% across different video types, underscoring the critical importance of maintaining balanced content representation in datasets.

\section{Details of AIGVDBench}
\label{7}
\subsection{Brief Introduction of Generation Models}
\label{7.1}
\textbf{A. Video Generation Models.}
The development of video generation models has undergone a significant evolution from basic generative capabilities to the production of high-quality, long-sequence videos. Early research\cite{show1,videofusion,magicvideo,preserve,align,text2video} primarily focused on generating short video clips (around 2–3 seconds), emphasizing fundamental inter-frame consistency and adherence to simple textual prompts. This was later extended to multimodal tasks~\cite{chen2023motion,esser2023structure,ouyang2024codef,qi2023fatezero}, though the emphasis remained largely on surface-level realism. In recent years, models like Sora~\cite{Sora,Sora2}, kling~\cite{Kling}, Gen-3~\cite{Gen-3}, and HunyuanVideo~\cite{hunyuanvideo} have shifted focus toward intrinsic realism. Leveraging large-scale training and Transformer-based architectures, these models achieve long-sequence generation with attention to physical laws, commonsense reasoning, and creative composition, driving video generation toward applications such as AI-assisted filmmaking and world model simulations. However, as AI-generated videos proliferate across social platforms, society is facing a new and far-reaching trust crisis.

\textbf{B. Used Video Generation Models in AIGVDBench .}
During the construction of AIGVDBench, we faced the objective constraint of limited GPU memory resources. Accordingly, the generative models were selected based on the following criteria: the model must be able to run within 80GB of GPU memory while generating a single video in under two minutes, or alternatively, operate within 24GB of GPU memory with a generation time of less than eight minutes per video. Despite these constraints, the dataset construction process still required nearly four months to complete; the full timeline is provided in \cref{7.8}. In the following, we briefly introduce the video generation models used in AIGVDBench, omitting some models that were applied repeatedly.
\begin{table*}[t]
\centering
\caption{AIGVDbench Used Video Generation Models Configuration.}
\label{tab:video_models}
\resizebox{\textwidth}{!}{
\begin{tabular}{lcccccccc}
\toprule
Model & TeaCache & Resolution & \makecell{Num\\Frames} & \makecell{Num\\Steps} & \makecell{CPU\\ Offload} & Precision & \makecell{Guidance\\Scale} & FPS \\
\midrule
Open-Sora & None & 256 (16:9) & 37 & 50 & True & bf16 & 7.5 & 24 \\
Hunyuan & 0.15 & 544 $\times$ 960 & 33 & 50 & True & fp8 & 6.0 & 24 \\
Wan2.1 & 0.08 & 480 $\times$ 832 & 49 & 50 & True & bf16 & 5.0 & 16 \\
AccVideo & None & 544 $\times$ 960 & 33 & 5 & True & fp16 & 1.0 & 24 \\
AnimateDiff-L & None & 512 $\times$ 512 & 16 & 8 & False & — & 1.0 & 10 \\
CogVideoX & None & 480 $\times$ 720 & 81 & 50 & False & fp16 & 6.0 & 16 \\
Pyramid-Flow & None & 384 $\times$ 640 & 89 & 10+20 & False & bf16 & 7.0 & 24 \\
VideoCrtafter & None & 320 $\times$ 512 & 16 & 50 & False & — & 12.0 & 10 \\
EasyAnimate & 0.08 & 384 $\times$ 672 & 49 & 50 & True & bf16 & 6.0 & 8 \\
LTX(T2V) & 0.05 & 512 $\times$ 768 & 101 & 50 & False & bf16 & 3.0 & 24 \\
LTX(I2V) & 0.05 & 513 $\times$ 768 & 50 & 50 & False & bf16 & 3.0 & 24 \\
LTX(V2V) & None & 514 $\times$ 768 & 50 & 50 & False & bf16 & 3.0 & 24 \\
RepVideo & None & 480 $\times$ 720 & 49 & 50 & False & fp16 & 6.0 & 8 \\
SEINE & None & 320 $\times$ 512 & 16 & 250 & False & fp16 & 8.0 & 8 \\
SVD & None & 320 $\times$ 513 & 16 & 25 & False & — & 1.0-3.0 & 5 \\
IPOC&   None& 480 $\times$ 720 & 49 & 50 & False & bf16 & 6.0 & 8\\
\bottomrule
\end{tabular}}
\vspace{0.1cm}
\begin{itemize}
    \item \textbf{TeaCache}: TeaCache~\cite{liu2025timestep} acceleration (enabled/disabled) and the associated $rel\_11\_thresh$ parameter.
    \item \textbf{Offload}: Whether the model employs memory offloading techniques.
    \item \textbf{FPS}: Frames per second in the generated video output.
    \item  \textbf{10+20}: $num\_inference\_steps: [20, 20, 20],
            video\_num\_inference\_steps=[10, 10, 10].$
\end{itemize}
\end{table*}

\begin{itemize}
    \item \textbf{Closed source video generation model.}
     \vspace{0.1cm}
    \begin{itemize}[leftmargin=10pt]
        \item \textbf{CausVid} (CVPR, 2025)  is a fast causal video generation model jointly developed by MIT and Adobe. It employs an innovative causal transformer architecture and knowledge distillation to generate high-quality videos with very low initial latency (1.3s for the first frame) and supports streaming generation with real-time instruction modification, excelling in speed and temporal coherence.  \textit{At the time of AIGVDBench's completion, this model was not yet open-sourced and was therefore categorized as proprietary.}
        \vspace{0.1cm}
        
        \item \textbf{Gen2} is a diffusion-transformer video model that turns text prompts or stills into short clips, lets users inject camera trajectories plus segmentation/depth controls, and wraps everything in a creator-friendly web workflow with asset libraries and multi-track timelines for agile ad, music, or narrative projects.
         \vspace{0.1cm}
        
        \item \textbf{Gen3} builds on the Gen2 stack with tighter text-visual alignment, large-scale live-action data, and physics-aware finetuning to deliver longer, steadier shots; it adds motion-reference uploads, mocap presets, and collaborative workspaces aimed at cinematic grading, dialogue-driven scenes, and bespoke brand content.
         \vspace{0.1cm}
        \item \textbf{Kling} is Kwai’s multimodal video suite built on a 3D diffusion transformer with spatio-temporal latents, covering text-to-video up to roughly a minute, image-to-video expansion, character animation, physics priors, control channels, and audio alignment so creators can preserve identity consistency, steer camera choreography, and rapidly deliver music videos, ad spots, or concept reels.
        \vspace{0.1cm}
        \item \textbf{Luma} Dream Machine is Luma Labs’ generative video and 3D creation suite, blending diffusion backbones with neural radiance field models to turn text, images, or captured footage into high-res videos, panoramas, and editable NeRF assets, and shipping browser-based tools for lighting, material tweaks, and camera path planning so teams can iterate on commercials, concept art, or digital twins with minimal friction.
       \vspace{0.1cm}
        \item \textbf{Open-Sora} is an open-source project initiative aimed at replicating and advancing the capabilities of OpenAI's Sora model. It focuses on significantly reducing the training costs associated with high-quality video generation models and promoting the development of accessible, open-source alternatives in the field, with its v2.0 version showing performance close to the closed-source Sora model.
        \vspace{0.1cm}
        \item \textbf{Pika} is an AI video generation tool developed by Pika Labs, enabling users to create and edit videos from text prompts. It is recognized for its user-friendly interface and ease of use, making it popular for social media content creation and quick video prototyping, allowing for efficient creative expression.
         \vspace{0.1cm}
         \item \textbf{Sora} is a groundbreaking text-to-video model developed by OpenAI. It is capable of generating high-fidelity, temporally coherent video clips up to one minute long directly from text descriptions, setting a notable benchmark for realism and narrative complexity in AI-generated video and demonstrating significant advancements in the field.
         \vspace{0.1cm}
         \item \textbf{Vidu} is an AI video generation model capable of producing videos from text descriptions. Its capabilities have been demonstrated in public showcases, highlighting its potential in the text-to-video generation space and indicating its role as a contender in the rapidly evolving field of AI video synthesis.
          \vspace{0.1cm}
          \item  \textbf{Wan} is a comprehensive and open video foundation model suite introduced by Alibaba's Wanxiang team. It includes models with 1.3B and 14B parameters, covering multiple downstream tasks like text-to-video and image-to-video generation. Notably, it is the first model capable of generating visual text in both Chinese and English.
    \end{itemize}
    \item \textbf{Open source video generation model.}
     \vspace{0.1cm}
    \begin{itemize}
    \item \textbf{EasyAnimate}: A high-performance, open-source long video generation project developed by Alibaba Cloud's PAI team, based on a Diffusion Transformer (DiT) architecture. It modifies the VAE and DiT structures to better support video generation, extending the original 2D image synthesis DiT framework into the 3D video domain and introducing a motion module to capture temporal dynamics. The proposed Slice VAE addresses GPU memory challenges during the encoding and decoding of long, large videos, enhancing compression efficiency along the time dimension. It offers a complete solution for HD long video generation, including data preprocessing, VAE training, DiT training, model inference, and evaluation.
     \vspace{0.1cm}

    \item \textbf{LTX}: An all-in-one generative AI platform designed for video production, catering to filmmakers, advertisers, and creative teams. It transforms creative ideas into high-quality, professional videos, streamlining stages from scripting and storyboarding to editing and final delivery. Its AI creative engine, LTX-2, powers synchronized audio and video generation, offers 4K fidelity, and includes multiple performance modes for efficiency. The platform provides tools for controlling camera movement, defining motion with keyframes, using visual references, and maintaining consistency with AI characters / objects.
     \vspace{0.1cm}

    \item \textbf{Pyramid-Flow} (ICLR 2025): An open-source high-definition video generation model jointly released by Peking University, Beijing University of Posts and Telecommunications, and Kuaishou Technology. It employs a pyramid flow matching algorithm, which decomposes the video generation process into multiple stages starting from low resolution and progressively upgrading to high resolution. This approach effectively handles high-dimensional video data, significantly reducing computational costs and token count while maintaining high visual quality.
     \vspace{0.1cm}

    \item \textbf{SEINE} (ICLR 2024): A diffusion model designed for generative transitions and prediction, focusing on generating videos from short to long durations. As part of the Vchitect video generation system, it can be integrated with text-to-video frameworks like LaVie. Based on Stable Diffusion v1.4, SEINE is particularly adept at creating smooth transitional effects and predicting video content, making it suitable for applications in filmmaking, advertising, and video editing.
     \vspace{0.1cm}

    \item \textbf{SVD}: A pioneering video generation model that applies the diffusion transformer architecture to video synthesis, developed by Google Research. It is capable of generating short video clips with coherent motion directly from a single static image, effectively animating still pictures. The model demonstrates significant capabilities in producing realistic and temporally consistent motions from image inputs, establishing a strong baseline for image-to-video generation research.
     \vspace{0.1cm}

    \item \textbf{VideoCrafter} (CVPR 2024): An open-source video generation model and toolkit, with versions like VideoCrafter2 supporting high-quality text-to-video and image-to-video generation. It aims to provide a reproducible and modifiable solution for the video generation community, often serving as a strong baseline for research and application development. The project typically offers a complete ecosystem, including model training, inference, and evaluation tools.
     \vspace{0.1cm}

    \item \textbf{AnimateDiff}: An influential open-source framework designed to inject motion dynamics into personalized text-to-image models, enabling them to generate video clips. Its core innovation lies in learning a universal motion module that can be plugged into various pre-existing personalized T2I models without modifying their original weights. This approach efficiently animates a wide range of personalized concepts, significantly advancing personalized video generation.
     \vspace{0.1cm}

    \item \textbf{CogVideoX} (ICLR 2025): A powerful open-source video generation model developed by Tsinghua University. Based on the Transformer architecture, it supports text-to-video and image-to-video generation, as well as video continuation. The model employs technologies like 3D VAE and 3D RoPE positional encoding to significantly enhance temporal coherence and long-range dependencies in the generated videos.
     \vspace{0.1cm}

    \item \textbf{HunyuanVideo}: A sophisticated text-to-video generation model developed by Tencent. It leverages a robust multi-stage training strategy, progressing from low to high resolution and from image pre-training to video training, to achieve high-fidelity output. A key strength is its deep integration of the proprietary Hunyuan large language model for enhanced text comprehension and semantic alignment.
     \vspace{0.1cm}

    \item \textbf{IPOC}: A research direction in video generation focused on complex scene composition and open-world synthesis from image prompts. It tackles challenges like seamlessly integrating multiple elements from reference images into a cohesive and dynamic video scene, aiming for greater creative control and visual consistency.
     \vspace{0.1cm}

    \item \textbf{RepVideo}: A video generation method that emphasizes efficient and high-fidelity reproduction of visual elements. The approach focuses on achieving high-quality replication of details, motions, or styles within the generated video content through innovative architectural designs or training strategies.
     \vspace{0.1cm}

    \item \textbf{Wan2.1}: A version iteration of the "Wan" video generation model suite, potentially from Alibaba's Wanxiang team. This version introduces improvements over its predecessors in areas such as generation quality, supported resolutions, and specific capabilities like generating visual text in multiple languages. It is designed to be an efficient model suitable for consumer-grade hardware.
     \vspace{0.1cm}

    \item \textbf{AccVideo}: A novel extraction method focused on accelerating existing video diffusion models. Its core innovation lies in using a pre-trained model to generate a high-quality synthetic dataset. Through trajectory-based few-step guidance and an adversarial training strategy, the student model learns to produce high-quality videos with very few inference steps, achieving significant speed boosts while maintaining comparable video quality.
\end{itemize}
\end{itemize}

\noindent\textbf{The detailed parameters of the generative models can be found in~\cref{tab:video_models}.}

\subsection{Introduction of the Used Detection Methods}
\label{7.2}
\textbf{A. Generated Image Detection.} Existing approaches for synthetic image detection can be broadly categorized into two main research directions.
The first line of methods focuses on forgery artifact learning, which aims to capture low-level traces left by generative models in either the spatial or frequency domain. For instance, some studies employ backbone networks such as  ResNet combined with specialized modules to extract frequency features~\cite{freqdet}, gradient information~\cite{Gram-Net,tan2023learning}, or reconstruction anomalies~\cite{wang2023dire}. Other works  analyze local pixel relationships arising from upsampling operations~\cite{NPR}. Although these methods perform well on known generative models, their generalization is often constrained by the training data distribution, and they typically require large-scale training from scratch.
The second category builds on frozen pre-trained vision models, such as CLIP-ViT-L/14, to extract general-purpose features, with only a lightweight classification head being fine-tuned for detection~\cite{Universal}. Such methods mitigate overfitting to specific artifacts and demonstrate promising cross-model generalization in few-shot scenarios. However, the frozen encoders often retain a substantial amount of forgery-irrelevant information, limiting further performance gains. Recent efforts have sought to incorporate forgery-aware priors by introducing prompt tuning or adapter modules into frozen architectures~\cite{effort,ForgeLens}, advancing this line towards more efficient and discriminative detection.

\noindent\textbf{B. Deepfake Detection.} The field of deepfake face detection has evolved to address the unique challenges of media forgery, with research directions broadly classified into the following categories:
1. Detection Based on Inherent Forgery Features: Early methods primarily leveraged convolutional neural networks (CNN) as end-to-end binary classifiers~\cite{rossler2019faceforensics++,afchar2018mesonet}. Subsequent work has specialized by targeting specific forgery traces. Spatial-domain methods focus on cues such as blending boundaries~\cite{facexray,zhao2021multi} or localized artifacts~\cite{nguyen2024laa}. 
2. Detection Enhanced by Data Augmentation: To improve generalization, a prominent strategy involves augmenting training data with synthetic forgeries. This includes generating pseudo deepfakes by applying face warping~\cite{facexray}, self-blending~\cite{shiohara2022detecting,chen2022self}, or adversarial blending simulations~\cite{yan2024transcending}. The core idea is to preemptively expose the detector to a wider variety of forgery types and artifacts, thereby enhancing its robustness against unseen manipulation techniques.
3. Detection Focused on Generalization Capabilities: Given the rapid evolution of generation models, a significant research thrust is dedicated to improving cross-dataset and cross-method generalization. Efforts span multiple dimensions, including advanced data augmentation~\cite{yan2023ucf,ma2025specificity,xu2023tall}, frequency feature analysis~\cite{durall2020watch,qian2020thinking}, and network architecture innovations~\cite{zhao2021multi}. 

\textbf{C.  Used Detection Methods in AIGVDBench.}
We categorize potential approaches for synthetic video detection into four groups: (1) video classification models, which employ specialized network architectures for video based tasks; (2) AI generated image detection models, which can be adapted as frame level detectors for synthetic videos; (3) dedicated AI generated video detection models specifically designed for this task; and (4) the transfer of vision language models, which also represents a highly promising direction. Based on this taxonomy, we selected 33 mainstream models for evaluation.
It should be noted that Deepfake detection models were excluded from our study, as they are designed specifically for facial forgery and are not suitable for general synthetic video detection. Additionally, methods requiring extensive input preprocessing, such as video reconstruction, were also omitted due to the substantial scale of our dataset, where such operations would significantly impede evaluation efficiency. In the following 
section, we introduce the 33 detectors included in our benchmark.
\vspace{0.1cm}
\begin{itemize}
    \item \textbf{Video Classification models.}
    \vspace{0.1cm}
    \begin{itemize}
        \item \textbf{MViT (CVPR, 2022)}: An open-source repository from Facebook Research (Meta AI) that implements Multiscale Vision Transformers (MViT), a model designed to address limitations in standard Vision Transformers regarding computation and multiscale feature modeling. MViT introduces a hierarchical transformer architecture that combines self-attention with pooling operations, enabling efficient capture of features at multiple scales. This leads to improved performance and reduced computational cost compared to conventional models. The repository provides modular PyTorch implementations for image classification, object detection, and video understanding tasks, achieving state of the art results on benchmarks such as ImageNet~\cite{deng2009imagenet} and Kinetics~\cite{kay2017kinetics}.
        \vspace{0.1cm}
\item \textbf{UniFormer (ICLR, 2022)}: A unified transformer framework that seamlessly integrates convolution and self-attention into a single network architecture for both video understanding and image recognition. It delivers state of the art performance by merging local and global modeling strengths, and features flexible scalability for various vision tasks.
\vspace{0.1cm}

\item \textbf{VideoSwin (CVPR, 2022)}: A video understanding model based on the Swin Transformer architecture, which efficiently captures spatio-temporal features in videos through a hierarchical window based attention mechanism. It is widely used in tasks such as action recognition and video classification, demonstrating excellent performance.
\vspace{0.1cm}

\item \textbf{VideoMAE (NeurIPS, 2022)}: A self-supervised video pre-training model that employs Masked Autoencoders (MAE) to efficiently learn spatiotemporal features from videos. It randomly masks a large portion of space-time blocks in the input video and reconstructs the full video from the remaining visible parts. This encourages the model to learn richer and more robust representations, drastically improving data efficiency. VideoMAE achieves state of the art performance in various video understanding tasks (like action recognition), advancing self-supervised learning in the video domain.
\vspace{0.1cm}

\item \textbf{TSM (ICCV, 2019)}: An efficient video understanding model that introduces the "Temporal Shift Module" for effective temporal modeling. This approach shifts a portion of feature channels along the temporal dimension, enabling information exchange between frames without introducing additional parameters or computation. TSM achieves a good balance between efficiency and performance, making it suitable for various video understanding tasks like action recognition, and it has achieved excellent results on several benchmark datasets.
\vspace{0.1cm}

\item \textbf{SlowFast (ICCV, 2019)}: utilizes dual pathways: a "slow" pathway and a "fast" pathway, capturing video spatiotemporal features at different temporal resolutions for precise understanding of dynamic visual content. The codebase supports tasks such as action recognition, making it widely used in both research and practical applications.
\vspace{0.1cm}

\item \textbf{TimeSformer}:  is an end to end video understanding model based purely on Transformers, applying attention mechanisms along both spatial and temporal dimensions to effectively extract video features. The model has driven the adoption of pure self-attention methods in tasks like video classification.
\vspace{0.1cm}

\item \textbf{I3D (CVPR, 2017)}: A convolutional neural network model for video classification based on the I3D (Inflated 3D ConvNet) architecture, trained on the large scale Kinetics human action recognition dataset. I3D extends 2D convolutions to 3D, effectively modeling spatiotemporal features for better recognition of dynamic actions in videos. This project provides a strong baseline for video understanding tasks and is widely used in action recognition research.
\vspace{0.1cm}

\item \textbf{UniFormerV2 (ICCV, 2023)}: A state of the art model for spatiotemporal learning in videos. UniFormerV2 innovatively combines the Image Vision Transformer (ViT) architecture with the video specific UniFormer framework, leveraging the strengths of both for spatial and temporal modeling. This model efficiently captures rich spatiotemporal information within videos, enhancing performance in tasks like action recognition, and offers a more unified and general solution for video representation learning.
\vspace{0.1cm}

\item \textbf{X3D (CVPR, 2020)}: A deep learning model for efficient video recognition. X3D expands existing 3D convolutional neural network architectures along multiple axes—such as spatial, temporal, and network width—achieving excellent video understanding performance with minimal computational cost. The model is lightweight and fast, making it well suited for action recognition and other tasks in resource constrained settings, and it has demonstrated strong results on various video benchmarks.

    \end{itemize}
    \vspace{0.1cm}
    \item \textbf{AI-Generated Image Detection Models.}
     \vspace{0.1cm}
    \begin{itemize}
        \item \textbf{Fredect (ICML, 2020)}: is designed to automatically identify AI-generated or manipulated content in images or videos. It integrates multiple feature analysis methods and deep learning techniques to detect anomalous patterns or artifacts in visual data, thereby improving the recognition accuracy of synthetic media. Fredect is applicable in academic research, content moderation, and digital forensics, serving as a key technology for enhancing media authenticity and security.
             \vspace{0.1cm}
\item \textbf{NPR (CVPR, 2024)}: A deepfake detection project leveraging Non-Photorealistic Rendering (NPR) features. By extracting NPR-based image representations, this method effectively discriminates real from fake content, improving robustness and generalization across diverse generators and complex backgrounds. The project is well-suited for applications in digital media security and content authentication.
     \vspace{0.1cm}

\item \textbf{Fusing}: A project focused on fusing global and local features for video understanding tasks. It aims to jointly model high-level contextual information and fine-grained local details in videos, enhancing the representation of complex dynamic content and improving video analysis accuracy. Such fusion strategies are valuable in domains such as action recognition and multimodal reasoning, supporting more precise interpretation of video content.
     \vspace{0.1cm}

\item \textbf{Gram-Net (CVPR, 2020)}: By enhancing global texture features in facial images, the model improves the accuracy and robustness of forged face detection, particularly against DeepFake and face-swapping techniques. Combining texture analysis with deep learning, the method advances fake face detection under real-world conditions and is widely applicable in media authentication and cybersecurity.
     \vspace{0.1cm}

\item \textbf{CNNspot (CVPR, 2020)}:  It detects CNN-synthesized images by identifying subtle statistical and textural artifacts specific to synthetic images. The method effectively distinguishes real from AI-generated content, demonstrating strong performance on early generative models and providing valuable insights for deepfake detection and image forensics.
     \vspace{0.1cm}

\item \textbf{D3 (CVPR, 2025)}: A Discrepancy Deepfake Detector designed for multi-generator deepfake detection. It employs a dual-branch architecture: a main branch that processes original image features and an auxiliary branch that handles specially distorted inputs. By fusing features from both branches, the model captures common discrepancy signals across generators and decomposes complex forgery patterns, leading to improved adaptability and generalization in detecting fake content from diverse sources.
     \vspace{0.1cm}

\item \textbf{ForgeLens (ICCV, 2025)}:  It introduces novel network structures and training strategies that effectively localize forged regions within images, achieving strong generalization even with limited annotated data. The method is applicable to deepfake and image manipulation detection, offering an efficient and universal solution for digital media security.  Based on tdistinct experimental configurations, we categorized the evaluation settings into ForgeLens1 and ForgeLens3.
     \vspace{0.1cm}

\item \textbf{Effort-AIGI-Detection (ICML, 2025)}: This work improves the generalization of AI-generated image detection via Orthogonal Subspace Decomposition (OSD), which decomposes image features into independent components to better separate real from fake characteristics. The approach significantly enhances cross-domain and multi-generator detection performance, making it suitable for real-world applications in digital media security and deepfake forensics.
\end{itemize}
     \vspace{0.1cm}
    \item \textbf{AI-Generated Video Detection Models.}
    \vspace{0.1cm}
    \begin{itemize}
        \item \textbf{DeMamba} focus on large-scale AI-generated video detection. In response to the rapid growth of video generation technology and the popularity of video content on social media, this project aims to develop methods and models capable of distinguishing between real and AI-generated videos over million-scale datasets (GenVideo Benchmark). The DeMamba framework extracts and analyzes features from both genuine and fake videos, helping curb the spread of misinformation and meeting the rising demand for efficient and accurate video forgery detection tools, especially for digital media security and content moderation use cases.
          \vspace{0.1cm}

        \item \textbf{DeCoF}  focus on detecting AI-generated videos via frame consistency analysis. By examining the consistency between video frames and leveraging temporal and sequential features, DeCoF effectively identifies AI-generated content and uncovers subtle artifacts that may not be apparent in individual frames. This approach enhances detection performance for complex generative videos and is suitable for applications such as media content moderation and video forensics, helping to limit the spread of fake video information.
    \end{itemize}
      \vspace{0.1cm}
      \item \textbf{Vision-Language Models.}
      \vspace{0.1cm}
      \begin{itemize}
          \item \textbf{DeepSeek-VL-7B-Chat}: An instruction-tuned multimodal model that pairs SigLIP-L and SAM-B encoders for 1024×1024 visual input with the DeepSeek-LLM-7B backbone. It was trained on approximately 2 trillion text tokens and 400 billion multimodal pairs. The model features conversational formatting, strong diagram and document comprehension capabilities, and flexible image grounding APIs, making it suitable for general-purpose assistant applications.
          \vspace{0.1cm}
\item \textbf{InternVL3-8B}: Integrates an InternViT-300M vision encoder with a Qwen2.5-7B decoder through native multimodal pretraining, mixed preference optimization, and variable visual position encoding. The model targets diverse applications including document analysis, GUI understanding, spatial reasoning, video processing, and multilingual tasks, while providing tool-calling capabilities and video-aware chat templates.
\vspace{0.1cm}

\item \textbf{Qwen2.5-VL-3B-Instruct}: A lightweight model combining a Vision Transformer front end (featuring windowed attention, SwiGLU, and RMSNorm) with a 3B Qwen2.5 decoder via a multimodal projector. Dynamic spatial and temporal sampling enables unified image and video processing, while the compact decoder ensures fast inference and precise instruction following.
\vspace{0.1cm}

\item \textbf{Qwen2.5-VL-7B-Instruct}: Scales the ViT–projector–LLM architecture to a 7B language core and higher-capacity vision encoder, maintaining dynamic resolution and mRoPE temporal encoding. Structured-output heads provide reliable performance for chart parsing, UI reasoning, and enterprise document extraction pipelines.
\vspace{0.1cm}

\item \textbf{Qwen2.5-VL-32B-Instruct}: Expands the modular ViT + projector architecture with a 32B parameter decoder, incorporating reinforcement-tuned preference heads and extended-position embeddings. Capable of handling 32K+ context windows, multi-image analytics, and complex agent-style planning with enhanced reasoning depth.
\vspace{0.1cm}

\item \textbf{DeepSeek-VL2}: A Mixture of Experts pipeline combining a SigLIP-style encoder and dynamic tiling with the DeepSeekMoE-27B language expert pool, activating only a subset of experts per token. This design balances latency with strong performance on chart understanding, OCR, and visual grounding, while supporting multi-image conversational prompts.
\vspace{0.1cm}

\item \textbf{DeepSeek-VL2-Small}: A compact version of VL2 based on the DeepSeekMoE-16B backbone with approximately 2.8B activated parameters. Retains the dynamic tiling preprocessor for up to two images, suitable for memory-constrained deployments while maintaining instruction stability and context-aware visual question answering.
\vspace{0.1cm}
\item \textbf{Emu3-Stage1}: First-stage Emu3 weights trained exclusively with next-token prediction over tokenized text and images using a single decoder-only transformer with 5120 context length. Leveraging a learned vision tokenizer and classifier-free guidance, it provides unified generation and perception capabilities, serving as the foundation for subsequent video-augmented stages.
\vspace{0.1cm}
\item \textbf{LLaVA-1.5-7B}: A classic architecture connecting a CLIP ViT-L/14 encoder to a Vicuna-7B decoder through a learned projector. Aligned via MSP/CC-SBU pretraining and GPT-curated VQA instructions, it excels at conversational image descriptions, simple grounding, and few-shot captioning with a well-documented chat template.
\vspace{0.1cm}
\item \textbf{FastVLM-7B}: Integrates the FastViTHD hybrid encoder—which generates fewer high-value tokens through strided attention—with a Qwen2-7B decoder and slim projector. Engineered for rapid time-to-first-token, it maintains competitive performance on OCR, chart understanding, and document comprehension while providing ready-to-use inference scripts.
\vspace{0.1cm}
\item \textbf{Kimi-VL-A3B-Instruct}: Combines a MoonViT native-resolution encoder with a 2.8B-activated Moonlight-16B MoE language core via an MLP bridge. Supporting 128K context windows, multi-image and video inputs, and agent-style workflows, it includes specialized prompts for UI automation and long-document analysis.
      \end{itemize}
\end{itemize}
\subsection{Introduction of Original Data}
\label{7.3}
The real video  in our AIGVDBench are sourced from the OpenVidHD dataset~\cite{nanopenvid}, a large-scale high-quality video collection specifically designed for high-definition video generation research. Derived from OpenVid-1M, it consists of a curated selection of approximately 433,000 high-resolution video clips at 1080p, totaling about 11,000 hours of content. Characterized by stringent quality control, OpenVidHD ensures superior visual quality and diverse thematic coverage, including natural landscapes, human activities, animal behaviors, and more. Each video clip is accompanied by accurate textual descriptions. Designed to address current limitations in high-definition video generation, the dataset provides a valuable foundation for training and evaluating text-to-video generation models, thereby advancing the development of video synthesis technologies. \textbf{In the OpenVidHD dataset, all videos are compressed using the MPEG-4 Part 2 format. For consistency in our evaluation, we have standardized the compression to H.264 across all videos in the AIGVBench benchmark. }
\subsection{Details of Prompts Categorization}
\label{7.4}
To balance classification accuracy and computational efficiency, we perform prompt categorization based on predefined WordNet synsets and keyword lists. This is achieved by defining a set of WordNet synsets and keywords for each category, then automatically scanning input prompts for words belonging to these lexical resources. When matches are found, the corresponding category labels are assigned to the prompt, enabling multi-label classification.
\begin{table*}[htbp]
\centering
\caption{Taxonomy of Predefined WordNet Synsets and Keywords (Major Content Aspect).}
\begin{subtable}[t]{\textwidth}
\centering
\resizebox{\linewidth}{!}{$
\begin{tabular}{@{}lcc@{}}
\toprule
                & \textbf{WordNet Synsets}                                                                                                                                                   & \textbf{Key Phrases/Words}                                                                                                                                                                                                                                                                                                                \\ \midrule
Actions         & \begin{tabular}[c]{@{}c@{}}'travel.v.01', 'compete.v.01', 'act.v.01', 'manipulate.v.02', 'eat.v.01', 'chew.v.01', 'drink', \\ 'move.v.03', 'move.v.02', 'change.v.01', 'make.v.03', 'make.v.01', 'run.v.01', 'crawl.v.01', \\ 'stretch.v.01', 'bend.v.01', 'twist.v.01', 'balance.v.01', 'crouch.v.01', 'leap.v.01', \\ 'analyze.v.01', 'memorize.v.01', 'predict.v.01', 'question.v.01', 'visualize.v.01', \\ 'calculate.v.01', 'evaluate.v.01', 'meditate.v.01', 'negotiate.v.01', 'persuade.v.01', \\ 'compliment.v.01', 'criticize.v.01', 'greet.v.01', 'introduce.v.01', 'console.v.01', \\ 'debate.v.01', 'fold.v.01', 'sweep.v.01', 'organize.v.01', 'repair.v.01', 'pack.v.01', \\ 'iron.v.01', 'polish.v.01', 'recycle.v.01', 'sketch.v.01', 'sculpt.v.01', 'compose.v.01', \\ 'edit.v.01', 'design.v.01', 'photograph.v.01', 'choreograph.v.01', 'improvise.v.01', \\ 'dribble.v.01', 'tackle.v.01', 'volley.v.01', 'parry.v.01', 'jog.v.01', 'sprint.v.01', \\ 'strike.v.01', 'dive.v.01', 'diagnose.v.01', 'program.v.01', 'audit.v.01', 'negotiate.v.01', \\ 'lecture.v.01', 'inspect.v.01', 'engineer.v.01', 'curate.v.01', 'scroll.v.01', 'upload.v.01', \\ 'debug.v.01', 'encrypt.v.01', 'stream.v.01', 'click.v.01', 'zoom.v.01', 'render.v.01', \\ 'sigh.v.01', 'glare.v.01', 'whisper.v.01', 'shrug.v.01', 'groan.v.01', 'cheer.v.01', \\ 'murmur.v.01', 'giggle.v.01', 'harvest.v.01', 'dig.v.01', 'plant.v.01', 'prune.v.01', \\ 'melt.v.01', 'evaporate.v.01', 'erode.v.01', 'ignite.v.01'\end{tabular} & None                                                                                                                                                                                                                                                                                                                                      \\
                & \multicolumn{1}{l}{}                                                                                                                                                       & \multicolumn{1}{l}{}                                                                                                                                                                                                                                                                                                                      \\
Kinetic Motions & \begin{tabular}[c]{@{}c@{}}'rotate.v.01', 'slide.v.01', 'pivot.v.01', 'vibrate.v.01', 'compress.v.01', 'expand.v.01', \\ 'lock.v.01', 'unlock.v.01', 'accelerate.v.01', 'decelerate.v.01', 'lubricate.v.01', \\ 'grind.v.01', 'weld.v.01', 'clamp.v.01', 'press.v.01', 'release.v.01', 'transmit.v.01', \\ 'absorb.v.01', 'ignite.v.01', 'discharge.v.01', 'charge.v.01', 'heat.v.01', 'cool.v.01', \\ 'filter.v.01', 'calibrate.v.01', 'adjust.v.01', 'trigger.v.01', 'activate.v.01', \\ 'deactivate.v.01', 'sense.v.01', 'measure.v.01', 'align.v.01', 'cut.v.01', 'drill.v.01', \\ 'screw.v.01', 'bolt.v.01', 'pump.v.01', 'vent.v.01', 'inject.v.01', 'extrude.v.01', \\ 'brake.v.01', 'clutch.v.01', 'gear.v.01', 'lever.v.01', 'hydrate.v.01', 'insulate.v.01', \\ 'conduct.v.01', 'resonate.v.01'\end{tabular}                          & \begin{tabular}[c]{@{}c@{}}'rotate', 'move', 'bounce', 'spin', 'sway', 'flythrough', 'fly', \\ 'panning', 'drone', 'run', 'walk', 'drive', 'zoom', 'chase', 'swim', \\ 'movement', 'fall', 'rise','sliding video', 'sliding camera', 'sliding shot',\\  'forward', 'backward', 'leftward', 'rightward', 'upward', 'downward'\end{tabular} \\
                & \multicolumn{1}{l}{}                                                                                                                                                       & \multicolumn{1}{l}{}                                                                                                                                                                                                                                                                                                                      \\
Fluid Motions   & \begin{tabular}[c]{@{}c@{}}'body\_of\_water.n.01', 'fluid.n.01', 'fluid.n.02', \\ 'atmospheric\_phenomenon.n.01', 'deformation.n.02'\end{tabular}                          & \begin{tabular}[c]{@{}c@{}}'fountain', 'float', 'firework', 'fire', 'cloud', 'clouds', 'candle', \\ 'smoke', 'wave', 'inflate', 'melt', 'shrink', 'ripple'\end{tabular}                                                                                                                                                                   \\
                & \multicolumn{1}{l}{}                                                                                                                                                       & \multicolumn{1}{l}{}                                                                                                                                                                                                                                                                                                                      \\
Light Change    & \begin{tabular}[c]{@{}c@{}}'burning.n.01', 'light.n.01', \\ 'light.n.02', 'light.n.04', 'light.n.07', 'light.n.09'\end{tabular}                               & \begin{tabular}[c]{@{}c@{}}'eclipse', 'sunset', 'sunrise', 'firework', 'fire', 'sunbeam', 'sun ray', \\ 'sunshine', 'sunny', 'burn', 'shine', 'luminous', 'glow','explode', 'milky', \\ 'galaxy', 'flash', 'sparkle', 'neon', 'reflection', 'bright', 'candle', 'light'\end{tabular}                                                               \\ \bottomrule
\end{tabular}
$}
\caption{Temporal categories under the "major content" aspect.}
\end{subtable}
\hfill
\begin{subtable}[t]{\textwidth}
\centering
\resizebox{\linewidth}{!}{$
\begin{tabular}{@{}lcc@{}}
\toprule
                                                                       & \textbf{WordNet Synsets}                                                                                                                                                      & \textbf{Key Phrases/Words}                                                                                                                      \\ \midrule
People                                                                 & 'person.n.01', 'people.n.01'                                                                                                                                                  & 'he', 'she', 'men', 'team'                                                                                                                      \\
                                                                       & \multicolumn{1}{l}{}                                                                                                                                                          & \multicolumn{1}{l}{}                                                                                                                            \\
Animals                                                                & 'animal.n.01'                                                                                                                                                                 & None                                                                                                                                            \\
                                                                       & \multicolumn{1}{l}{}                                                                                                                                                          & \multicolumn{1}{l}{}                                                                                                                            \\
Vehicles                                                               & 'vehicle.n.01'                                                                                                                                                                & 'drone'                                                                                                                                         \\
                                                                       & \multicolumn{1}{l}{}                                                                                                                                                          & \multicolumn{1}{l}{}                                                                                                                            \\
Artifects                                                              & 'artifact.n.01'                                                                                                                                                               & None                                                                                                                                            \\
                                                                       &                                                                                                                                                                               &                                                                                                                                                 \\
\begin{tabular}[c]{@{}l@{}}Buildings \&\\ Infrastructures\end{tabular} & 'building.n.01', 'structure.n.01'                                                                                                                                             & 'building', 'cityscape', 'town', 'city'                                                                                                         \\
                                                                       &                                                                                                                                                                               &                                                                                                                                                 \\
\begin{tabular}[c]{@{}l@{}}Scenery \&\\ Natural Object\end{tabular}    & \begin{tabular}[c]{@{}c@{}}'natural\_object.n.01', 'body\_of\_water.n.01', \\ 'geological\_formation.n.01', \\ 'atmospheric\_phenomenon.n.01', 'atmosphere.n.05'\end{tabular} & \begin{tabular}[c]{@{}c@{}}'mountainous', 'fire', 'firework', 'solar eclipse', \\ 'water current', 'water drop', 'cloud', 'desert'\end{tabular} \\
                                                                       &                                                                                                                                                                               &                                                                                                                                                 \\
Plants                                                                 & 'plant.n.02', 'vegetation.n.01'                                                                                                                                               & None                                                                                                                                            \\
                                                                       &                                                                                                                                                                               &                                                                                                                                                 \\
\begin{tabular}[c]{@{}l@{}}Food \& \\ Beverage\end{tabular}            & 'food.n.01', 'food.n.02'                                                                                                                                                      & 'mushroom'                                                                                                                                      \\
                                                                       &                                                                                                                                                                               &                                                                                                                                                 \\
Illustrations                                                          & 'shape.n.02', 'symbol.n.01'                                                                                                                                                   & \begin{tabular}[c]{@{}c@{}}'pattern', 'abstract', 'pattern', 'particle', \\ 'gradient', 'loop', 'graphic'\end{tabular}                          \\ \bottomrule
\end{tabular}
$}
\caption{Spatial categories under the "major content" aspect.}
\end{subtable}
\label{tab:auto_rule_major_content}
\end{table*}

\begin{table*}[htbp]
\centering
\caption{Taxonomy of Predefined WordNet Synsets and Keywords (Attribute Control Aspect).}
\begin{subtable}[t]{\textwidth}
\centering
\resizebox{\linewidth}{!}{$
\begin{tabular}{@{}lcc@{}}
\toprule
                 & \textbf{WordNet Synsets}                                                                                                                                & \textbf{Key Phrases/Words}                                                                                                                                                                                                                                                                                                                                              \\ \midrule
Color            & 'color.n.01'                                                                                                                                           & 'white'                                                                                                                                                                                                                                                                                                                                                                 \\
                 & \multicolumn{1}{l}{}                                                                                                                                    & \multicolumn{1}{l}{}                                                                                                                                                                                                                                                                                                                                                    \\
Camera View      & \begin{tabular}[c]{@{}c@{}}"perspective.n.01", "angle.n.01", "view.n.01", "shot.n.01", "panorama.n.01", \\ "closeup.n.01", "wideangle.n.01", "telephoto.n.01", "macro.n.01", "aerial.n.01", \\ "birdview.n.01", "fisheye.n.01", "tilt.n.01", "pan.n.01", "zoom.n.01", "focus.n.01", \\ "depth.n.01", "frame.n.01", "composition.n.01", "horizon.n.01", "vertical.n.01", \\ "horizontal.n.01", "aspect.n.01"\end{tabular} & \begin{tabular}[c]{@{}c@{}}'macro shot', 'medium shot', 'wide shot', 'close up', 'close-up', 'close view',\\ 'close shot', 'front view', 'front-facing', 'front facing', 'backside view', 'backside shot', \\'profile view', 'profile shot', 'side view', 'side shot', 'top view', 'top-down view', \\'top down view', 'overhead view', 'overhead shot', 'bottom view', 'bottom shot', 'low angle', \\'high angle', 'aerial', 'drone view', "bird's eye view", 'first person', 'first-person', \\'1st person', 'third person', 'third-person', '3rd person'\end{tabular}                               \\
                 &                                                                                                                                                         &                                                                                           \\
Quantity         & 'integer.n.01'                                                                                                                                          & 'a'                                                                                                                                                                                                 \\
                 &                                                                                                                                                         &                                                                               \\

 \bottomrule
\end{tabular}
$}
\caption{Spatial categories under the "attribute control" aspect.}
\end{subtable}
\hfill
\begin{subtable}[t]{\textwidth}
\centering
\resizebox{\linewidth}{!}{$
\begin{tabular}{@{}lcc@{}}
\toprule
                 & \textbf{WordNet Synsets}                                                                                                                                & \textbf{Key Phrases/Words}                                                                                                                                                                                                                                                                                                                                              \\ \midrule
Speed            & \begin{tabular}[c]{@{}c@{}}"velocity.n.01", "speed.n.01", "acceleration.n.01", "deceleration.n.01", \\ "momentum.n.01", "rapidity.n.01", "swiftness.n.01", "haste.n.01", "pace.n.01", \\ "tempo.n.01", "vector.n.01", "terminal.n.01", "escape.n.01", "angular.n.01", \\ "linear.n.01", "relative.n.01", "initial.n.01", "final.n.01", "constant.n.01", \\ "variable.n.01", "light.n.01", "sound.n.01"\end{tabular} & \begin{tabular}[c]{@{}c@{}}'slow', 'fast', 'slowly', 'fastly', 'timelapse', 'timelapse', 'time-lapse', \\ 'stop motion', 'velocity', 'speed', 'acceleration', 'deceleration', 'momentum', \\ 'rapidity', 'swiftness', 'haste', 'pace', 'tempo', 'velocity vector', \\ 'terminal velocity', 'escape velocity', 'angular velocity', 'linear velocity', \\ 'relative velocity', 'initial velocity', 'final velocity', 'constant speed', \\ 'variable speed', 'speed of light', 'speed of sound'\end{tabular}       \\
                 &                                                                                                                                                         &                                                                                                                                                                                                                                                                                                                                                                         \\
Motion Direction & \begin{tabular}[c]{@{}c@{}}"north.n.01", "south.n.01", "east.n.01", "west.n.01", "forward.n.01", \\ "backward.n.01", "left.n.01", "right.n.01", "up.n.01", "down.n.01", "vertical.n.01", \\ "horizontal.n.01", "radial.n.01", "tangential.n.01", "axial.n.01", "diagonal.n.01", \\ "lateral.n.01", "longitudinal.n.01", "reverse.n.01", "clockwise.n.01", \\ "counterclockwise.n.01", "oblique.n.01", "azimuth.n.01", "bearing.n.01", \\ "orientation.n.01", "trajectory.n.01", "normal.n.01", "parallel.n.01", "perpendicular.n.01"\end{tabular} & \begin{tabular}[c]{@{}c@{}}'ahead', 'anticlockwise', 'away from', 'clockwise', 'counterclockwise', 'downward', \\ 'eastbound', 'northbound', 'southbound', 'westbound', 'homeward', 'leftwards',\\  'rightwards', 'upward', 'left', 'right', 'forward', 'backward', 'toward', 'out of',\\  'approach', 'leave', 'against to', 'lift', 'opposite direction'\end{tabular} \\
                 &                                                                                                                                                         &                                                                                                                                                                                                                                                                                                                                                                         \\
Event Order      & \begin{tabular}[c]{@{}c@{}}"sequence.n.01", "order.n.01", "phase.n.01", "step.n.01", "stage.n.01", \\ "priority.n.01", "predecessor.n.01", "successor.n.01", "time.n.01", "past.n.01", \\ "future.n.01", "present.n.01", "duration.n.01", "interval.n.01", "epoch.n.01", \\ "era.n.01", "period.n.01", "schedule.n.01", "calendar.n.01", "chronology.n.01", \\ "timing.n.01", "frequency.n.01", "cycle.n.01", "simultaneous.n.01", "anachronism.n.01"\end{tabular} & 'and then'                                                                                                                                                                                                                                                                                                                                                              \\ \bottomrule
\end{tabular}
$}
\caption{Temporal categories under the "attribute control" aspect.}
\end{subtable}
\label{tab:auto_rule_attribute_control}
\end{table*}

\begin{table*}[t]
\centering
\vspace{2cm}
\caption{Examples of Prompt Classification Results. Misclassified properties are denoted by red text.}
\label{tab:data_example}
\renewcommand{\arraystretch}{1.5} 
\resizebox{0.95\textwidth}{!}{
\begin{tabular}{@{}m{0.45\textwidth}cccc@{}}
\toprule
\centering\textbf{Prompt} & \textbf{Spatial Content} & \textbf{Temporal Content} & \textbf{Spatial Attribute} & \textbf{Temporal Attribute} \\
\midrule

\vspace{0.3cm}\begin{minipage}[c]{0.45\textwidth}two men are engaged in a conversation in a car showroom. The older man, dressed in a black polo shirt, is speaking to the younger man, who is wearing a black t-shirt. The showroom is filled with various cars, including a Bugatti, and posters adorn the walls. The men are standing in front of a car, suggesting that they might be discussing the vehicle$\dots$ \end{minipage} & 
\begin{tabular}[c]{@{}c@{}}People\\ Artifects\\ Buildings\&Infrastructures\\ Vehicles\end{tabular} & 
\begin{tabular}[c]{@{}c@{}}Actions\end{tabular} & 
\begin{tabular}[c]{@{}c@{}}Color\\  Quantity\end{tabular} & 
None \\
\addlinespace[0.3cm]

\cmidrule(lr){1-5}
\vspace{0.2cm}\begin{minipage}[c]{0.45\textwidth}a close-up of a doll with curly brown hair and large, expressive eyes. The doll is wearing a denim dress with a floral design on the front. In the background, there are cans of pumpkin pie filling, suggesting a setting related to baking or cooking$\dots$ \end{minipage} & 
\begin{tabular}[c]{@{}c@{}}Artifects\\ Scenery\&Natural Object\\ Plants\\ Food\&Beverage\end{tabular} & 
\begin{tabular}[c]{@{}c@{}}Actions\\ Light Change\end{tabular} & 
\begin{tabular}[c]{@{}c@{}}Color\\ Camera View\\ Quantity\end{tabular} & 
None \\
\addlinespace[0.3cm]
\cmidrule(lr){1-5}
\vspace{0.3cm}\begin{minipage}[c]{0.45\textwidth}three young men standing on a race track, smiling and posing for the camera. They are dressed in casual clothing, with one wearing a tie-dye shirt, another in a tank top, and the third in a black t-shirt with a cat design$\dots$ \end{minipage} & 
\begin{tabular}[c]{@{}c@{}}People\\ Animals\\ Artifects\\ Buildings\&Infrastructures\\ Scenery\&Natural Object\end{tabular} & 
\begin{tabular}[c]{@{}c@{}}Actions\\ Fluid Motions\end{tabular} & 
\begin{tabular}[c]{@{}c@{}}Color\\ Quantity\end{tabular} & 
None \\
\addlinespace[0.3cm]

\cmidrule(lr){1-5}
\vspace{0.2cm}\begin{minipage}[c]{0.45\textwidth}a white Mustang GT500 parked in a parking lot. The car is adorned with blue stripes on the hood and side, and it features a black grille and black wheels. The car is parked in a spot with a white line marking the boundary $\dots$ \end{minipage} & 
\begin{tabular}[c]{@{}c@{}}People\\ \color{red}{Animals}\\ Vehicles\\ Artifects\\ Illustrations\end{tabular} & 
\begin{tabular}[c]{@{}c@{}}Actions\\ Kinetic Motions\\ Light Change\end{tabular} & 
\begin{tabular}[c]{@{}c@{}}Color\\ Camera View\\ Quantity\end{tabular} & 
None \\
\addlinespace[0.3cm]

\cmidrule(lr){1-5}
\vspace{0.2cm}\begin{minipage}[c]{0.45\textwidth}a young man standing next to a red Volkswagen Beetle car on a road surrounded by trees. The man is dressed in a casual blue shirt and jeans, and he has his hands clasped together. The car is parked on the side of the road, and it has a shiny red finish$\dots$ \end{minipage} & 
\begin{tabular}[c]{@{}c@{}}People\\ \color{red}{Animals} \\ Artifects\\ Scenery\&NaturalObject\\ Plants\end{tabular} & 
\begin{tabular}[c]{@{}c@{}}Actions\\ \color{red}{Fluid Motions}\\ LightChange\end{tabular} & 
\begin{tabular}[c]{@{}c@{}}Color\\ Camera View\\ Quantity\end{tabular} & 
None \\
\addlinespace[0.3cm]

\cmidrule(lr){1-5}
\vspace{0.2cm}\begin{minipage}[c]{0.45\textwidth}a white SUV driving down a snowy road. The vehicle is equipped with a roof rack and has a distinctive orange stripe running along its side. The road is lined with trees, and the snow-covered landscape suggests a cold, winter day $\dots$  \end{minipage} & 
\begin{tabular}[c]{@{}c@{}}People\\ Vehicles\\ Artifects\\ Buildings\&Infrastructures\\ Scenery\&NaturalObject\\ Plants\end{tabular} & 
\begin{tabular}[c]{@{}c@{}}Actions\\ Kinetic Motions\end{tabular} & 
\begin{tabular}[c]{@{}c@{}}Color\\ Camera View\\ Quantity\end{tabular} & 
\begin{tabular}[c]{@{}c@{}}Speed\\ Motion Direction\end{tabular} \\
\addlinespace[0.3cm]

\bottomrule
\vspace{2cm}
\end{tabular}}
\end{table*}

The WordNet synsets and corresponding keyword lists are presented in ~\cref{tab:auto_rule_major_content,tab:auto_rule_attribute_control}. It should be acknowledged that, despite our best efforts to ensure classification accuracy, the long-text nature of our prompts inevitably introduced a certain degree of misclassification. Representative examples of such prompts, including incorrectly categorized cases, are provided in ~\cref{tab:data_example}. However, given the substantial volume of data (approximately over 400,000 entries) and the considerable length of individual prompts, it was not feasible to employ large language models (LLMs) for this classification task. It is worth noting that we also attempted to utilize closed-source LLMs such as GPT-4, but found that this approach still could not fully resolve the issue.

\begin{table}[t]
\centering
\caption{Balance Comparison with Prior Datasets($\alpha=2.0$).}
\label{tab:pgbs_comparison}
\resizebox{\linewidth}{!}{
\begin{tabular}{lcccc}
\toprule
\textbf{Metric} & \textbf{AIGVDBench} & \textbf{GVF} & \textbf{GenVidBench} \\
\midrule
\multicolumn{4}{l}{\textbf{Complete Uniformity (CU) by Attribute}} \\
\midrule
Spatial\_Content & 0.9592 & 0.7818 & 0.8495 \\
Temporal\_Content & 0.8785 & 0.6866 & 0.7677 \\
Spatial\_Attribute & 0.9942 & 0.4069 & 0.7492 \\
Temporal\_Attribute & 1.0000 & 0.8762 & 0.9966 \\
\midrule
\multicolumn{4}{l}{\textbf{Global Metrics}} \\
\midrule
MCU & 0.9580 & 0.6879 & 0.8408 \\
UCO & 0.0023 & 0.0308 & 0.0095 \\
PGBS & \textbf{0.9557} & 0.6667 & 0.8328 \\
\midrule
\multicolumn{4}{l}{\textbf{Prompt Length Analysis}} \\
\midrule
Avg. Prompt Length & 552.39 & 46.29 & 88.69 \\
\bottomrule
\end{tabular}}
\end{table}
\subsection{Balance Comparison with Prior Datasets}
\label{7.5}

To quantitatively evaluate the balance and diversity of the dataset, we introduced \textbf{Penalized Global Balance Score (PGBS).}
At the attribute level, we first introduce the \textbf{Relative Uniformity (RU)} to quantify the distribution uniformity of values within a single attribute: 
\begin{equation}
    RU_i=\frac{H_i}{H_i^{unif}}
\end{equation}
where denotes the empirical entropy of the $i$-th attribute, $H_i^{unif}= \log n_i$ represents the entropy under an ideal uniform distribution, and $n_i$ is the total number of theoretical categories for that attribute. An \textbf{RU} value close to 1 indicates that the distribution of the attribute approaches uniformity.
Considering that datasets often suffer from missing categories, we extend RU to \textbf{Complete Uniformity (CU)} as follows:
\begin{equation}
    CU_i=RU_i \times (R_i)^\alpha
\end{equation}
where $R_i=\frac{|S_i^{obs}|}{n_i}$
 is the completeness ratio, $|S_i^{obs}|$ denotes the number of actually observed categories, and $\alpha$ is an adjustable penalty intensity parameter. This design ensures that when category missing occurs (i.e., $R_i$<1), the CU value is penalized appropriately, thereby more accurately reflecting the impact of distribution completeness on balance.
After obtaining the CU values for all attributes, we further construct two core metrics from a global perspective. The \textbf{Mean Complete Uniformity (MCU)}, defined as
\begin{equation}
MCU=\frac{1}{m}\sum_{i=1}^{m} CU_i
\end{equation}
measures the average balance level across all $m$ attributes. The \textbf{Uniformity Coordination (UCO)}, given by
\begin{equation}
    UCO= \frac{1}{m}\sum_{i=1}^{m}(CU 
i- MCU)^2
\end{equation}
captures the dispersion of balance across attributes, reflecting their coordination consistency.
Finally, by integrating MCU and UCO, we construct the \textbf{Penalized Global Balance Score}:
\begin{equation}
    PGBS=MCU\times(1-UCO)
\end{equation}
PGBS simultaneously accounts for the uniformity of distributions within attributes and the coordination among attributes, resulting in a composite score bounded between $[0, 1]$. A higher score indicates a more desirable overall balance characteristic of the dataset. The balance of the dataset and the effectiveness of the Attribute Balancing Selection Algorithm are evaluated using the PGBS metric. The experimental results are shown in ~\cref{tab:pgbs_comparison}. \textbf{Our dataset achieves the highest balance score despite employing more extensive prompts. More significantly, this result was attained through a systematic and scalable algorithmic approach, eliminating the need for manual selection and its inherent biases.}
\begin{table}[H]
\centering
\caption{Impact of Video Compression Formats on Detector Performance.}
\label{tab:cross_format_generalization}
\resizebox{\linewidth}{!}{
\begin{tabular}{lccccc}
\toprule
\multirow{2}{*}{Model} & \multirow{2}{*}{Training Scheme} & \multicolumn{2}{c}{Accuracy (\%)} & \multirow{2}{*}{Gap} \\
\cmidrule(lr){3-4}
 & & H.264 Test & MPEG-4 Test & \\
\midrule
\multirow{2}{*}{I3D} 
 & Mixed & 100.00 & 100.00 & 0.00 \\
 & Unified & 100.00 & 100.00 & 0.00 \\
\cmidrule{1-5}
\multirow{2}{*}{X3D} 
 & Mixed & 99.93 & 99.93 & 0.00 \\
 & Unified & 99.97 & 100.00 & \textbf{0.033} \\
\cmidrule{1-5}
\multirow{2}{*}{CNN\_Spot} 
 & Mixed & 99.97 & 99.97 & 0.00 \\
 & Unified & 99.97 & 99.97 & 0.00 \\
\cmidrule{1-5}
\multirow{2}{*}{UnivFD} 
 & Mixed & 99.83 & 99.83 & 0.00 \\
 & Unified & 99.83 & 99.83 & 0.00 \\
\bottomrule
\end{tabular}}
\vspace{0.2cm}
\begin{itemize}
    \item \small Mixed: Real videos (MPEG-4), Generated videos (H.264); 
    \item \small Unified: All videos compressed with H.264. 
\end{itemize}     
\end{table}
\begin{figure*}[t]
\centering
\caption{Logit distribution comparison across different models under two training schemes.}

\begin{tabular}{cccc}
\parbox{0.22\textwidth}{
  \centering
  \includegraphics[width=\linewidth]{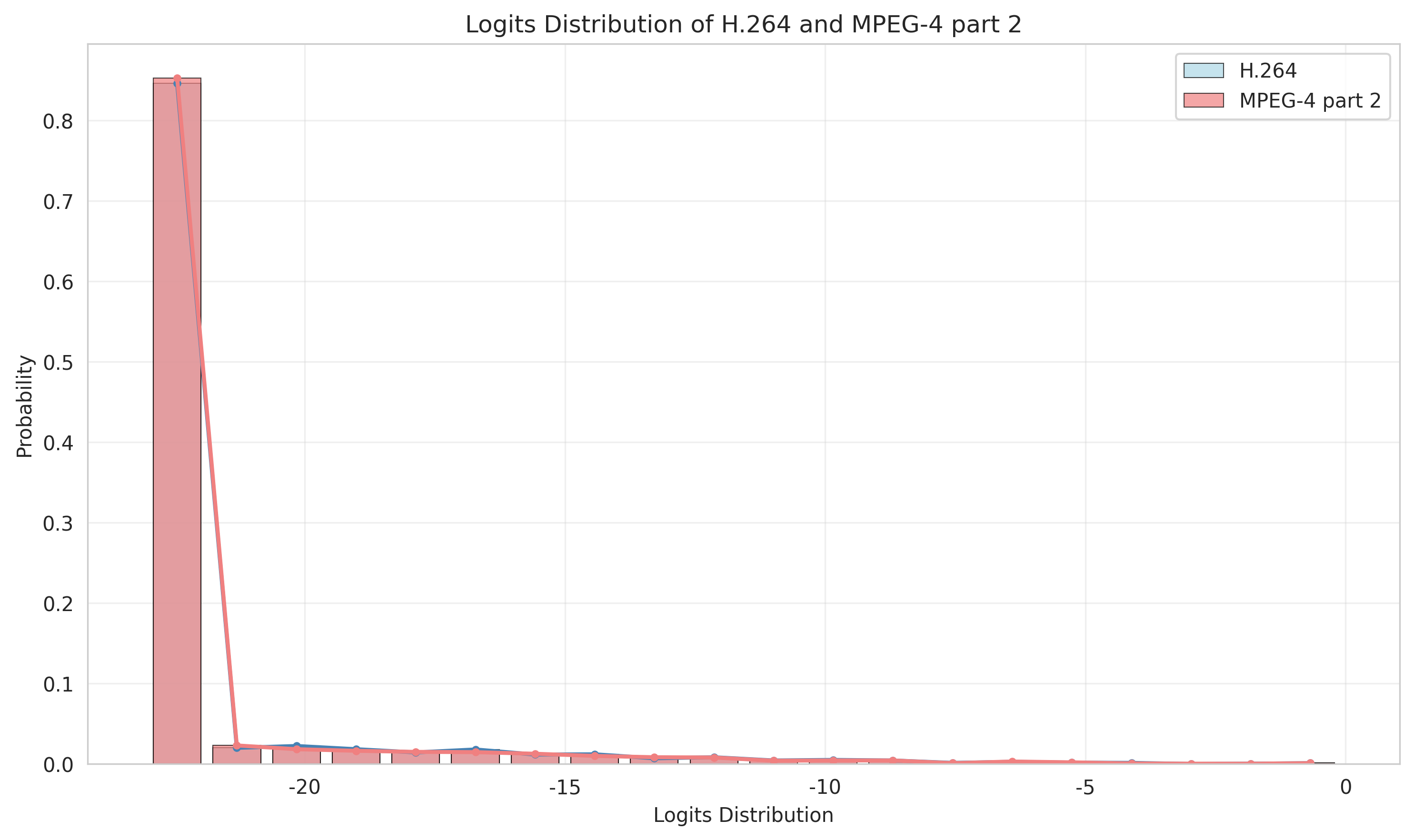}\\
  \small(a) I3D (H.264)
} &
\parbox{0.22\textwidth}{
  \centering
  \includegraphics[width=\linewidth]{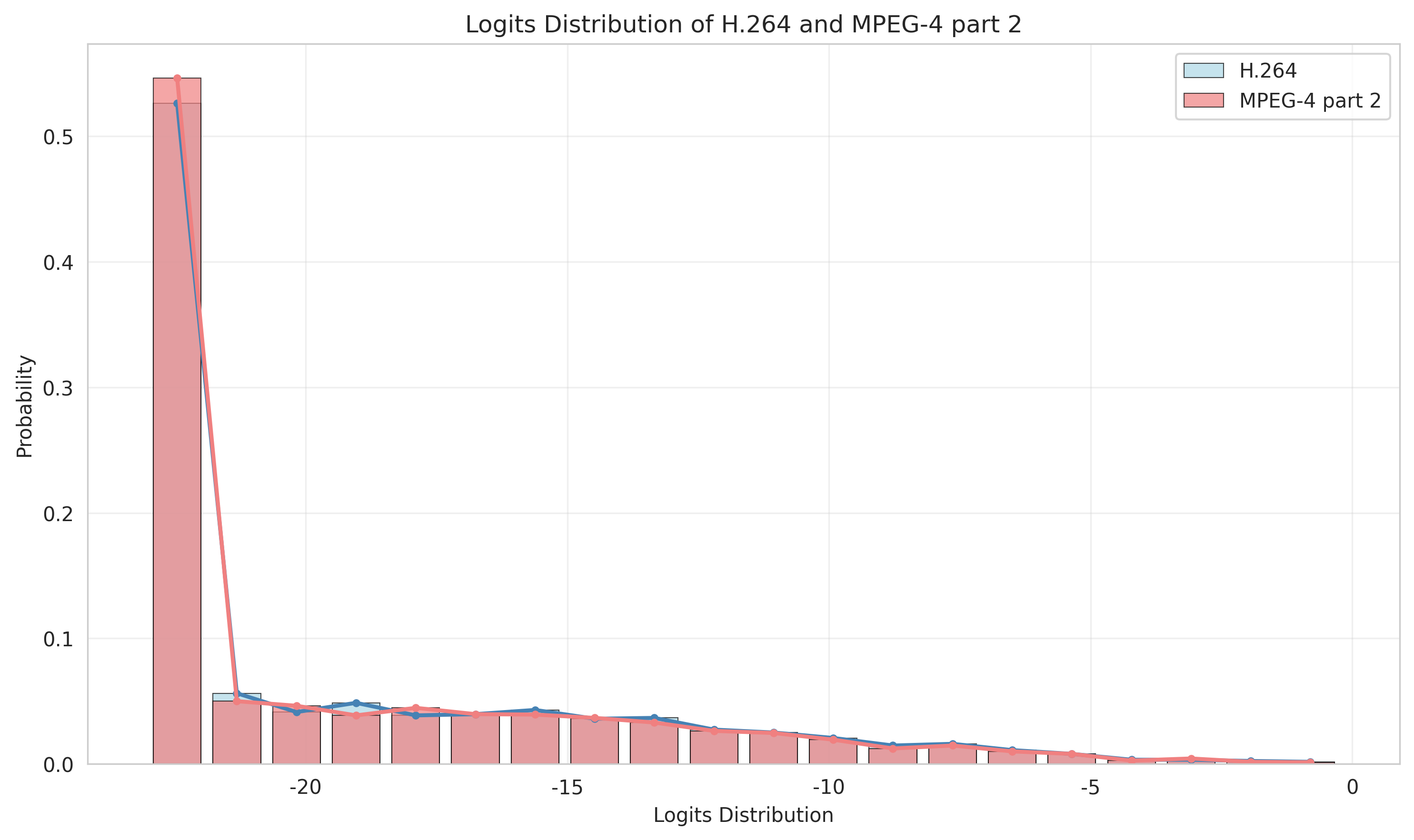}\\
  \small(b) I3D (Mixed)
} &
\parbox{0.22\textwidth}{
  \centering
  \includegraphics[width=\linewidth]{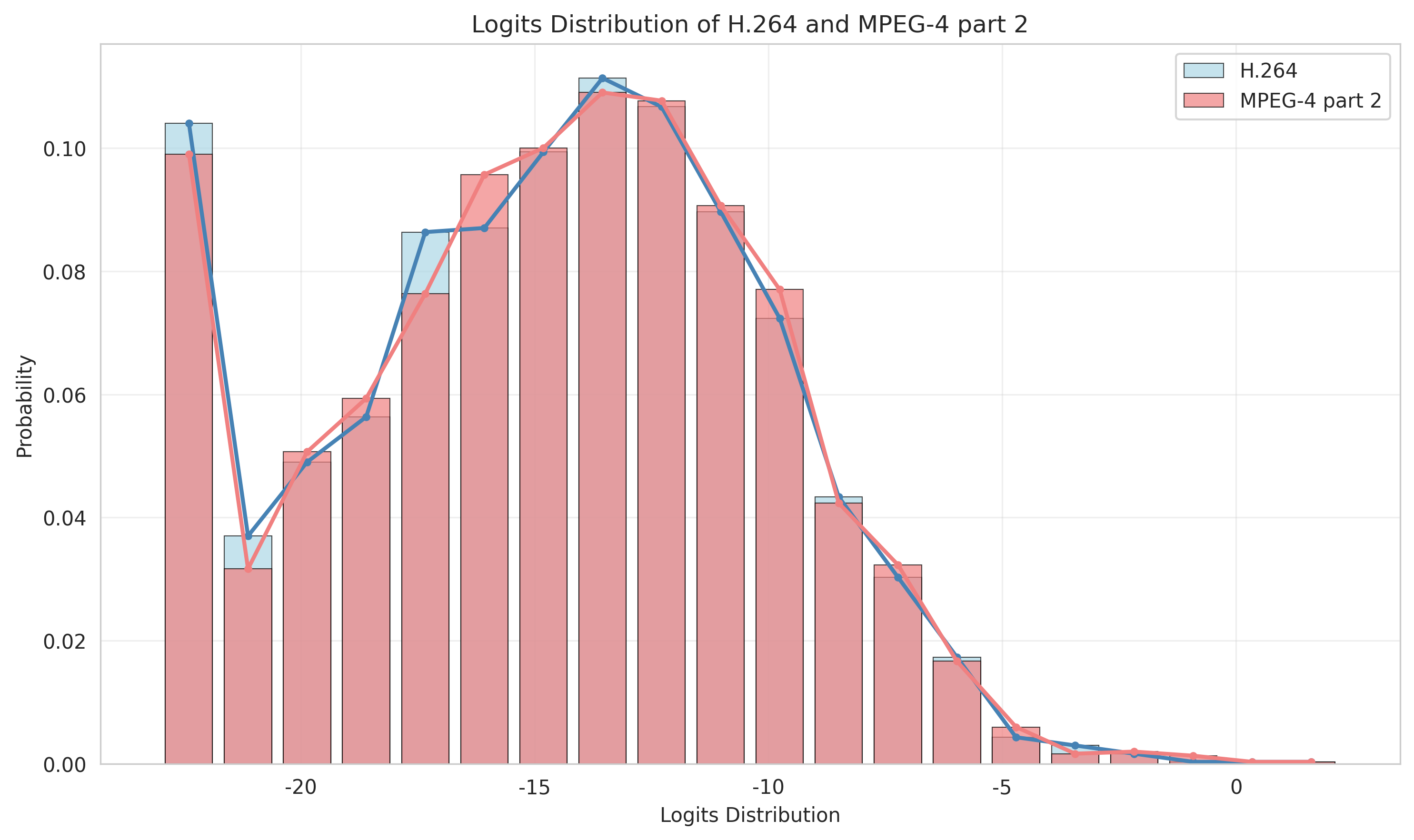}\\
  \small(c) X3D (H.264)
} &
\parbox{0.22\textwidth}{
  \centering
  \includegraphics[width=\linewidth]{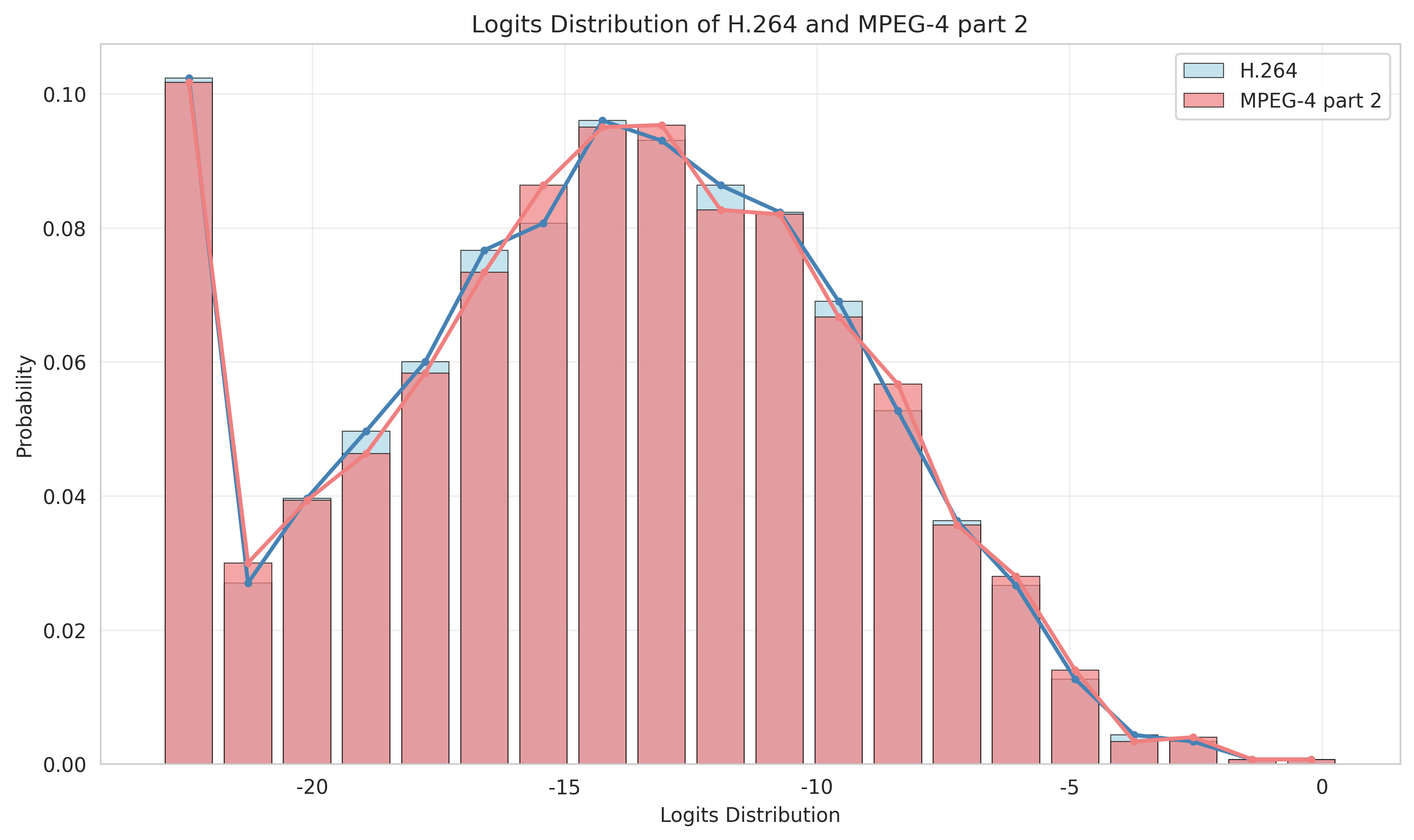}\\
  \small(d) X3D (Mixed)
} \\
\end{tabular}

\begin{tabular}{cccc}
\parbox{0.22\textwidth}{
  \centering
  \includegraphics[width=\linewidth]{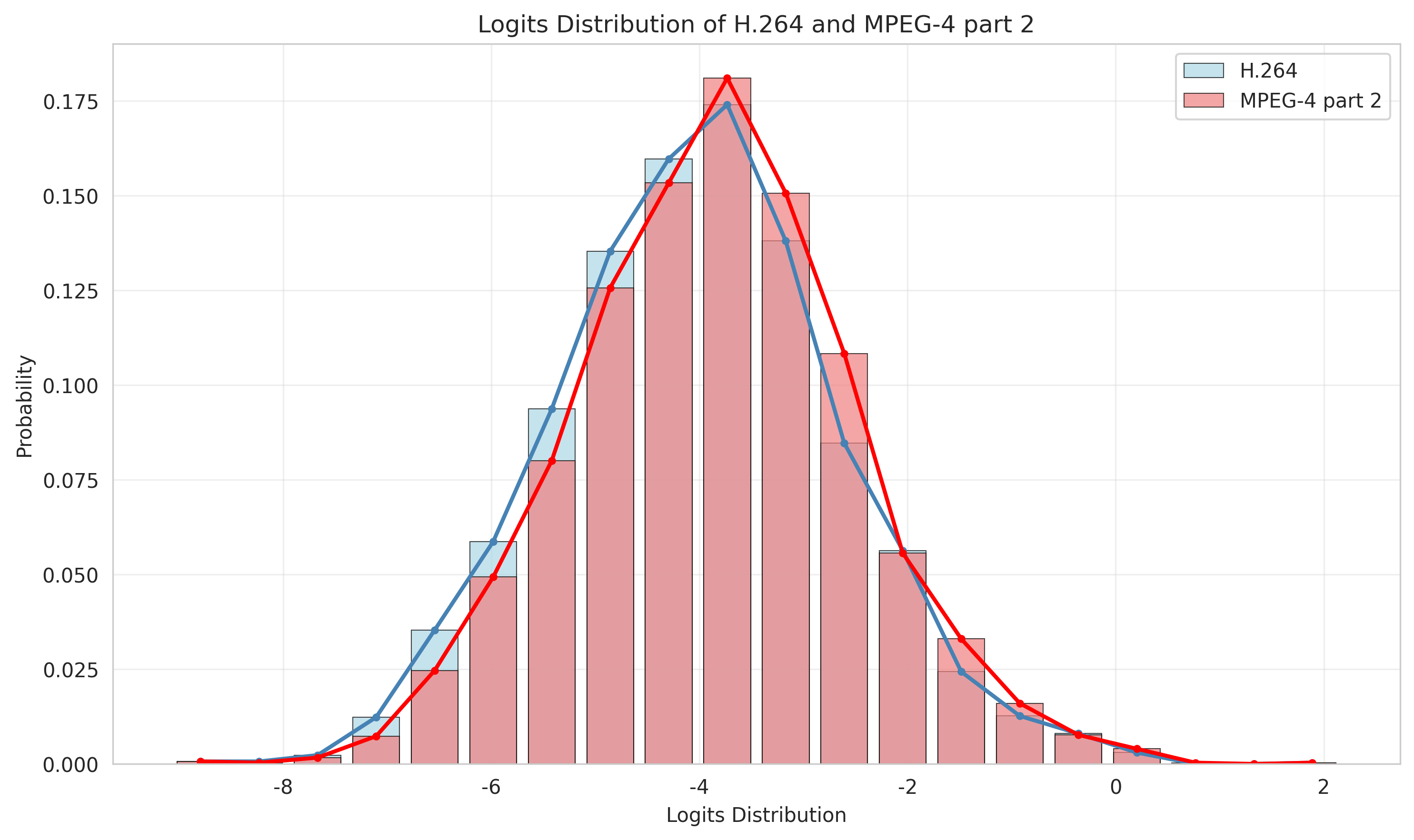}\\
  \small(e) UnivFD (H.264)
} &
\parbox{0.22\textwidth}{
  \centering
  \includegraphics[width=\linewidth]{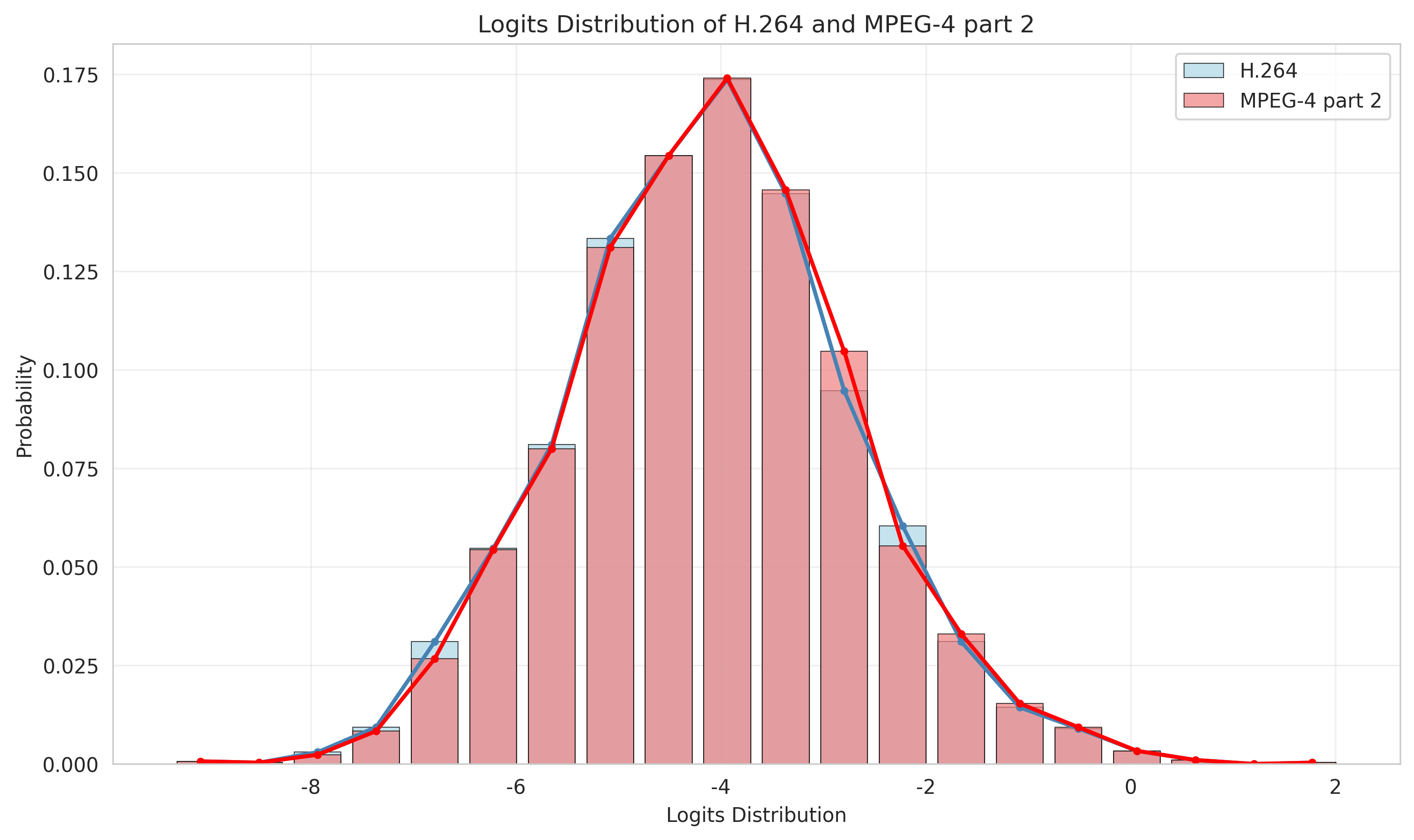}\\
  \small(f) UnivFD (Mixed)
} &
\parbox{0.22\textwidth}{
  \centering
  \includegraphics[width=\linewidth]{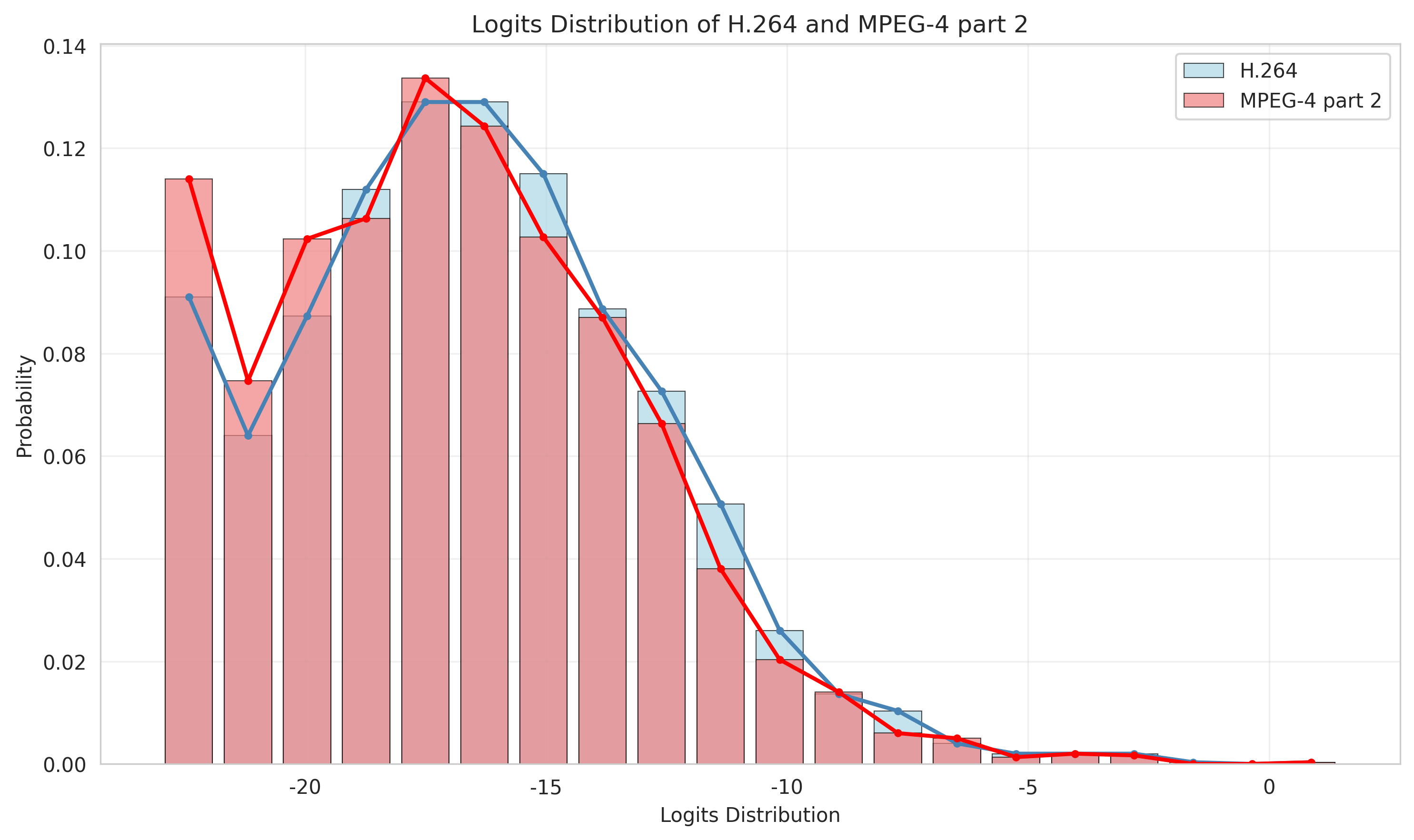}\\
  \small(g) CNNSpot (H.264)
} &
\parbox{0.22\textwidth}{
  \centering
  \includegraphics[width=\linewidth]{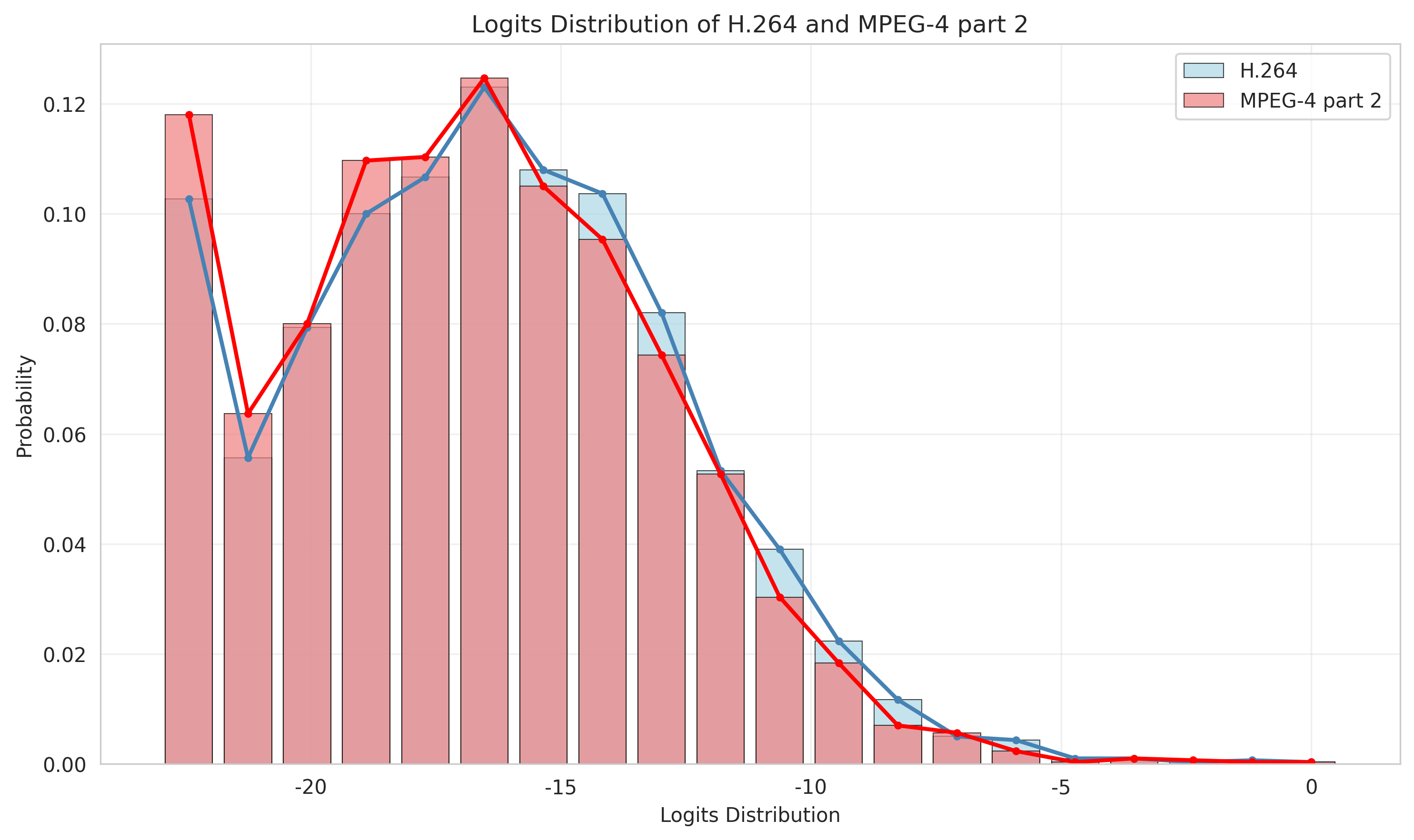}\\
  \small (h) CNNSpot (Mixed)
} \\
\end{tabular}

\vspace{0.2cm}
\footnotesize
\textbf{Note:} H.264 = models trained with unified H.264 compression; Mixed = models trained with mixed compression (real videos: MPEG-4 Part 2, generated samples: H.264). The figure shows feature distributions when tested on datasets containing both compression formats. In all subfigures, the \textcolor{lightcoral}{light coral} curves represent test results on MPEG-4 Part 2 compressed videos, while the \textcolor{lightblue}{light blue} curves represent test results on H.264 compressed videos.

\label{fig:distribution_comparison_compact}
\end{figure*}
\subsection{Discussion on Video Encoding Methods}
\label{7.6}
In early explorations of generated image detection~\cite{zhu2023genimage,Universal,wang2023dire}, an interesting phenomenon was observed: real videos were typically stored in PNG format, while generated images were saved as JPEG. Models~\cite{wang2023dire} trained on such datasets could easily distinguish between positive and negative samples, achieving high accuracy. This unintentional format discrepancy led the models to learn the compression differences between PNG and JPEG, rather than the intrinsic distinctions between authentic and generated content.

Motivated by this finding, we investigated whether a similar effect would occur when real videos are compressed using MPEG-4 Part 2 and generated samples are compressed with H.264. As shown in~\cref{tab:cross_format_generalization}, we selected several common network architectures and trained them under two conditions: one where real videos were compressed with MPEG-4 Part 2 and generated samples with H.264, and another where all videos were uniformly compressed with H.264. The models were then evaluated on real videos compressed with either MPEG-4 Part 2 or H.264.
From the perspective of accuracy metrics, the impact of this compression discrepancy is negligible. Only the X3D model, trained on the mixed-compression dataset (real: MPEG-4 Part 2, generated: H.264), exhibited any noticeable effect. Furthermore, although we observed some variations in output logits~\cref{fig:distribution_comparison_compact}, these changes had almost no influence on the final accuracy. Nevertheless, to mitigate potential confounding factors, all videos in AIGVDBench were uniformly compressed using H.264.

\subsection{Training details}
\label{7.7}
\textbf{Training Protocol for Non-VLM Models: } To maximize the inherent potential of each individual model (excluding VLMs), we deliberately refrained from standardizing key hyperparameters such as learning rates and data augmentation schemes. This decision was motivated by the fact that such parameters are intrinsically tied to each model's architecture, and arbitrary unification could compromise their original performance. Furthermore, given the large number of models involved, enforcing acommon parameter set was deemed impractical.
To mitigate overfitting, we setan upper limit of 50 training epochs for all models, ensuring that each had adequately converged before termination. For models originally proposed with fewer than 50 epochs in their respective papers, we adhered to their prescribed training length. In terms of input resolution, most models were trained and evaluated using 256×256 images, while Vision Transformer (ViT)-based models utilized 224×224 inputs.
For video classification models, we adopted the MMAction2~\cite{MMAction2} and followed the recommended hyperparameter settings. For generated image detection, early-stage methods were implemented based on the AIGCDetectBenchmark~\cite{zhong2023patchcraft} repository, while other models were trained using officially released code. Across all experiments, the only modification made was to the data input pipeline; all other training configurations were preserved as provided in the original implementations.
\begin{table}[H]
\centering
\caption{Prompt template assignment for VLM models.}
\label{tab:vlm_prompts}
\resizebox{\linewidth}{!}{
\begin{tabular}{ll}
\toprule
\textbf{Prompt Template Group} & \textbf{Models} \\
\midrule
\multirow{7}{*}{Common Template} & Qwen2.5-VL-3B-Instruct \\
 & Qwen2.5-VL-7B-Instruct \\
 & Qwen2.5-VL-32B-Instruct \\
 & Kimi-VL-A3B-Instruct \\
 & FastVLM-Apple-7B \\
 & InternVL-8B \\
 & LLaVA-1.5-7B \\
\midrule
\multirow{3}{*}{Individual Templates} & Emu3-Stage1  \\
 & DeepSeek-VL-7B \\
 & DeepSeek-VL2-Small (DeepSeek-VL2) \\
\bottomrule
\end{tabular}}
\end{table}
\noindent\textbf{VLM Models:} It must be acknowledged that Vision-Language Models (VLMs) have demonstrated remarkable potential across a range of tasks. In our experimental setup, we adopted a prompt-based zero-shot approach to objectively evaluate VLMs' capabilities in generated video detection. Due to variations in prompt-following styles across different models, we adjusted the prompts for models that initially exhibited poor classification performance to ensure a fair comparison of their ``basic classification capabilities''.  The specific prompt templates are illustrated in ~\cref{prompt1,prompt2,prompt3,prompt4}. ~\cref{prompt1} presents the common prompt template employed for multiple VLM models, while ~\cref{prompt2,prompt3,prompt4} display the individually tailored prompt templates designed for Emu3-Stage1, DeepSeek-VL-7B, and DeepSeek-VL2-Small respectively.   For Vision-Language Models (VLMs) lacking temporal reasoning capabilities, we implement a 

\begin{figure}[t]
\begin{tcolorbox}[
  colback=white,
  colframe=black,
  colupper=gray!70,
  fontupper=\footnotesize,
  coltitle=white,
  coltext=black,
  boxrule=0.5mm,
  title=Common Prompt Template
]

\#\# Critical Instruction\\
You MUST respond with EXACTLY ONE WORD only.\\

\#\# Task\\
Analyze this video frame for deepfake detection. Determine if the content is real or AI-generated.\\

\#\# Detection Criteria\\
Look for the following features:\\
\begin{itemize}
\setlength\itemsep{0.2em}
\item Visual artifacts
\item Inconsistencies
\item Unnatural movements  
\item Authentication markers
\end{itemize}

\#\# Response Format\\
\begin{itemize}
\setlength\itemsep{0.2em}
\item If content is authentic/real: Answer 'REAL'
\item If content is AI-generated/deepfake: Answer 'FAKE'
\end{itemize}

\#\# Strict Rules\\
\begin{itemize}
\setlength\itemsep{0.2em}
\item Output ONLY the classification word
\item No explanations, descriptions, or other text allowed
\item Any response other than 'REAL' or 'FAKE' will be considered invalid
\end{itemize}

\end{tcolorbox}
\caption{Common prompt template for AI-generated video detection, used with multiple VLM models.}
\label{prompt1} 
\end{figure}

\begin{figure}[t]
\begin{tcolorbox}[
  colback=white,
  colframe=black,
  colupper=gray!70,
  fontupper=\footnotesize,
  coltitle=white,
  coltext=black,
  boxrule=0.5mm,
  title=Emu3-Stage1 Prompt Template
]

\#\# Role\\
You are analyzing video frames for deepfake detection.\\

\#\# Task\\
Look carefully at these frames and determine if they show real content or AI-generated/deepfake content.\\

\#\# Detection Criteria\\
Consider the following features:\\
\begin{itemize}
\setlength\itemsep{0.2em}
\item Visual artifacts
\item Inconsistencies
\item Unnatural movements
\item Authenticity markers
\end{itemize}

\#\# Examples\\
\begin{itemize}
\setlength\itemsep{0.2em}
\item If content is authentic: Answer 'REAL'
\item If content is AI-generated/deepfake: Answer 'FAKE'
\end{itemize}

\#\# Important Rules\\
\begin{itemize}
\setlength\itemsep{0.2em}
\item You must respond with ONLY one word - either 'REAL' or 'FAKE'
\item Do not provide any explanation, description, or other text
\item Just the single classification word
\end{itemize}

\end{tcolorbox}
\caption{Individual prompt template for AI-generated video detection with Emu3-Stage1 model.}
\label{prompt2} 
\end{figure}

\vspace{0.2cm}

\begin{figure}[H]
\begin{tcolorbox}[
  colback=white,
  colframe=black,
  colupper=gray!70,
  fontupper=\footnotesize,
  coltitle=white,
  coltext=black,
  boxrule=0.5mm,
  title=DeepSeek-VL-7B Prompt Template
]
\#\# $<$Image\_placeholder$>$\\

\#\# Role\\
You are a forensic deepfake detector. Inspect the provided video frame carefully.\\

\#\# Task\\
Decide whether the frame is REAL (authentic) or FAKE (AI-generated / manipulated).\\

\#\# Focus Areas\\
Analyze the following aspects:\\
\begin{itemize}
\setlength\itemsep{0.2em}
\item Artifacts
\item Lighting inconsistencies
\item Texture glitches
\item Boundary errors
\item Unnatural details
\end{itemize}

\#\# Critical Instruction\\
Respond with EXACTLY ONE WORD: 'REAL' or 'FAKE'.\\

[CONTEXT PLACEHOLDER]

Classification:

\end{tcolorbox}
\vspace{0.2cm}
\caption{Individual prompt template for AI-generated video detection with DeepSeek-VL-7B model.}
\label{prompt3} 
\end{figure}

\begin{figure}[t]
\begin{tcolorbox}[
  colback=white,
  colframe=black,
  colupper=gray!70,
  fontupper=\footnotesize,
  coltitle=white,
  coltext=black,
  boxrule=0.5mm,
  title=DeepSeek-VL2 Prompt Template
]

\#\#  $<$Image$>$\\

\#\# Role\\
You are a forensic deepfake detector. Some samples are suspicious.\\

\#\# Task\\
Analyze this video frame for deepfake detection. Determine if the content is real or AI-generated.\\

\#\# Detection Criteria\\
Look for the following features:\\
\begin{itemize}
\setlength\itemsep{0.2em}
\item Visual artifacts
\item Inconsistencies
\item Unnatural movements
\item Authentication markers
\end{itemize}

\#\# Critical Rule\\
Respond ONLY with one word: REAL or FAKE.\\

[CONTEXT PLACEHOLDER]

Classification:

\end{tcolorbox}
\vspace{0.2cm}
\caption{Individual prompt template for AI-generated video detection with DeepSeek-VL2-Small model.}
\label{prompt4} 
\end{figure}

\begin{figure*}[t]
\centering
\includegraphics[width=0.92\linewidth]{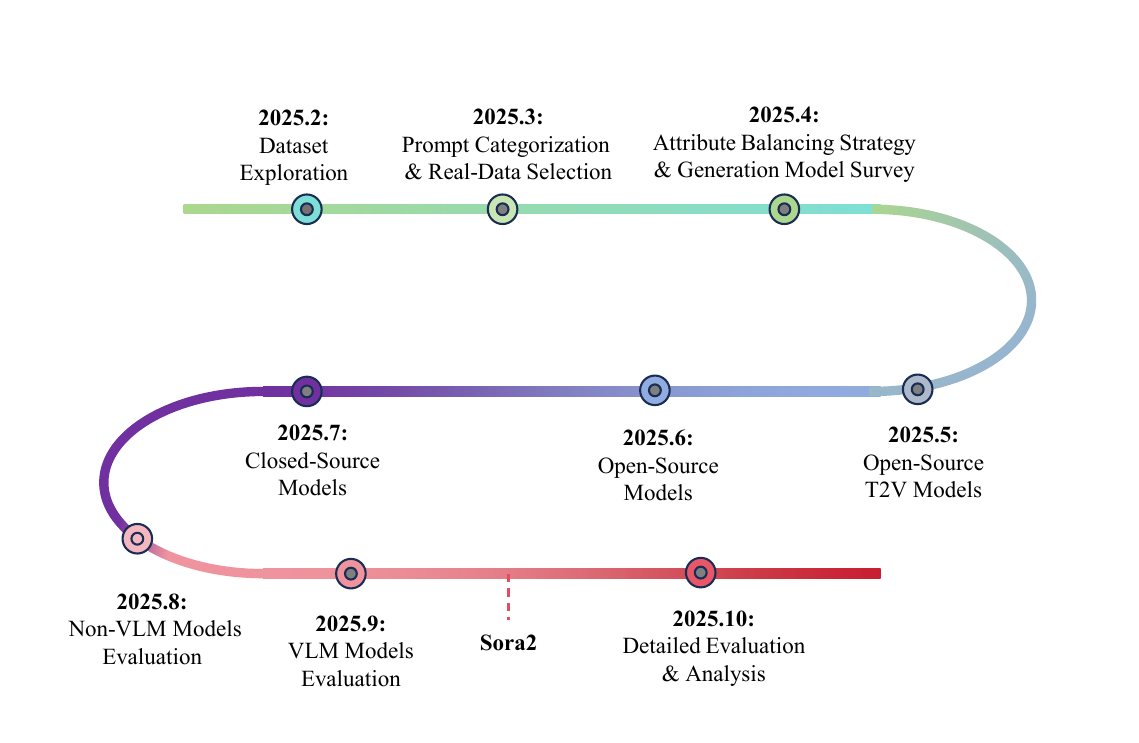} 
\caption{Timeline for AIGVDBench: Key Milestones and Development Phases from February to October 2025.}
\label{fig:research_timeline}
\end{figure*}
\noindent frame-based evaluation protocol: a video is labeled as "REAL" if at least one frame is classified as "REAL", and as "FAKE" only if all frames are consistently classified as "FAKE". In cases where the model fails to produce a valid output, the video is categorized as "No Answer".

\noindent\textbf{Hardware Configuration:} During the dataset construction phase, we utilized eight RTX 3090 GPUs and one 80GB A800 GPU. For the evaluation stage, all non-VLM models were assessed using four RTX 3090 GPUs, while the evaluation of VLM models was conducted on a single  80GB A800 GPU.

\subsection{Timeline of AIGVDBench Construction}
\label{7.8}

As illustrated in ~\cref{fig:research_timeline}, the project followed a structured development process with key milestones achieved from February to October 2025.

\section{Limitation}
\label{8}
In this paper, we introduce AIGVDBench and conduct a series of explorations based on it. While substantial efforts and resources have been devoted to this work, several limitations should be acknowledged. From a data perspective, although we aimed to encompass a wide range of existing generative models, we were constrained by computational resources and generation speed, which necessitated the use of smaller-parameter versions of some models. Additionally, due to the high cost associated with generating samples using closed-source models, we resorted to collecting existing videos rather than regenerating the entire test set from scratch. Furthermore, certain closed-source models, such as Sora 2 and Veo 3, were excluded due to regional access restrictions.

In terms of evaluation, while we incorporated several state-of-the-art models, storage and time limitations led to the exclusion of certain approaches, such as reconstruction-based methods. All analytical findings are derived from rigorously designed experiments; however, we cannot guarantee absolute correctness, as the dataset construction process may have introduced extraneous factors, despite our careful efforts to minimize such interference.

We hope that this work will facilitate and accelerate research in the field of generated video detection.



\end{document}